\documentclass{article}

\usepackage[preprint]{neurips_2020}
\usepackage[utf8]{inputenc} 
\usepackage[T1]{fontenc}    
\usepackage{hyperref}       
\usepackage{url}            
\usepackage{booktabs}       
\usepackage{amsfonts}       
\usepackage{nicefrac}       
\usepackage{microtype}      
\usepackage{subfigure}
\usepackage{algorithm}
\usepackage{algorithmicx}
\usepackage{algpseudocode}
\usepackage{amsmath}
\usepackage{amsthm}
\usepackage{graphicx}
\usepackage{wrapfig}
\usepackage{epstopdf}
\setcitestyle{numbers,square, comma,sort&compress}
\newcommand{\norm}[1]{\left\lVert#1\right\rVert}
\title{Robust Federated Recommendation System}

\author{%
  Chen Chen\textsuperscript{\rm 1}\thanks{Equal Contributions.}$\;\,$, Jingfeng Zhang\textsuperscript{\rm 2}$^{*}$, Anthony K. H. Tung\textsuperscript{\rm 2}, Mohan
Kankanhalli\textsuperscript{\rm 2}, Gang Chen\textsuperscript{\rm 1} \\
  \textsuperscript{\rm 1} College of Computer Science and Technology,
  Zhejiang University, China \\
  \textsuperscript{\rm 2}  School of Computing, National University of Singapore, Singapore \\
  \texttt{cc33@zju.edu.cn, \{j-zhang, atung, mohan\}@comp.nus.edu.sg, cg@zju.edu.cn}
}

\begin{document}

\maketitle

\begin{abstract}

Federated recommendation systems can provide good performance without collecting users' private data, making them attractive. However, they are susceptible to low-cost poisoning attacks that can degrade their performance. In this paper, we develop a novel federated recommendation technique that is robust against the poisoning attack where Byzantine clients prevail. We argue that the key to Byzantine detection is monitoring of gradients of the model parameters of clients. We then propose a robust learning strategy where instead of using model parameters, the central server computes and utilizes the gradients to filter out Byzantine clients. Theoretically, we justify our robust learning strategy by our proposed definition of Byzantine resilience. Empirically, we confirm the efficacy of our robust learning strategy employing four datasets in a federated recommendation system.
\end{abstract}
\section{Introduction}
Recommendation systems (RS), which are machine learning models that try to predict the user (client) preferences, are increasingly being deployed commercially
~\cite{ma2019learning,ben2018game,conf/icml/PurushothamL12,wang2018modeling,conf/nips/VolkovsYP17,wang2015collaborative}. Learning a conventional RS requires centralized storage of clients' data,
which raises privacy concerns~\cite{beye2013social}.
In order to make RS compliant with privacy regulation while preserving the quality of recommendation service, federated recommendation system (FRS) has attracted recent attention~\cite{journals/corr/abs-1901-09888,journals/corr/abs-1906-05108,journals/corr/abs-2003-00602}. An FRS is able to learn a quality recommendation model without holding clients' data centrally.


However, due to the decentralized data storage, FRS is susceptible to low-cost poisoning attacks~\cite{conf/uai/XieKG19,conf/nips/BlanchardMGS17,fang2019local,conf/icml/MhamdiGR18,conf/nips/LiWSV16,journals/iacr/WangT15}. An unscrupulous competitor can easily create a small number of malicious clients, i.e., Byzantine clients, to bias the recommendations \cite{baruch2019little,bagdasaryan2018backdoor,conf/uai/XieKG19,journals/iacr/WangT15}. Consequently, a non-robust federated recommendation system with irrelevant recommendations will soon lose its reputation and trust. Therefore, developing a federated recommendation system that is robust against  poisoning attacks is necessary.


It appears that learning a robust FRS can directly employ existing defense strategies, i.e., utilize the model parameters among clients to detect Byzantine clients~\cite{conf/nips/BlanchardMGS17,conf/icml/DamaskinosMGPT18,conf/icml/MhamdiGR18,journals/corr/abs-1903-06996,journals/corr/abs-2002-00211}.
The tacit assumption of the existing defense strategies is that the clients optimize their local models with (stochastic) gradient descent (SGD), where the update of the model parameter is identical to the gradient (of the model parameter). However, in learning FRS, SGD often leads to poor performance due to the problem of  vanishing gradients, slow convergence, and its inability to handle sparse data~\cite{conf/cvpr/ZouSJZL19,journals/corr/abs-1808-05671}. Consequently, it is common for clients to employ momentum-based optimizers such as Adam~\cite{journals/corr/KingmaB14} and SGD with momentum~\cite{conf/icml/SutskeverMDH13} to optimize their local models~\cite{journals/corr/abs-1901-09888,conf/www/HeLZNHC17,conf/sigir/Cao0MAYH18}. Our experimental findings in Appendix H.2 also substantiate this. Naturally, the following question arises:
\begin{quote}
\emph{Can we simply employ the model parameters of clients to detect Byzantine clients when momentum-based optimizers are used to learn the federated recommendation system?}
\end{quote}

\begin{figure}
  \centering
  \subfigure[Model parameters]{\includegraphics[width=0.3\linewidth]{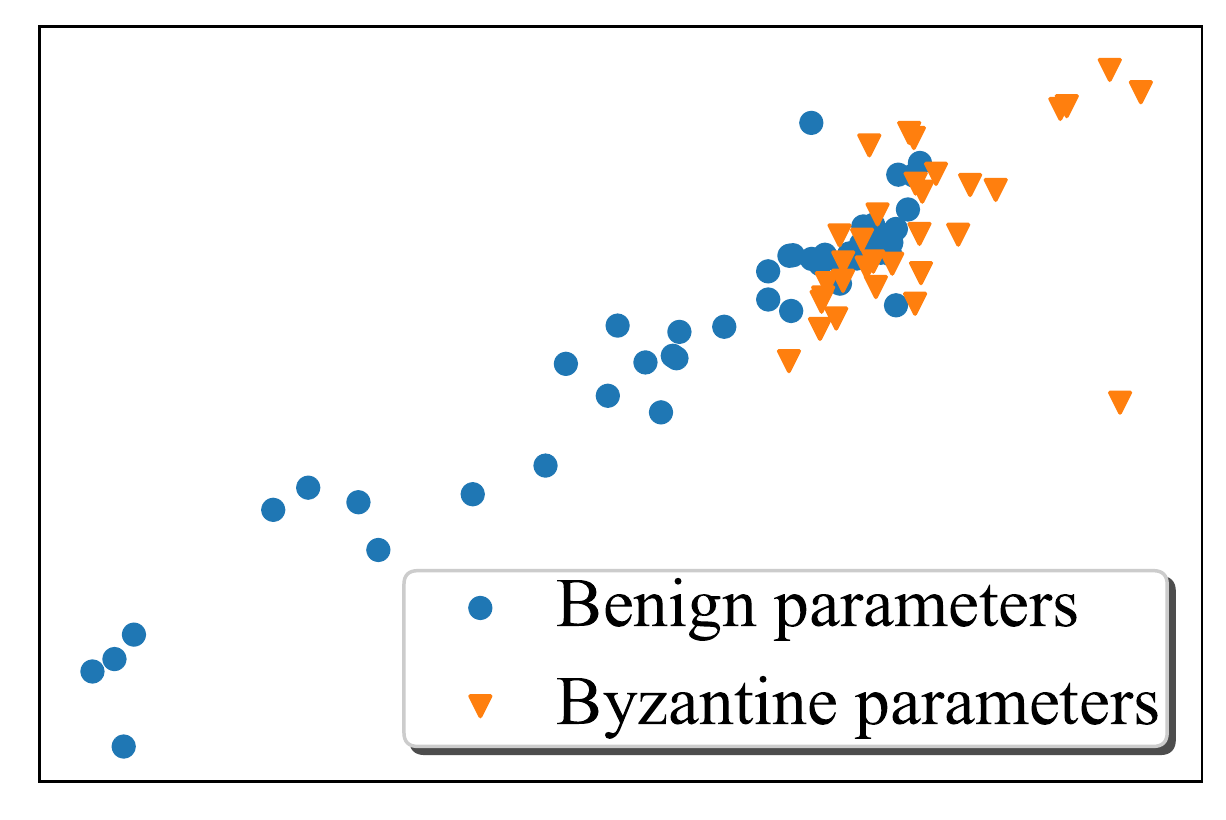}}
  \subfigure[Gradients]{\includegraphics[width=0.3\linewidth]{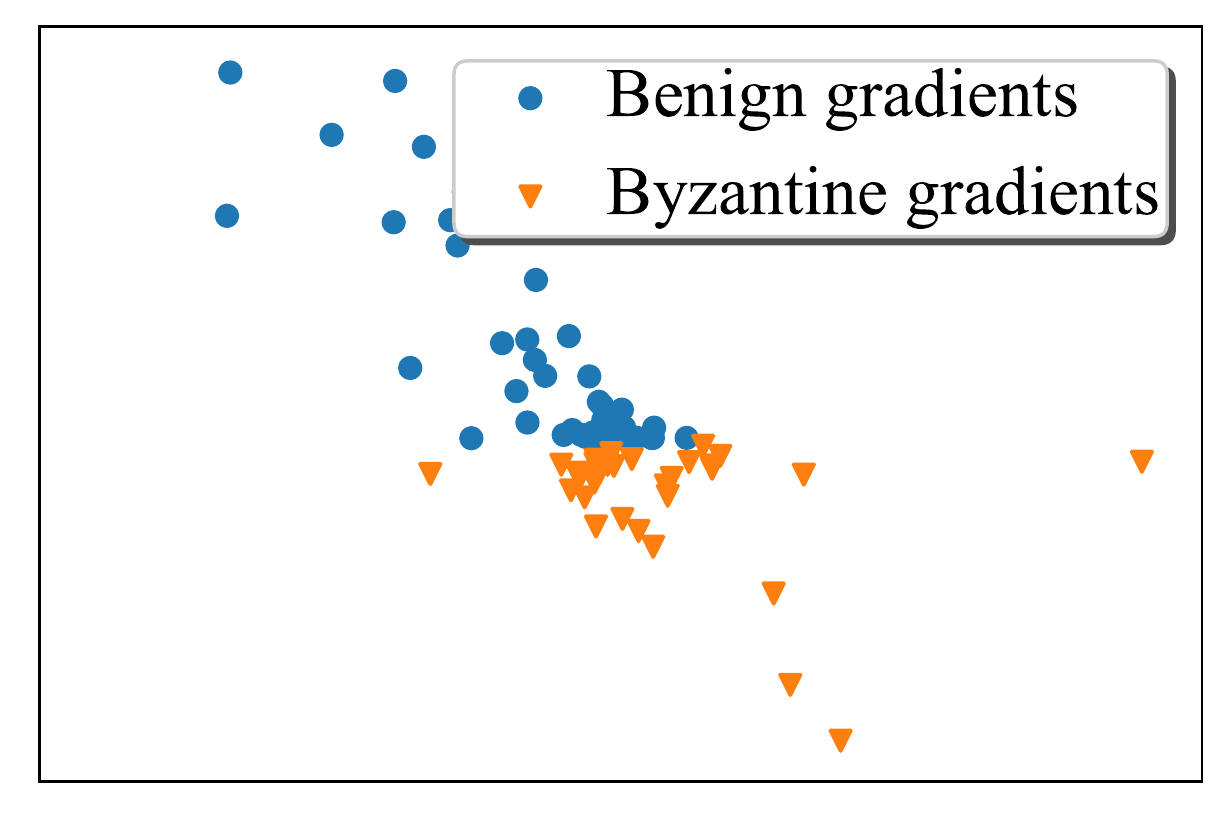}}
  \vspace{-2mm}
  \caption{Comparison between model parameters and gradients.
  58 benign clients (blue dots) and 32 Byzantine clients (orange triangles) federally use Adam to jointly optimize a matrix factorization model~\cite{journals/corr/abs-1901-09888}. We extract model parameters (left panel) and gradients (right panel) from all 90 clients at the $24$-th training round and use principal components analysis (PCA) to project them into 2-D dimension for visualization.}
  \label{fig:visualization}
  \vspace{-3mm}
\end{figure}

We argue that the answer to the above question is negative. We further propose that we should utilize gradients rather than model parameters to detect Byzantine clients. For example, when a client uses Adam to optimize its local model, the update to the model parameter is not identical to the gradient of the local model~\cite{journals/corr/KingmaB14}. Actually, the model parameter is adaptively updated based on the current gradient and gradients in the previous rounds.
If the client is Byzantine, its model gradient could largely deviate from that of benign clients (e.g., Byzantine clients perform gradient ascent while benign clients perform gradient descent \cite{conf/uai/XieKG19}). However, the model parameter of the Byzantine client could be similar to that of the benign client due to the fact that the model parameter accumulates gradients (including other benign gradients) of the previous training rounds.
As a result, Byzantine model parameters are less distinguishable from benign ones compared to Byzantine gradients.
To illustrate this fact, in Figure~\ref{fig:visualization}, we trained a simple federated recommendation system with 58 benign clients (blue dots) and 32 Byzantine clients (orange triangles) using Adam optimizer. We plotted the clients' model parameters (left panel) and the clients' gradients (right panel) at a randomly selected training round. From Figure~\ref{fig:visualization}, model parameters from Byzantine and benign clients are less distinguishable than the gradients.
This issue affects all momentum-based optimizers including Adam, and we will elaborate in Section~\ref{sec:rfrs}.

To learn a robust federated recommendation system, rather than using clients' model parameters, we propose to utilize gradients to detect Byzantine clients.
Our main contributions are:
\begin{enumerate}
\item We first employ factored item similarity model (FISM)~\cite{conf/kdd/KabburNK13} in learning a federated recommendation system (FRS), which achieves state-of-the-art performance. Our method is able to handle real-time personalization and sparse client feedback~\cite{journals/tkde/HeHSLJC18} better than the existing methods ~\cite{journals/corr/abs-1901-09888,journals/corr/abs-1906-05108,journals/corr/abs-2003-00602}.

\item To the best of our knowledge, we are the first to develop a robust federated recommendation system against the poisoning attack.
We show that when clients use momentum-based optimizers such as Adam, Byzantine clients are able to camouflage their model parameters and launch effective attacks. However if we employ gradients for detection, it can effectively thwart Byzantine clients.
We propose a new definition of Byzantine resilience catering to momentum-based optimized FRS, and we provide a theoretically guarantee that our robust learning strategy (gradient-based detection) is Byzantine resilient.
Empirically, we conduct extensive experiments on real-world datasets verifying the efficacy of our robust learning strategy against poisoning attacks.
Besides momentum-based optimizers, we further show that our robust learning strategy can be easily adapted to other well-known optimizers such as AdaGrad \cite{journals/jmlr/DuchiHS11} and RMSProp \cite{Tieleman2012} and still preserves the theoretical guarantee of Byzantine resilience.

\end{enumerate}

\section{Background and notations}
In this section, we review factored item similarity model (FISM) and Adam optimizer used in federated recommendation system (FRS).

\paragraph{Notations.} We use bold lower-case letters such as $\mathbf{m}$ to represent vectors, lower-case letters such as $m$ to represent scalars, and upper-case curlicue letters such as $\mathcal{S}$ to represent sets.
Aggregated vectors are denoted by a line over vectors such as $\overline{\mathbf{m}}$.
Byzantine vectors are denoted by a tilde over vectors such as $\widetilde{\mathbf{m}}$.
$\norm{\mathbf{m}}$ denotes the Euclidean norm of $\mathbf{m}$.
$\left|\mathcal{S}\right|$ is the cardinality of set $\mathcal{S}$.
$\odot$ denotes element-wise multiplication (Hadamard product). All operations between vectors are element-wise operations in this paper (except inner products of vectors).

\paragraph{Factored item similarity model (FISM).}
In online services (e.g., YouTube, Amazon, etc.),  clients constantly update their preference data~\cite{besbes2016optimization}. It is imperative for recommendation systems to handle real-time personalization~\cite{grbovic2018real}.
Moreover, many clients' activities are not frequent, i.e., the client-item rating matrix is sparse~\cite{popescul2013probabilistic}.
To deal with real-time personalization and sparse data, factored item similarity model (FISM) was proposed~\cite{conf/kdd/KabburNK13}.
FISM utilizes the average embedding vector of items that have been rated by client $i$ to represent client $i$'s feature and then uses the inner product of client $i$'s and item $j$'s embedding vectors to calculate the prediction score of client $i$ on item $j$.
Formally, the predictive model is:
\begin{equation}\label{item_based_CF}
\begin{split}
\hat{y}^{ij}=\underbrace{\Bigg(\mathbf{p}^j\Bigg)^\top}_{\text{item $j$'s embedding vector}}\underbrace{\Bigg(\frac{1}{|\mathcal{S}^{i+}\backslash\{j\}|^\gamma}\sum_{k\in \mathcal{S}^{i+}\backslash\{j\}}\mathbf{q}^k\Bigg)}_{\text{client $i$'s embedding vector}}.
\end{split}
\end{equation}
$\hat{y}^{ij}$ is the prediction score of client $i$ on item $j$. $\mathcal{S}^{i+}$ is the set of client $i$'s rated items.
$\gamma$ is a hyperparameter controlling the normalization effect.
$\mathbf{p}^j,\mathbf{q}^k\in\mathbb{R}^d$ are the trainable embedding vectors for item $j$ and $k$ respectively. $d$ is the dimension of the embedding vectors. Each item $j$ has two embedding vectors $\mathbf{p}^j$ and $\mathbf{q}^j$: $\mathbf{p}^j$ represents the item feature for prediction, while $\mathbf{q}^j$ is a historical interaction used for representing client feature. For clarity, we use $\boldsymbol{\theta}$ to represent trainable embedding vectors $\mathbf{p}^j$ and $\mathbf{q}^j$ for all items $j$.

When client $i$ interacts with a new item $j$, we just need to add item $j$ to $\mathcal{S}^{i+}$, i.e., $\mathcal{S}^{i+}=\mathcal{S}^{i+}\cup \{j\}$, and do not need to retrain the model.
On the other hand, even if a client has less activity, it can be represented by an average embedding vector of items, which prevents overfitting of the model.
\paragraph{Adam optimizer.}
Adam optimizer has been widely used in learning FRS, due to its fast convergence property and its ability to handle sparse gradients~\cite{journals/corr/abs-1901-09888}. 
Different from SGD that directly uses gradients to update model parameters, Adam updates the model parameters using estimates of first and second moments of the gradients.
At round $t$, client $i$ uses Adam optimizer to update its model parameter $\boldsymbol{\theta}^{ti}$ according to:
\begin{align}
\mathbf{m}^{ti}&=\beta_1 \overline{\mathbf{m}}^{t-1}+(1-\beta_1) \mathbf{g}^{ti} \label{eq:adam_m}\\
\mathbf{v}^{ti}&=\beta_2 \overline{\mathbf{v}}^{t-1}+(1-\beta_2) \mathbf{g}^{ti}\odot \mathbf{g}^{ti} \label{eq:adam_v}\\
\eta^t &=\eta  \frac{\sqrt{1-\beta_2^t}}{1-\beta_1^t} \label{eq:adam_eta}\\
\mathbf{u}^{ti}&= \frac{\mathbf{m}^{ti}}{\sqrt{\mathbf{v}^{ti}}+\mathbf{\epsilon}} \label{eq:adam_u}\\
\boldsymbol{\theta}^{ti}&=\overline{\boldsymbol{\theta}}^{t-1}-\eta^t\mathbf{u}^{ti}, \label{eq:adam_theta}
\end{align}
where $\mathbf{g}^{ti}$, $\mathbf{m}^{ti}$, $\mathbf{v}^{ti}$, $\mathbf{u}^{ti}$ and $\boldsymbol{\theta}^{ti}$ are gradient, first moment, second moment, update and model parameter of the $i$-th client at $t$-th (communication) round. $\overline{\mathbf{m}}^{t-1}$, $\overline{\mathbf{v}}^{t-1}$ and  $\overline{\boldsymbol{\theta}}^{t-1}$ that are the inputs at round $t$ respectively represent aggregated first moment, aggregated second moment and aggregated model parameter at ($t-1$)-th round. $\beta_1$ and $\beta_2$ are hyperparameters related to first and second moment vectors. $\eta$ is the learning rate. $\overline{\mathbf{m}}^0$, $\overline{\mathbf{v}}^0$ are initialized to $\mathbf{0}$. $\overline{\boldsymbol{\theta}}^0$ is initialized with random values sampled from a standard Gaussian distribution.  $\mathbf{\epsilon}$ is a small constant for numerical stability.

\section{Federated recommendation system}
\label{sec:frs}
In this section, we employ factored item similarity model (FISM)~\cite{conf/kdd/KabburNK13} in federated recommendation system (FRS), namely, Adam-based federated recommendation system (A-FRS). Then, we introduce the training process for A-FRS (Algorithm \ref{alg:rfrs}).

Conventional federated recommendation systems based on matrix factorization (MF) \cite{journals/corr/abs-1901-09888,journals/corr/abs-1906-05108,journals/corr/abs-2003-00602} fail to deal with real-time personalization and sparse data \cite{conf/kdd/KabburNK13,rendle2010factorization,conf/www/HeLZNHC17}.
To solve this problem, we propose Adam-based federated recommendation system (A-FRS), which extends FISM to be a FRS. A-FRS inherits the advantages of FISM \cite{conf/kdd/KabburNK13}. It can handle real-time personalization and deal with sparse data with the ability to learn a quality recommendation model without collecting clients' data.

The learning of A-FRS is shown in Algorithm \ref{alg:rfrs} (Option $\mathbf{I}$).
In Algorithm \ref{alg:rfrs}, A-FRS consists of two parts: a server and $|\mathcal{S}|$ clients. The server distributes first moment, second moment, and model parameters to the clients for training and aggregates those trained by all clients.

\begin{algorithm}[!t]
    \caption{\underline{A}dam-based (\underline{R}obust) \underline{F}ederated \underline{R}ecommendation \underline{S}ystem -  A-(R)FRS}
    \label{alg:rfrs}
    \textbf{Input:} Client set $S$, number of training rounds $T$, and fraction of training clients per round $e$ \\
    \textbf{Output:} Trained model parameter $\overline{\boldsymbol{\theta}}^T$ (Represents the trainable parameter of Eq.~(1))
    \begin{algorithmic} 
        \Procedure{Server Aggregation}{}
            \State Initialize $\overline{\mathbf{m}}^{0}$,  $\overline{\mathbf{v}}^{0}$, and  $\overline{\boldsymbol{\theta}}^{0}$
            \For{each round $t$ = 1, 2, ..., $T$}
                \State $\mathcal{S}^t\gets$ random subset of $\mathcal{S}$
                \Comment{$\left|\mathcal{S}^t\right|=e*\left|S\right|$.}
                \For{each client $i \in \mathcal{S}^t$} {\textbf{in parallel}}
                    \State $\mathbf{m}^{ti},\mathbf{v}^{ti},\boldsymbol{\theta}^{ti} \gets$ ClientUpdate$(\overline{\mathbf{m}}^{t-1},\overline{\mathbf{v}}^{t-1},\overline{\boldsymbol{\theta}}^{t-1})$
                    \Comment{Algorithm \ref{alg:client}.}
                    \State $\mathbf{g}^{ti}\gets$ Compute with $\mathbf{m}^{ti}$ and $\overline{\mathbf{m}}^{t-1}$ \Comment{Use Eq. (\ref{eq:adam_m}) for Byzantine clients detection.}
                \EndFor
                \State Option $\mathbf{I}$:
                $\mathcal{F}_t\gets \mathcal{S}^t$
                \Comment{Non-robust FRS. Ignore the computed $\mathbf{g}^{ti}$.}
                \State Option $\mathbf{II}$:
                $\mathcal{F}_t\gets  F(\mathbf{g}^{t1},...,\mathbf{g}^{t{\left|\mathcal{S}^t\right|}})$
                \Comment{Robust FRS. Use the computed $\mathbf{g}^{ti}$.}
                \State $N^t=\sum_{i\in \mathcal{F}_t} n^i$
                \Comment{$n^i$ is the number of training data of client $i$.}
                \State $\overline{\mathbf{m}}^t \gets \sum_{i\in \mathcal{F}_t}\frac{n^i}{N^t}\mathbf{m}^{ti}$
                \State $\overline{\mathbf{v}}^t \gets \sum_{i\in \mathcal{F}_t}\frac{n^i}{N^t}\mathbf{v}^{ti}$
                \State $\overline{\boldsymbol{\theta}}^t \gets \sum_{i\in \mathcal{F}_t}\frac{n^i}{N^t}\boldsymbol{\theta}^{ti}$
            \EndFor
        \EndProcedure
    \end{algorithmic}
\end{algorithm}

\begin{algorithm}[!t]
    \caption{Client update using Adam optimizer}
    \label{alg:client}
    \textbf{Input:} Aggregated first moment $\overline{\mathbf{m}}^{t-1}$, aggregated second moment $\overline{\mathbf{v}}^{t-1}$ and aggregated model parameter $\overline{\boldsymbol{\theta}}^{t-1}$ at round ($t-1$) from the server\\
    \textbf{Output:} First moment $\mathbf{m}^{ti}$, second moment $\mathbf{v}^{ti}$ and model parameter $\boldsymbol{\theta}^{ti}$ of client $i$ at round $t$
    \begin{algorithmic}
        \Procedure{Client Update}{}
        \Comment{Run on the $i$-th client.}
            \State $\mathbf{g}^{ti} \gets \nabla\ell(\overline{\boldsymbol{\theta}}^{t-1};D_i)$
            \Comment{Compute gradient by Eq. (\ref{eq:loss}) using local dataset $D_i$.}
            \State $\mathbf{m}^{ti}, \mathbf{v}^{ti}, \boldsymbol{\theta}^{ti} \gets$ AdamUpdate$(\mathbf{g}^{ti}, \overline{\mathbf{m}}^{t-1}, \overline{\mathbf{v}}^{t-1})$
            \Comment{Apply Eq. (\ref{eq:adam_m}-\ref{eq:adam_theta}).}
        \EndProcedure
    \end{algorithmic}
\end{algorithm}

Clients optimize their local recommendation model based on the ranking loss ~\cite{conf/uai/RendleFGS09} defined as follows:
\begin{equation}\label{eq:loss}
\begin{split}
\ell =-\sum_{i \in \mathcal{S}} \sum_{j \in \mathcal{S}^{i+}} \sum_{k \in \mathcal{S}^{i-}} \log\sigma(\hat{y}^{ij}-\hat{y}^{ik})+\lambda\norm{\boldsymbol{\theta}},
\end{split}
\end{equation}
where $\mathcal{S}$ is the client set. $\sigma(\cdot)$ is the sigmoid function.
$\boldsymbol{\theta}$ denotes trainable model parameter. $\lambda$ controls the strength of $L_2$ regularization to prevent overfitting.
$\mathcal{S}^{i+}$ and $\mathcal{S}^{i-}$ are sets of client $i$'s rated items and unrated items respectively. $\hat{y}^{ij}$ and $\hat{y}^{ik}$ can be calculated with Eq. (\ref{item_based_CF}). Similar to other federated recommendation methods~\cite{journals/corr/abs-1901-09888}, clients utilize Adam to optimize model parameters. Compared with SGD, two additional
terms (the first moment $\mathbf{m}$ and the second moment $\mathbf{v}$) are introduced in Adam. Thus, clients also need to transfer these two terms to the server for aggregation.

\section{Robust federated recommendation system}
\label{sec:rfrs}
In this section, we first show that Byzantine clients can camouflage the model parameters and launch effective attacks. Then, we propose our robust learning strategy in Adam-based FRS and theoretically prove that our strategy is Byzantine resilient. Besides, in Appendix C, D, and E, we show that our robust learning strategy is also suitable in FRS based on other well-known optimizers (e.g., SGD with momentum, AdaGrad, and RMSProp) with theoretical guarantees.

\subsection{Motivation - Byzantine clients can camouflage model parameters}
\label{sec:camouflage}
We demonstrate that when clients use momentum-based optimizers such as Adam, the Byzantine clients can camouflage their model parameters while launching an effective poisoning attack\footnote{Byzantine clients will obey the Adam update rules (Eq.~(\ref{eq:adam_m}-\ref{eq:adam_theta})).
Otherwise, the server can easily detect the abnormal behavior by verifying the Adam update rules.}. Thus, the existing defense methods~\cite{conf/nips/BlanchardMGS17,conf/icml/DamaskinosMGPT18,conf/icml/MhamdiGR18} catering to SGD optimizer can not be simply applied, because they detect Byzantine clients based on the model parameters.

\begin{figure}[tp!]
  \centering
  \includegraphics[width=0.35\linewidth]{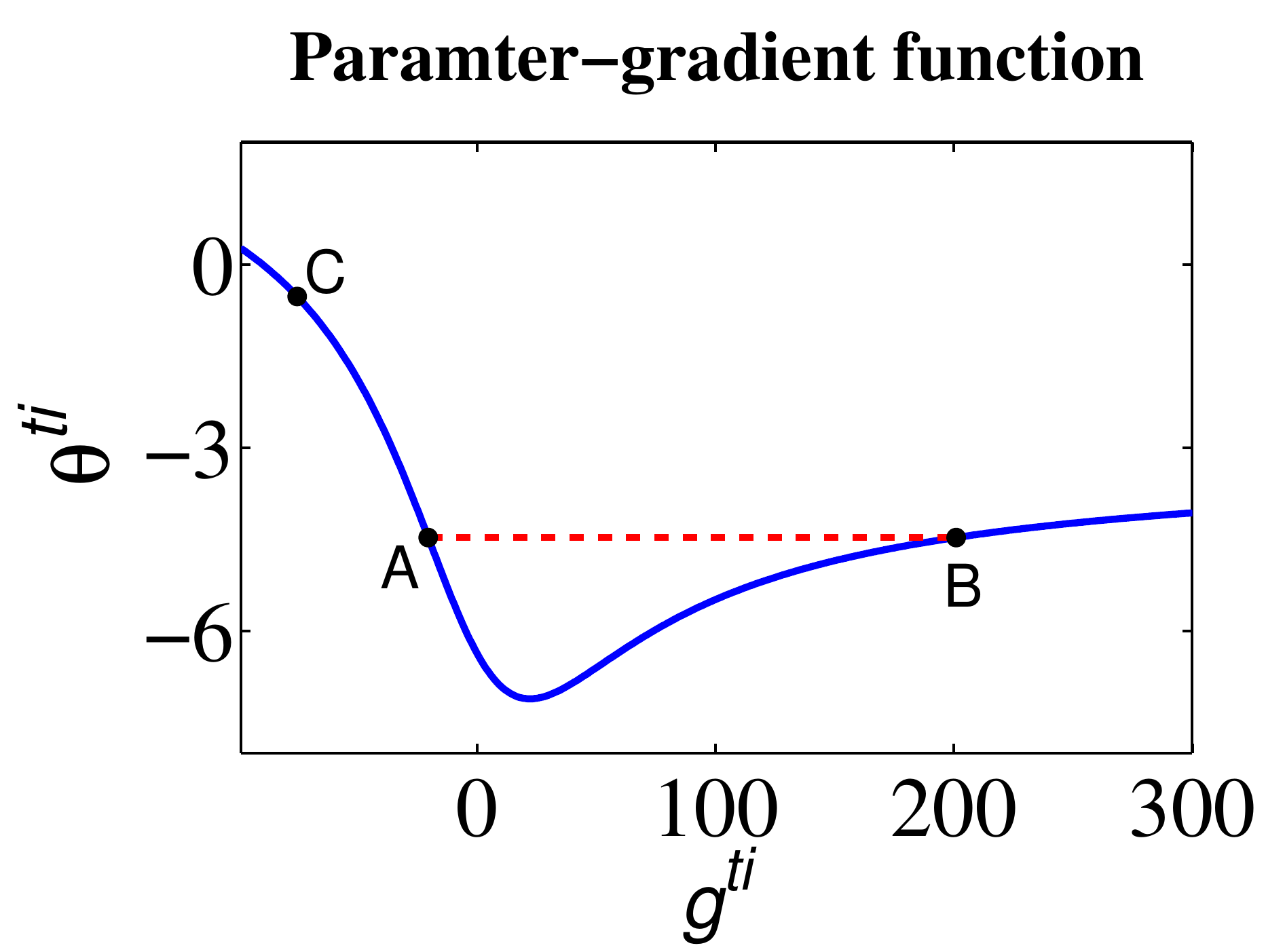}\label{fig:theta_g}
  \hspace{4mm}
  \includegraphics[width=0.35\linewidth]{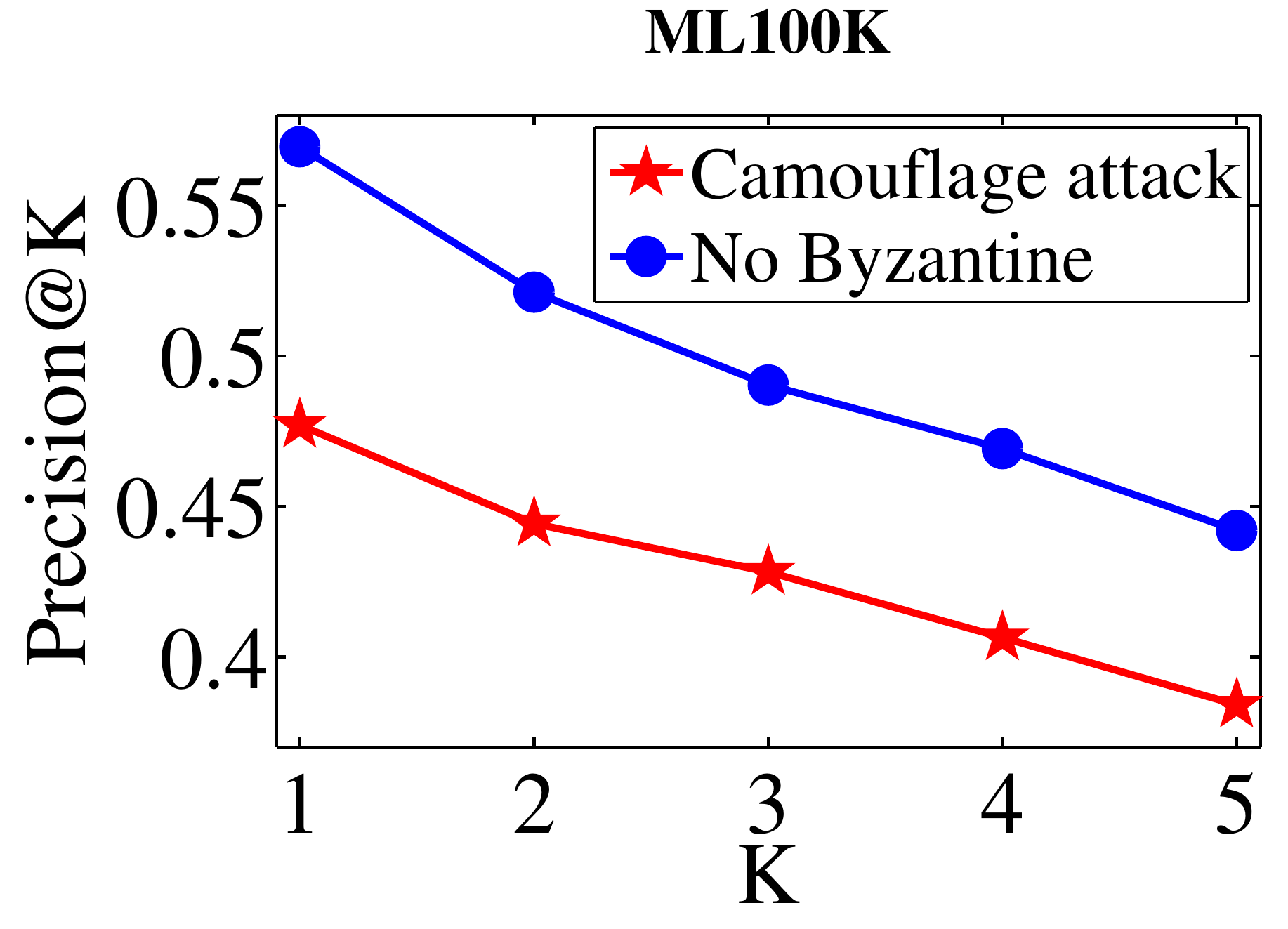}\label{fig:camouflage_result}
  \vspace{-2mm}
  \caption{Left panel shows the relation between the model parameter ${\theta}^{ti}$ and the gradient $g^{ti}$ of client $i$ at round $t$ according to Eq.~(\ref{eq:adam_m}-\ref{eq:adam_theta}). Right panel shows the Byzantine clients can launch the effective poisoning attack while keep model parameters the same as those of benign ones.}
  \vspace{-3mm}
  \label{fig:para-gradient-camouflage_attack}
\end{figure}

In the left panel of Figure~\ref{fig:para-gradient-camouflage_attack}, the $i$-th client at the round $t$ uses Adam to optimize the local model. We plot the parameter $\theta^{ti}$-gradient $g^{ti}$ relationship.
For simplicity, let $\theta^{ti}$ and $g^{ti}$ be scalars.
Suppose point A is the benign point with benign $\theta^{ti}$ and $g^{ti}$, the Byzantine client can choose Byzantine point B which has the same $\theta^{ti}$ but completely different $\widetilde{g}^{ti}$ that is calculated by
\begin{align}
\label{eq:relation-parameter-update}
    \widetilde{g}^{ti}=\frac{2\beta_1\beta_2\left(1-\beta_1\right)\overline{m}^{t-1}\overline{v}^{t-1}
+\beta_2\left(1-\beta_1\right)^2 \overline{v}^{t-1} g^{ti}
 -\beta_1^2\left(1-\beta_2\right)(\overline{m}^{t-1})^2 g^{ti}}
 {\beta_1^2\left(1-\beta_2\right)(\overline{m}^{t-1})^2
 +2\beta_1\left(1-\beta_1\right)\left(1-\beta_2\right)\overline{m}^{t-1}g^{ti}
-\beta_2\left(1-\beta_1\right)^2\overline{v}^{t-1}}.
\end{align}
The detailed derivation is provided in Appendix F. Eq.~(\ref{eq:relation-parameter-update}) is the case for one dimension. It can be easily generalized into higher-dimensional model parameters due to element-wise operations of Eq.~(\ref{eq:adam_m}-\ref{eq:adam_theta}).
In particular, when the dimension of the model parameter $\boldsymbol{\theta}^{ti}$ is large, the number of such Byzantine gradients $\widetilde{\mathbf{g}}^{ti}$ is exponential, because the Byzantine client can craft each component of the Byzantine gradients $\widetilde{\mathbf{g}}^{ti}$.
Thus, the Byzantine client can easily choose a $\widetilde{\mathbf{g}}^{ti}$ that can effectively poison the global model.

In the right panel of Figure~\ref{fig:para-gradient-camouflage_attack}, we conduct an experiment showing the efficacy of the camouflage attack (red line). The blue line represents Precision@$K$ ~\cite{wu2016collaborative} of A-FRS (Option $\mathbf{I}$ in Algorithm~\ref{alg:rfrs}) without any attack.
The red line represents the Precision@$K$ with $40\%$ Byzantine clients. The Byzantine clients maintain the same model parameters for update but calculate the Byzantine $\widetilde{\mathbf{g}}^{ti}$ that has the largest Euclidean distance from the benign correspondence $\mathbf{g}^{ti}$. Consequently, from the right panel of  Figure~\ref{fig:para-gradient-camouflage_attack}, the red line is lower than the blue line. It shows that the camouflage attack can effectively degrade the global model.


It is worth noting that not all model parameters can be camouflaged.
For example, in the left panel of Figure~\ref{fig:para-gradient-camouflage_attack}, point C has a unique mapping from $\theta^{ti}$ to $g^{ti}$. Thus, the Byzantine client has a certain chance of failing to camouflage the model parameter. However, this chance is afflicted with the curse of dimensionality. As the dimensionality of model parameters is typically very large, the Byzantine client can confidently camouflage the model parameter and launch effective attacks.

Motivated by the above observations, rather than using model parameters, we propose to use gradients of the models to detect Byzantine clients in the following sections.
\subsection{Byzantine resilience}
The current defense methods employ the existing definitions of Byzantine resilience \cite{conf/nips/BlanchardMGS17,conf/icml/MhamdiGR18,xie2018phocas}. However, these definitions only provide restrictions to the model parameters of clients. For example, if a client's model parameter is very distinct from those of others, this client is deemed to be Byzantine. These restrictions do not apply to first moment and second moment in Algorithm \ref{alg:client}. As a result, Byzantine clients can camouflage the model parameters but change the gradient along with first and second moments to be very different (shown in Section \ref{sec:camouflage}). Thus, existing definitions have this serious limitation, which can lead to a security breach.


To solve this, we give a new definition, \textit{Adam-Byzantine resilience} to evaluate defense methods when Adam is used in FRS. Other types of Byzantine resilience definitions catering to other optimizers such as SGD with momentum, AdaGrad, and RMSProp can be found in Appendix C, D, and E.

Suppose $\widetilde{n}$ out of $n$ clients are Byzantine.
Let $G^t=\{\mathbf{g}^{ti}|i\in\{1,...,n-\widetilde{n}\}\}$ be the gradient set of $n-\widetilde{n}$ benign clients at round $t$. Let  $\widetilde{G}^t=\{\widetilde{\mathbf{g}}^{ti}|i\in\{1,...,\widetilde{n}\}\}$ be the gradient set of $\widetilde{n}$ Byzantine clients at round $t$.
Let $\mathcal{F}_t$ be the set of selected clients for aggregation.
Let $\hat{G}^t=\{\hat{\mathbf{g}}^{ti}|\text{client} \, i\in\mathcal{F}_t\}$ be the gradient set of clients in $\mathcal{F}_t$.
We define Adam-Byzantine resilience as follows:

\textbf{Definition 1} Adam-Byzantine Resilience.
For any client $i$ in $\mathcal{F}_t$ at training round $t$, we denote its first moment, second moment and model parameter as $\hat{\mathbf{m}}^{ti}$, $\hat{\mathbf{v}}^{ti}$ and $\hat{\boldsymbol{\theta}}^{ti}$.  For any benign client $j$ at training round $t$, we denote its first moment, second moment, and model parameter as $\mathbf{m}^{tj}$, $\mathbf{v}^{tj}$ and $\boldsymbol{\theta}^{tj}$.
A defense method is Adam-Byzantine resilient, if for the round $T$ there exist positive constant numbers $C_m$, $C_v$ and $C_\theta$, such that
\begin{enumerate}
    \item $\sum\limits_{t=1}^T\sum\limits_{\text{client} \, i \in \mathcal{F}_t}\sum\limits_{\text{benign client} \, j}\norm{\hat{\mathbf{m}}^{ti}-\mathbf{m}^{tj}}\leq C_m$;
    \item $\sum\limits_{t=1}^T\sum\limits_{\text{client} \, i \in \mathcal{F}_t}\sum\limits_{\text{benign client} \, j}\norm{\hat{\mathbf{v}}^{ti}-\mathbf{v}^{tj}}\leq C_v$;
    \item $\sum\limits_{t=1}^T\sum\limits_{\text{client} \, i \in \mathcal{F}_t}\sum\limits_{\text{benign client} \, j}\norm{\hat{\boldsymbol{\theta}}^{ti}-\boldsymbol{\theta}^{tj}}\leq C_\theta$.
\end{enumerate}
The definition of Adam-Byzantine resilience provides constraints on first moments, second moments, and model parameters.
If a defense method is Adam-Byzantine resilient, no matter how Byzantine clients attack the server (e.g., by using gradient ascent~\cite{conf/nips/BlanchardMGS17} or by adding random noise~\cite{journals/corr/abs-2002-00211}), the attack will have little influence on the global model, which guarantees the efficacy of the defense method.
In comparison, the existing definitions of Byzantine resilience~\cite{conf/nips/BlanchardMGS17,conf/icml/DamaskinosMGPT18,conf/icml/MhamdiGR18} can provide no guarantees on condition 1 and condition 2.

\subsection{Adam-based robust federated recommendation system (A-RFRS)}
When clients federally learn a recommendation system using Adam optimizer, we propose Adam-based robust federated recommendation system (A-RFRS) (Option $\mathbf{II}$ in Algorithm \ref{alg:rfrs}).
In A-RFRS, $\mathbf{g}^{ti}$ are used to detect Byzantine clients, since $\mathbf{m}^{ti}$, $\mathbf{v}^{ti}$, and $\boldsymbol{\theta}^{ti}$ are computed by $\mathbf{g}^{ti}$ of client $i$ at round $t$. Compared with non-robust A-FRS (Option $\mathbf{I}$), our robust version A-RFRS performs a filtering operation before the aggregation of updates from clients. The filtering function $F(\cdot)$ is flexible. In our paper, we utilize the strategy of Krum \cite{conf/nips/BlanchardMGS17} to filter out Byzantine clients based on the computed gradients.

To theoretically justify our proposed A-RFRS, we prove that A-RFRS is Adam-Byzantine resilient.

\textbf{Assumption 1}. For any gradient $\mathbf{g}$, its norm is upper bounded by a positive constant number $g_{max}$.
Formally, $\norm{\mathbf{g}}\leq g_{max}$ with $\mathbf{g}\in G^t\cup \widetilde{G}^t, t\in \mathbb{N}^*$.

\textbf{Assumption 2}. After $T'$ rounds of training, each component of $\overline{\mathbf{v}}^{t-1}$ is lower bounded by a positive constant number $v_{min}$. Formally, for any round $t$ with $t>T'$, $\overline{v}^{t-1}_k\geq v_{min}$, where $\overline{v}^{t-1}_k$ denotes the $k$-th component of $\overline{\mathbf{v}}^{t-1}$.

\textbf{Theorem 1}. A-RFRS is Adam-Byzantine resilient, if Assumption 1 and Assumption 2 hold, and for any client $i$ in $\mathcal{F}_t$ with gradient $\hat{\mathbf{g}}^{ti} \in \hat{G}^t$, for any benign client $j$ with gradient $\mathbf{g}^{tj}\in G^t$, and for training round $T\in\mathbb{N}^*$, there exist a positive constant number $C_g$, such that
$$\sum\limits_{t=1}^T\sum\limits_{\text{client} \, i \in \mathcal{F}_t}\sum\limits_{\text{benign client} \, j}\norm{\hat{\mathbf{g}}^{ti}-\mathbf{g}^{tj}}\leq C_g.$$

The proof of Theorem 1 is in Appendix B. Theorem 1 claims that, if our gradient-based filter algorithm (e.g., gradient-based Krum) guarantees that $\hat{\mathbf{g}}^{ti}$ is close to any benign gradient $\mathbf{g}^{tj}$, then the first moment $\hat{\mathbf{m}}^{ti}$, second moment $\hat{\mathbf{v}}^{ti}$ and model parameter $\hat{\boldsymbol{\theta}}^{ti}$ will also be close to the benign $\mathbf{m}^{tj}$, $\mathbf{v}^{tj}$ and $\boldsymbol{\theta}^{tj}$. This indicates that our defense method is robust to Byzantine attacks. This further shows that our learning strategy, which utilizes gradients to filter out Byzantine clients, is effective when Adam optimizer is used.


We also show that our proposed robust learning strategy that utilizes gradients to detect Byzantine clients, can be adapted to other well-known optimizers such as SGD with momentum, AdaGrad, and RMSProp with theoretical guarantees.
Those results and the proofs are in Appendix C, D, and E, respectively. We also conduct experiments comparing our robust learning strategy with existing defense methods on FRS based on SGD with momentum and AdaGrad in Appendix H.3.

\begin{figure*}[!tp]
	\centering
	\subfigure{
		\includegraphics[width=0.22\linewidth]{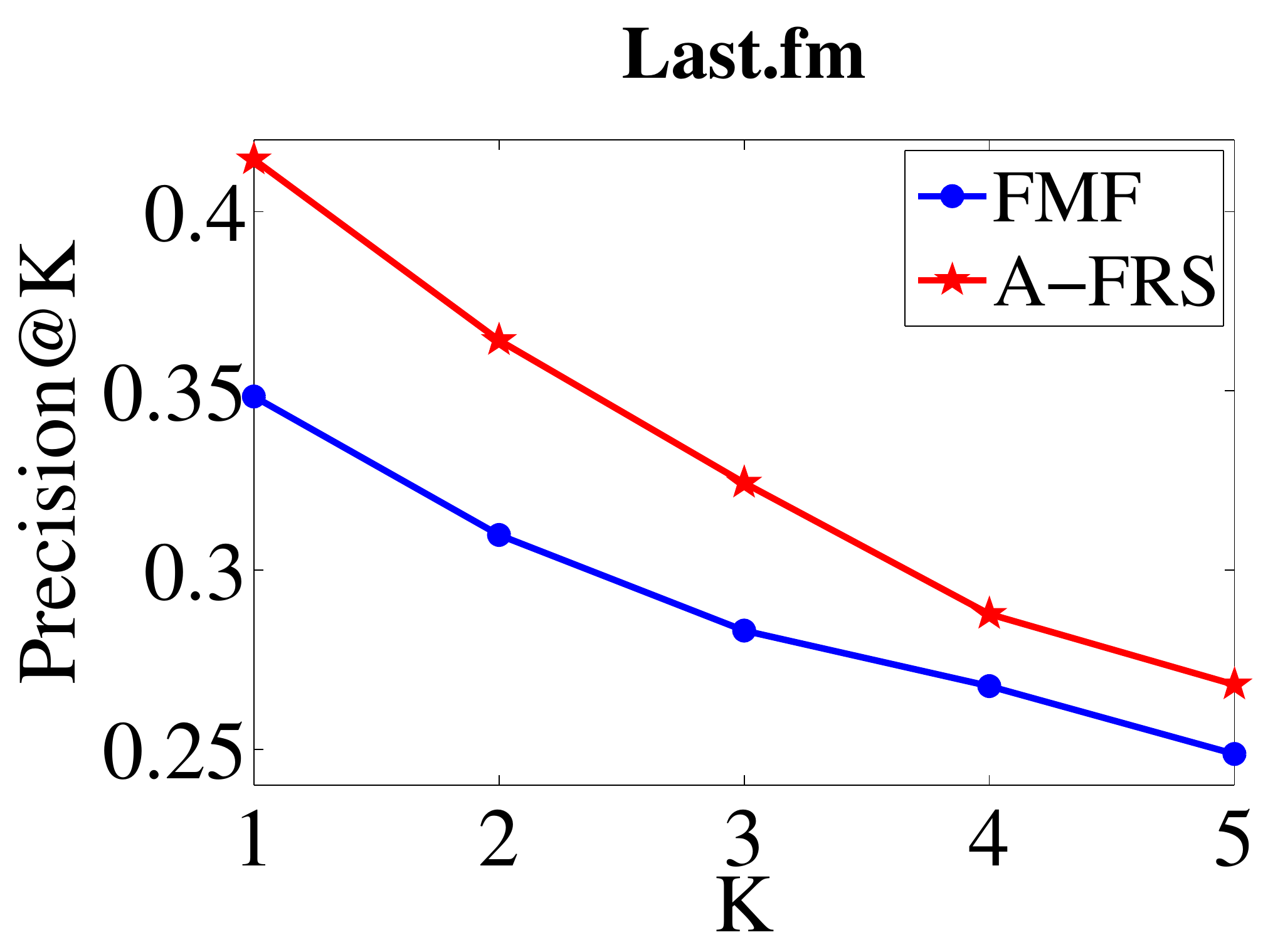}
	}
	\subfigure{
		\includegraphics[width=0.22\linewidth]{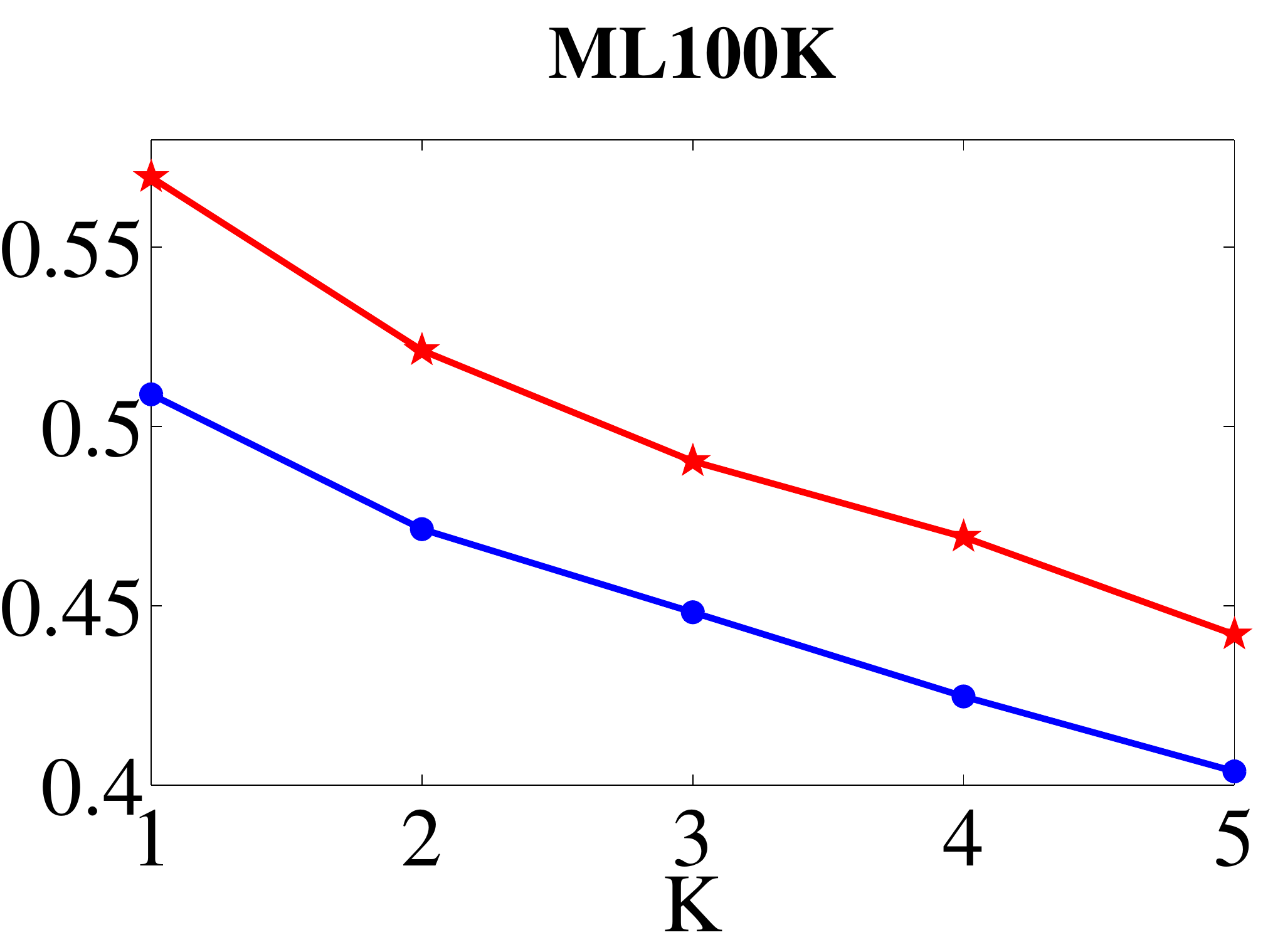}
	}
	\subfigure{
		\includegraphics[width=0.22\linewidth]{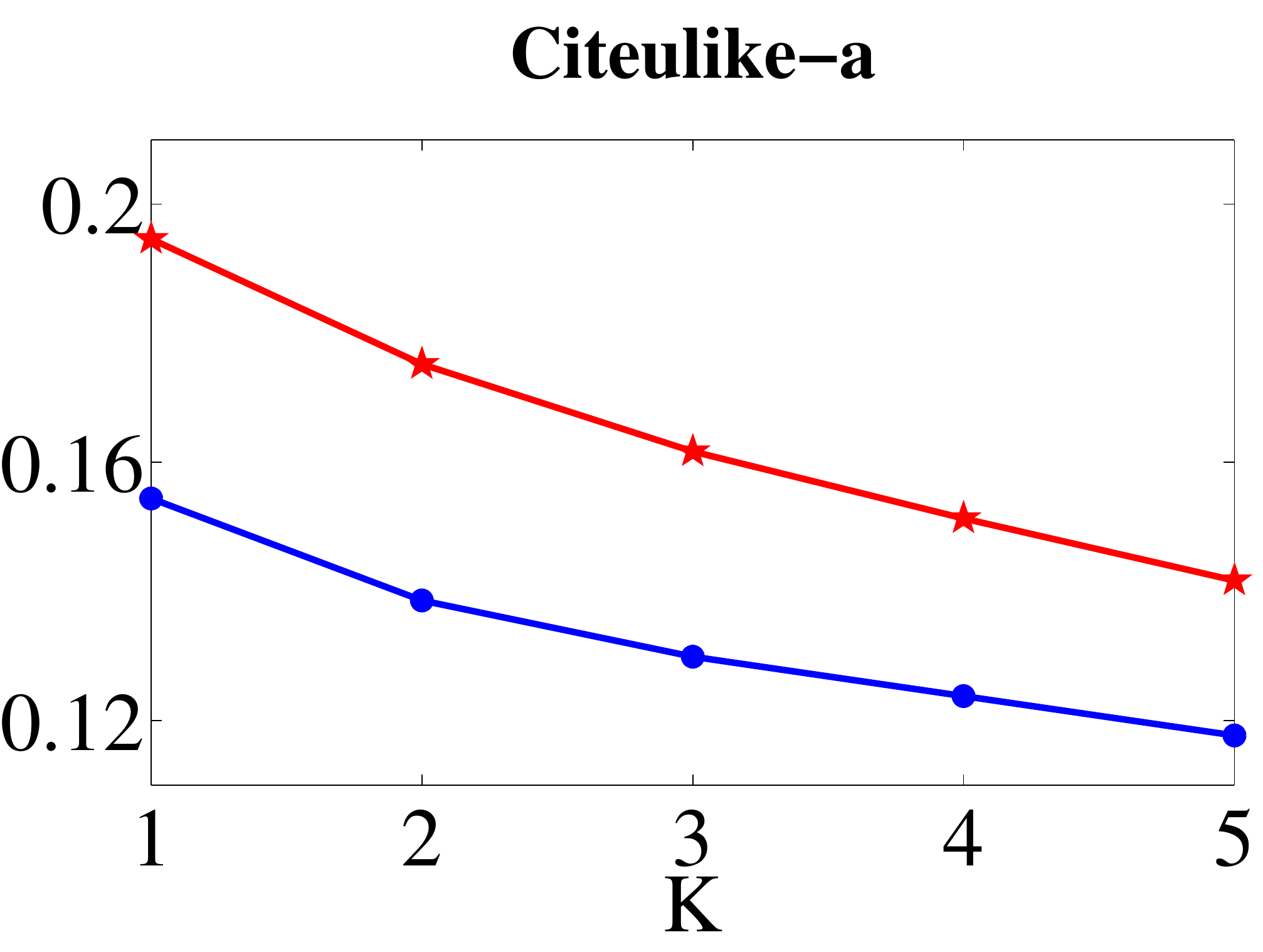}
	}
	\subfigure{
		\includegraphics[width=0.22\linewidth]{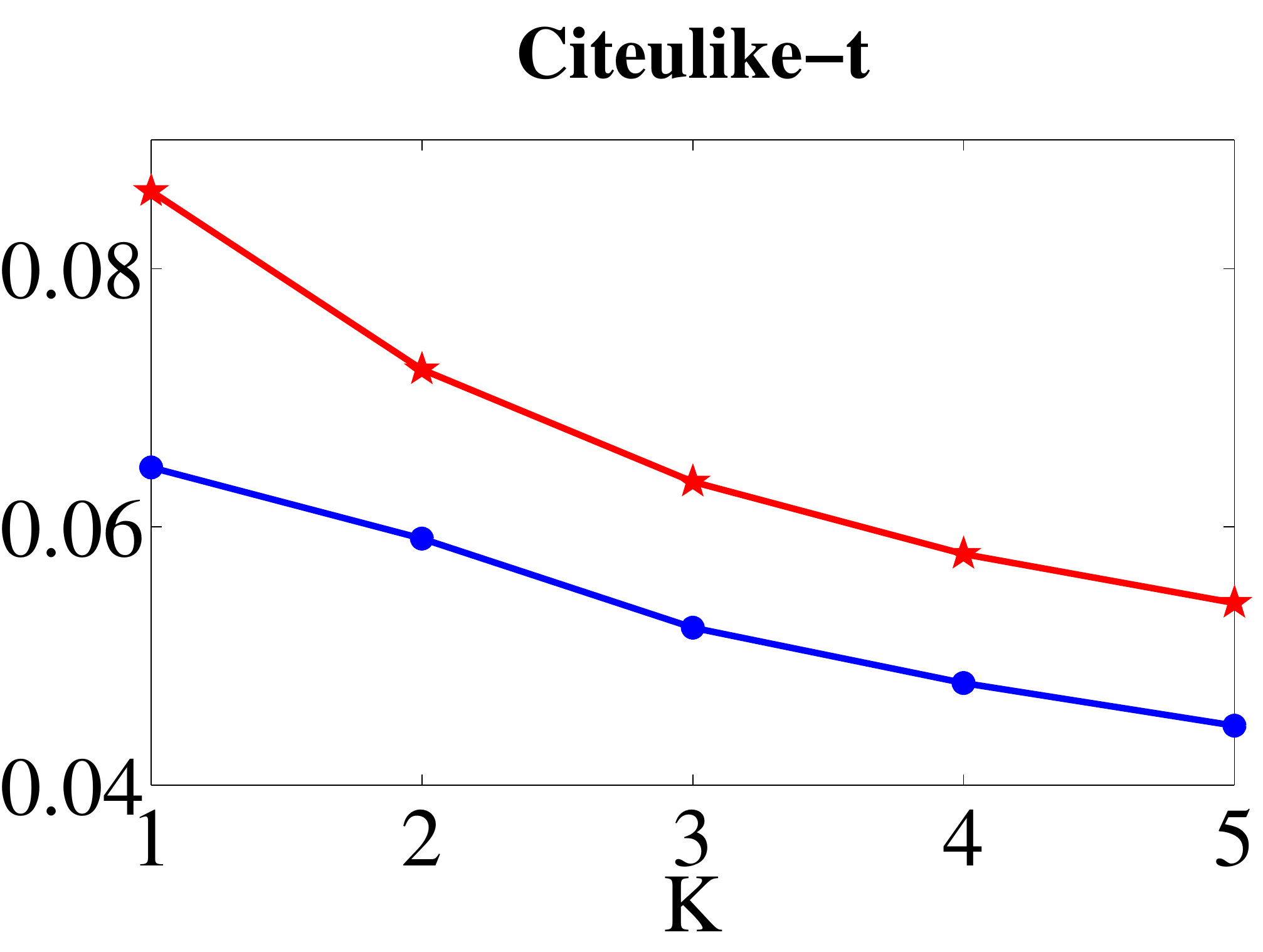}
	}
	\caption{Precision@$K$ of FRSs on 4 datasets. A-FRS (red line) is our proposed federated recommendation method.}
	\label{fig:performance1}
\end{figure*}

\section{Experiments}
In this section, we evaluate the performance of our proposed A-FRS (Option $\mathbf{I}$ in Algorithm~\ref{alg:rfrs}) and A-RFRS (Option $\mathbf{II}$ in Algorithm~\ref{alg:rfrs}) on 4 real-world datasets from various domains (i.e., Last.fm \cite{Cantador_RecSys2011}, ML100K \cite{harper2015movielens}, Citeulike-a \cite{conf/kdd/WangB11}, and Citeulike-t \cite{conf/ijcai/WangCL13}).

Our experiments consist of two parts: firstly, we show the efficacy of our A-FRS; secondly, we demonstrate A-RFRS can outperform other defense methods when clients use Adam optimizer to optimize recommendation models. The evaluation metric are Precision@$K$~\cite{wu2016collaborative} and Recall@$K$~\cite{ma2019learning} with the ranking position $K$ ranges from $1$ to $5$.
The detailed description of datasets and training configurations are in Appendix G.


\paragraph{Federated recommendation system.} In the first part of our experiments, we compare our A-FRS (Option $\mathbf{I}$ in Algorithm~\ref{alg:rfrs}) with existing federated matrix factorization (FMF) \cite{journals/corr/abs-1901-09888}, which employs matrix factorization in FRS.
For each client, we randomly select 80\% of its local data as the training set and evaluate the global model with the remaining 20\%.

Figure~\ref{fig:performance1} shows Precision@$K$ of FRSs on different datasets where the ranking position $K$ ranges from $1$ to $5$. We also report Recall@$K$~\cite{ma2019learning} in Appendix H.1. The results demonstrate that our proposed A-FRS (red line) outperforms FMF (blue line) on all datasets.
Recommendation datasets are often sparse, and our A-FRS employs factored item similarity model that is capable of dealing with sparse data.

\begin{figure*}[!tp]
	\centering
	\subfigure{
		\includegraphics[width=0.22\linewidth]{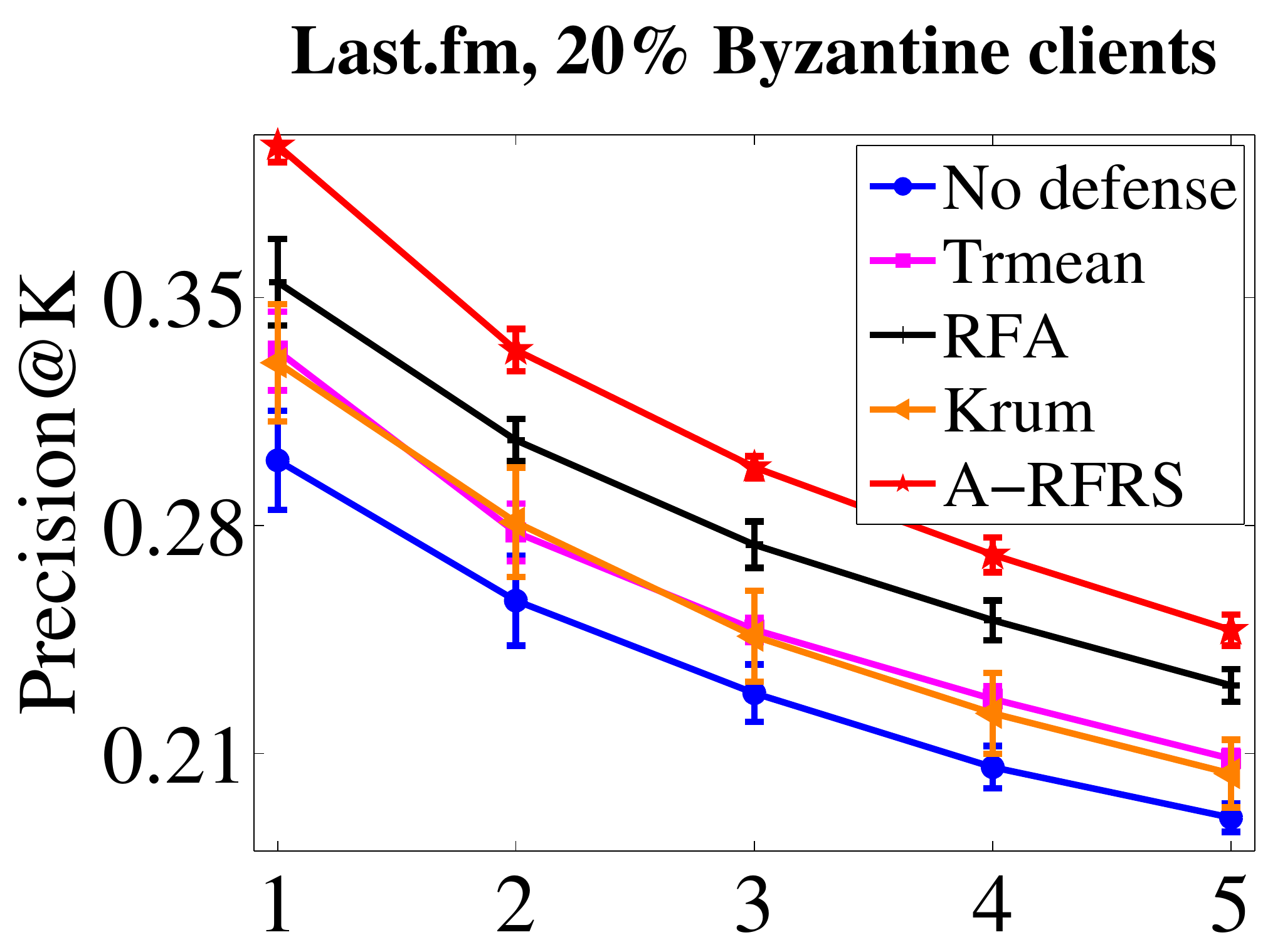}
	}
	\subfigure{
		\includegraphics[width=0.22\linewidth]{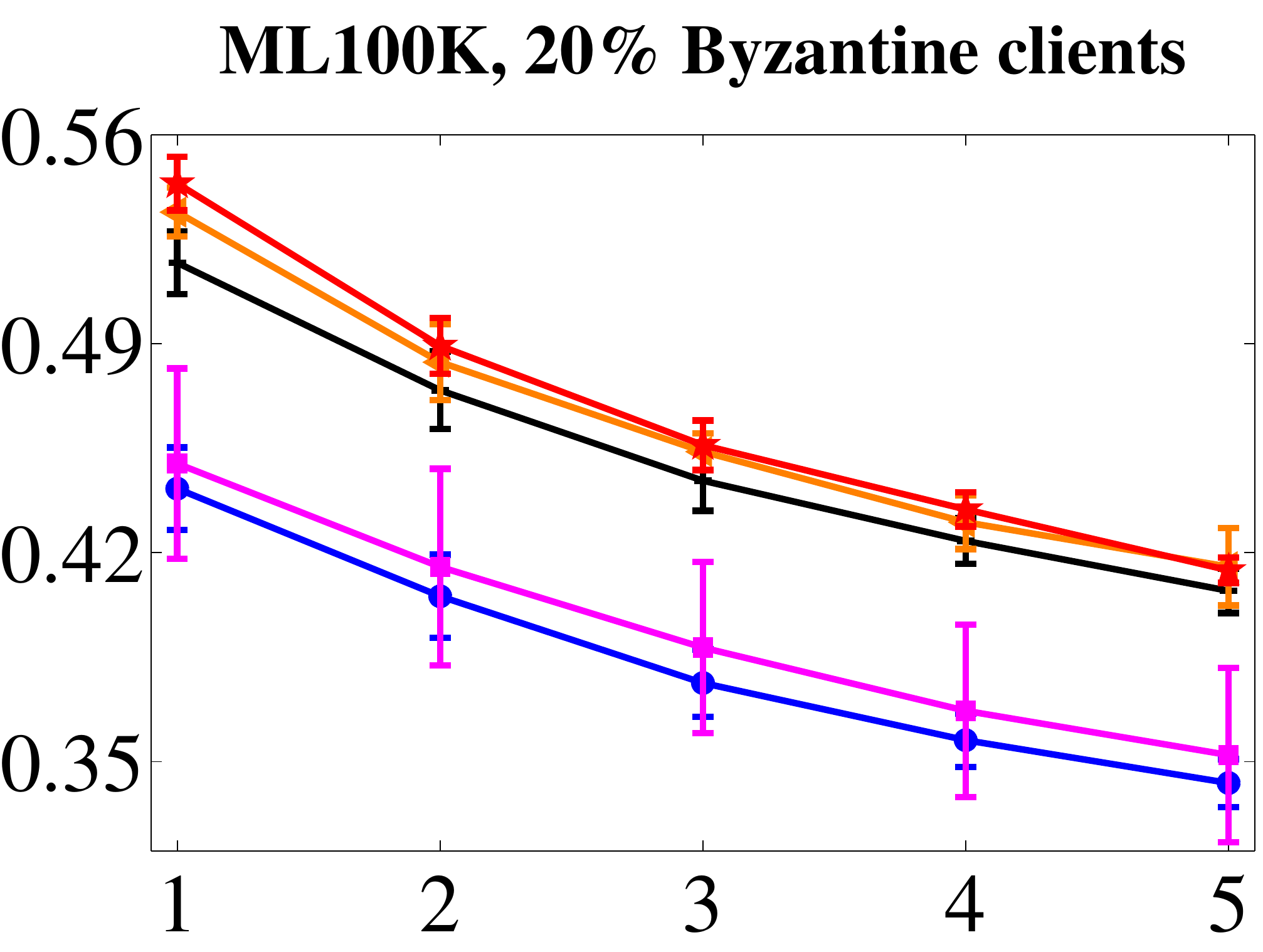}
	}
	\subfigure{
		\includegraphics[width=0.22\linewidth]{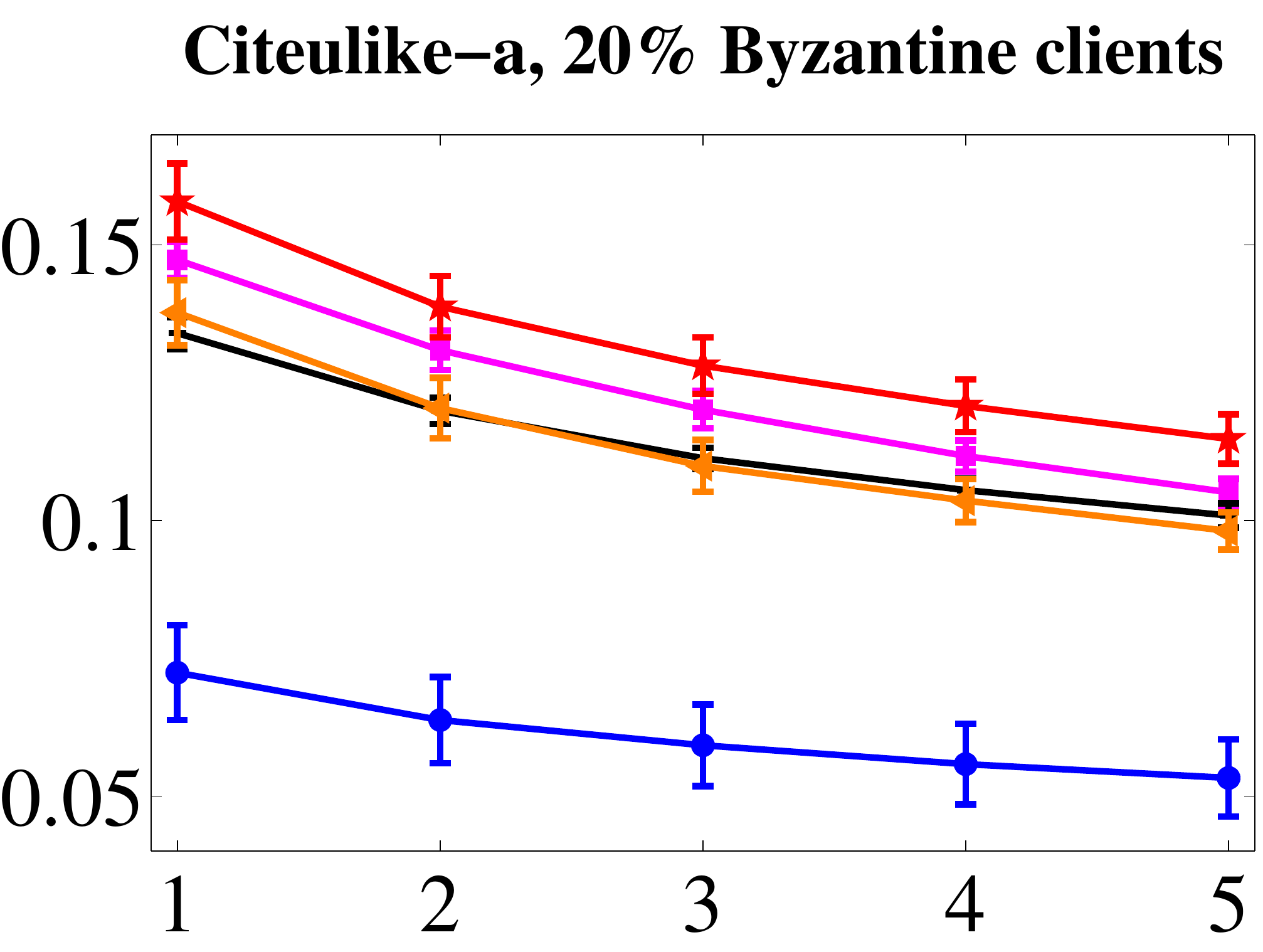}
	}
	\subfigure{
		\includegraphics[width=0.22\linewidth]{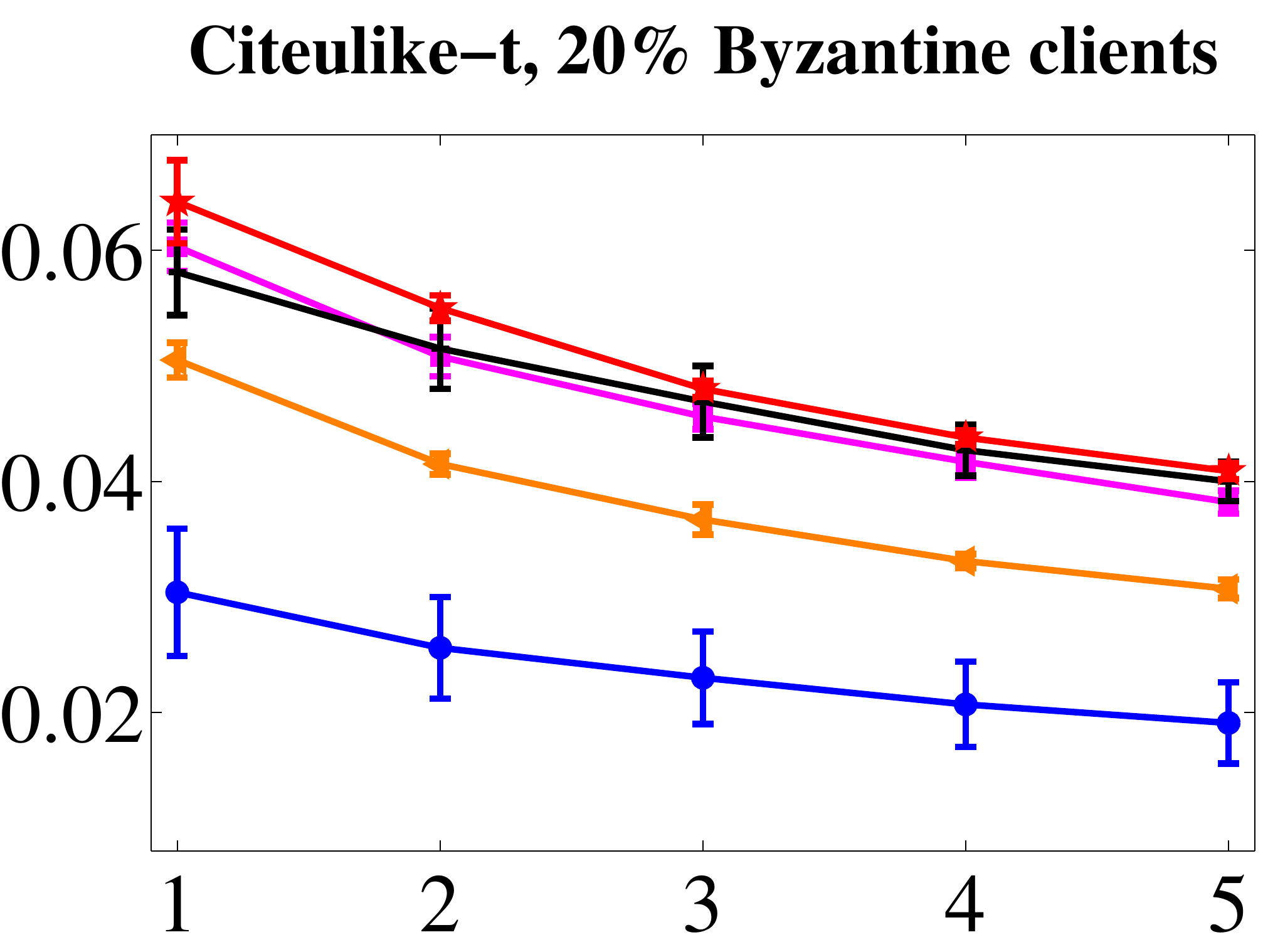}
	}
    \subfigure{
		\includegraphics[width=0.22\linewidth]{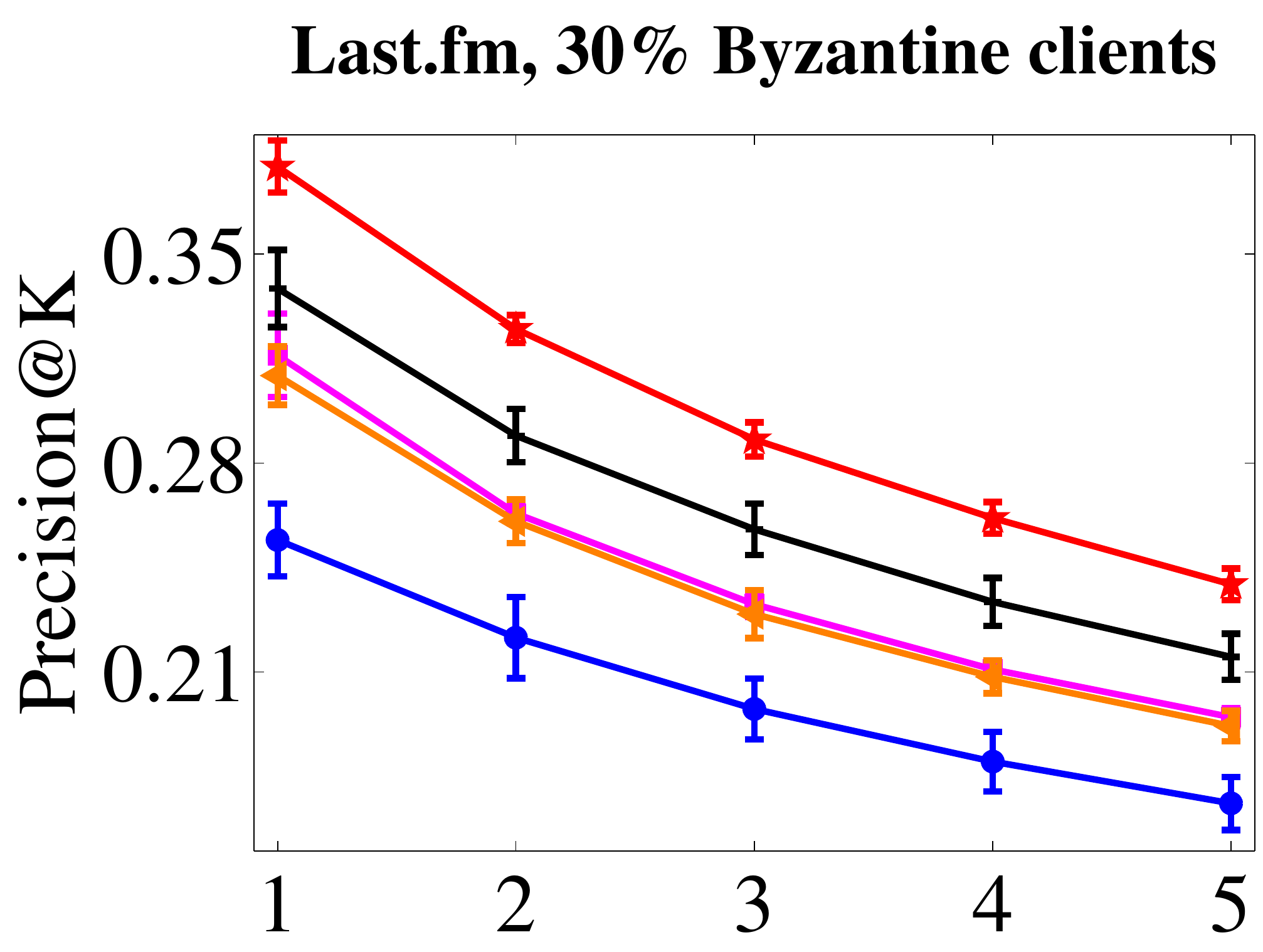}
	}
	\subfigure{
		\includegraphics[width=0.22\linewidth]{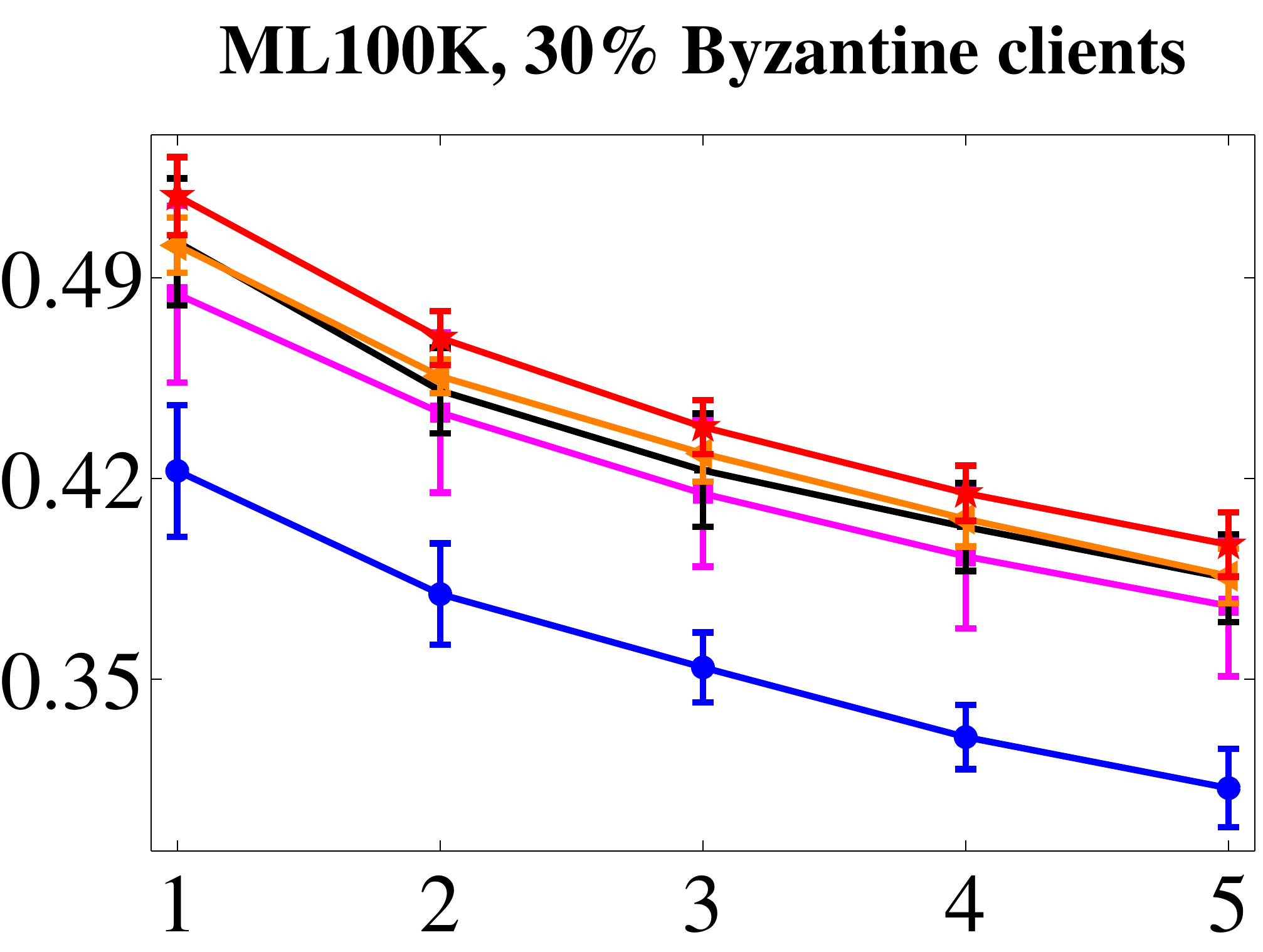}
	}
	\subfigure{
		\includegraphics[width=0.22\linewidth]{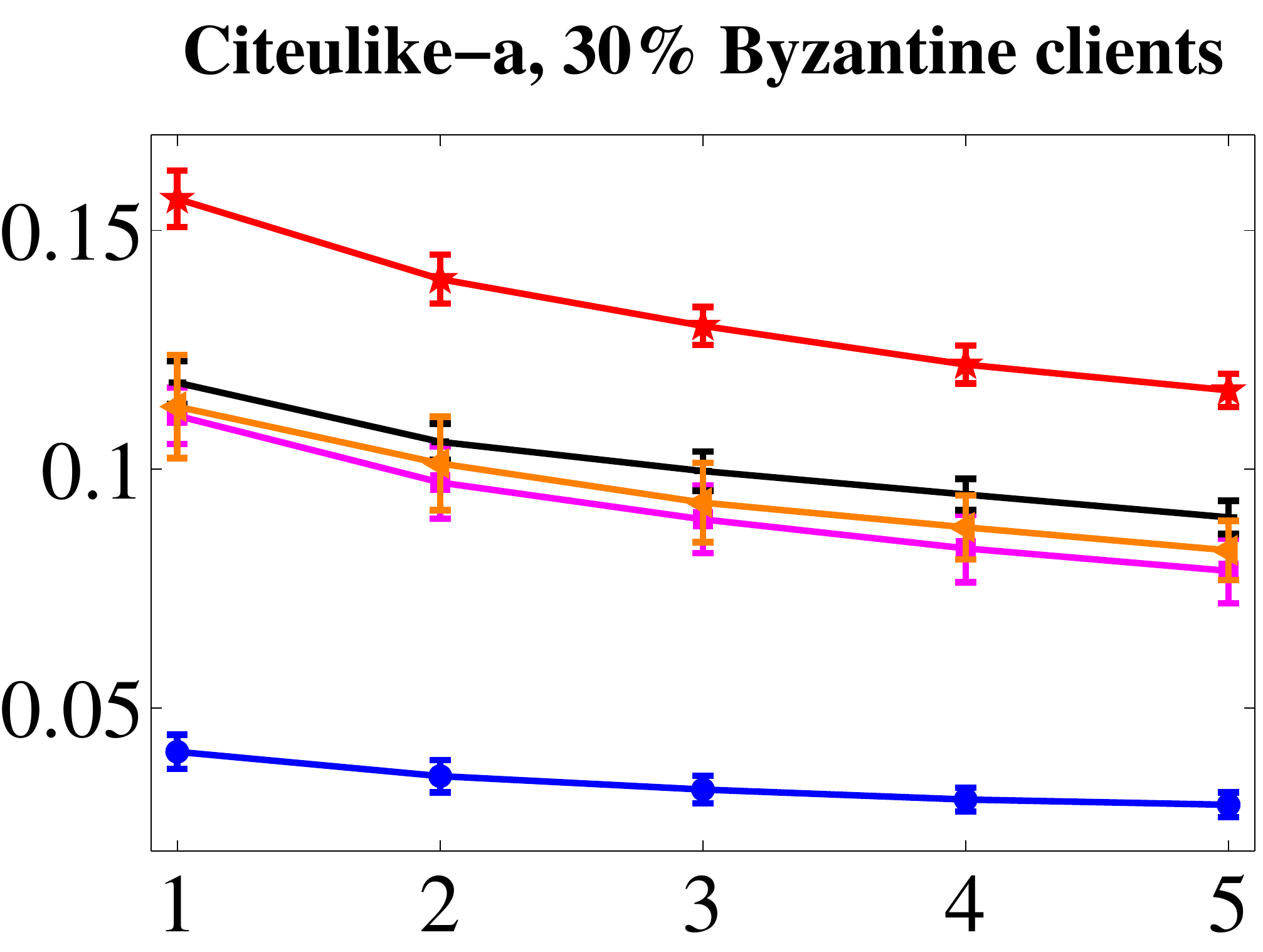}
	}
	\subfigure{
		\includegraphics[width=0.22\linewidth]{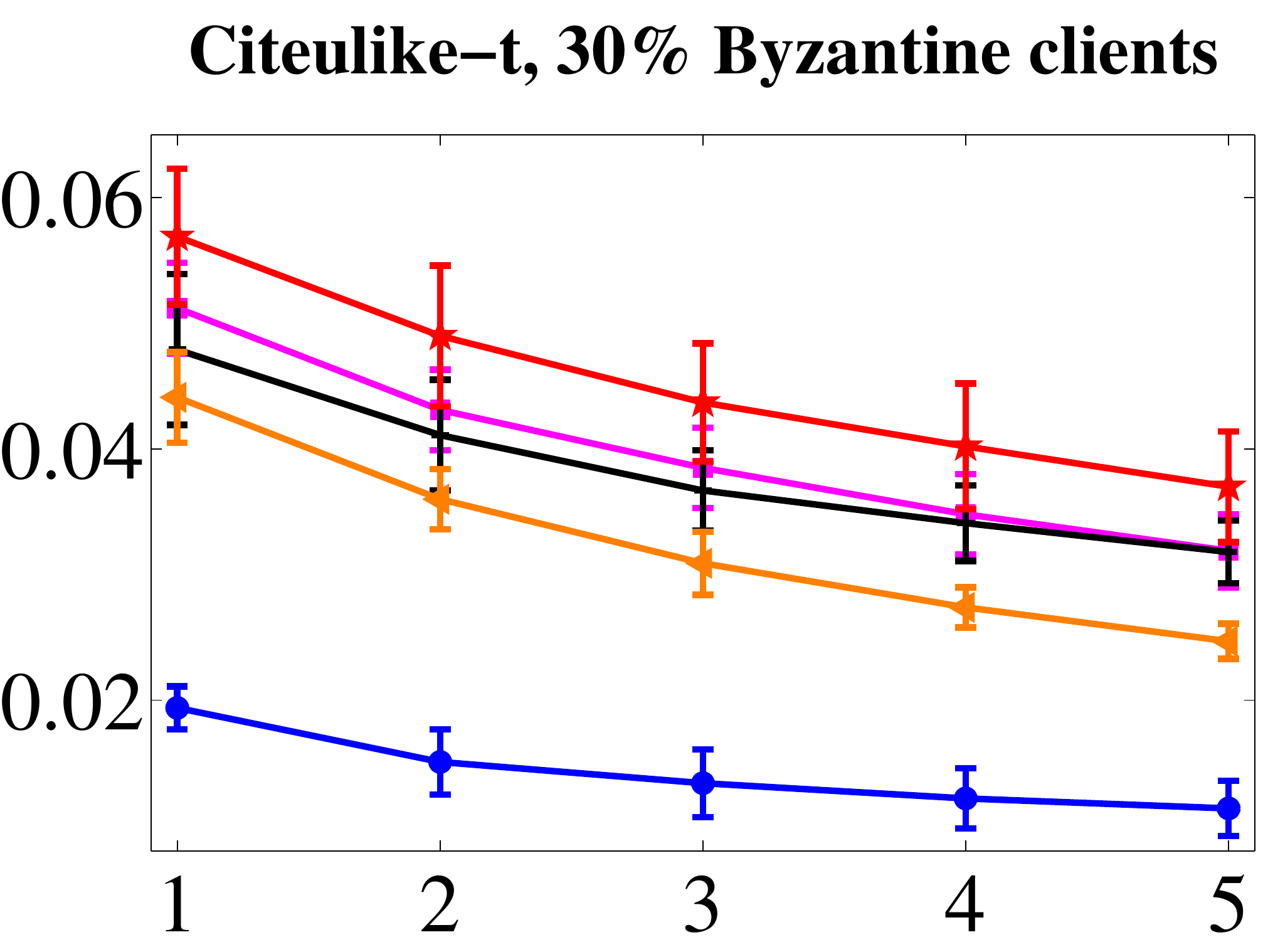}
	}
	\subfigure{
		\includegraphics[width=0.22\linewidth]{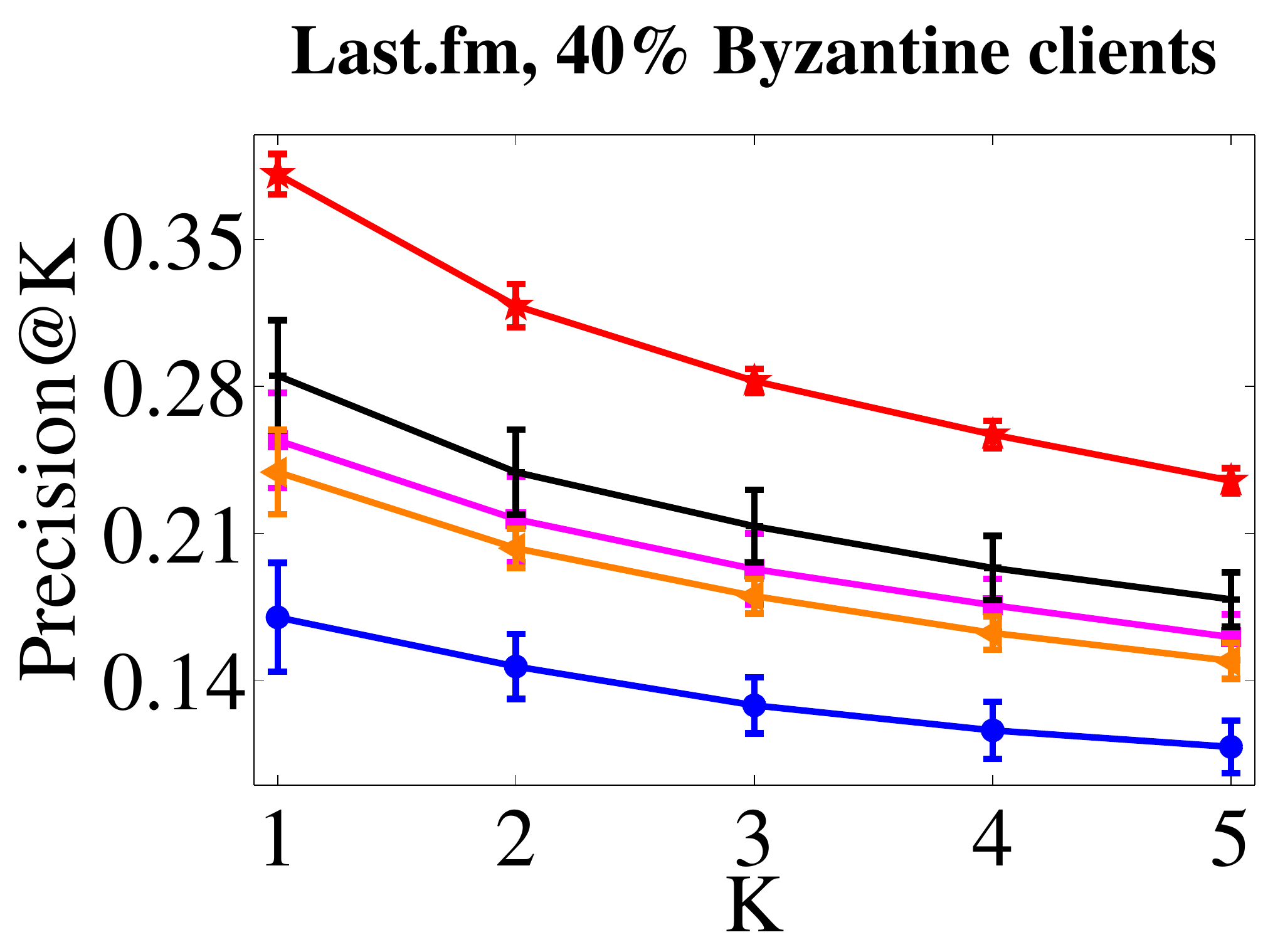}
	}
	\subfigure{
		\includegraphics[width=0.22\linewidth]{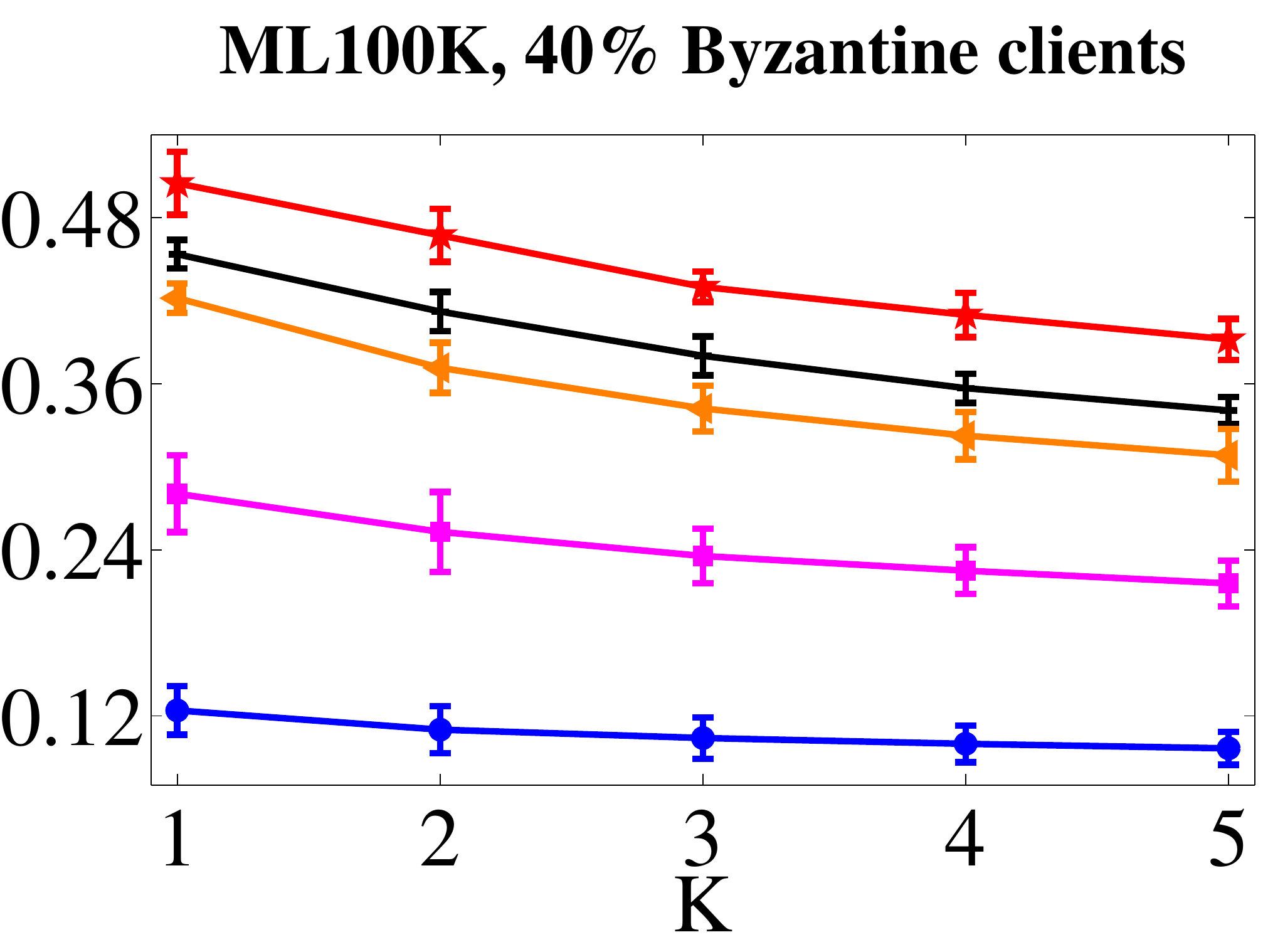}
	}
	\subfigure{
		\includegraphics[width=0.22\linewidth]{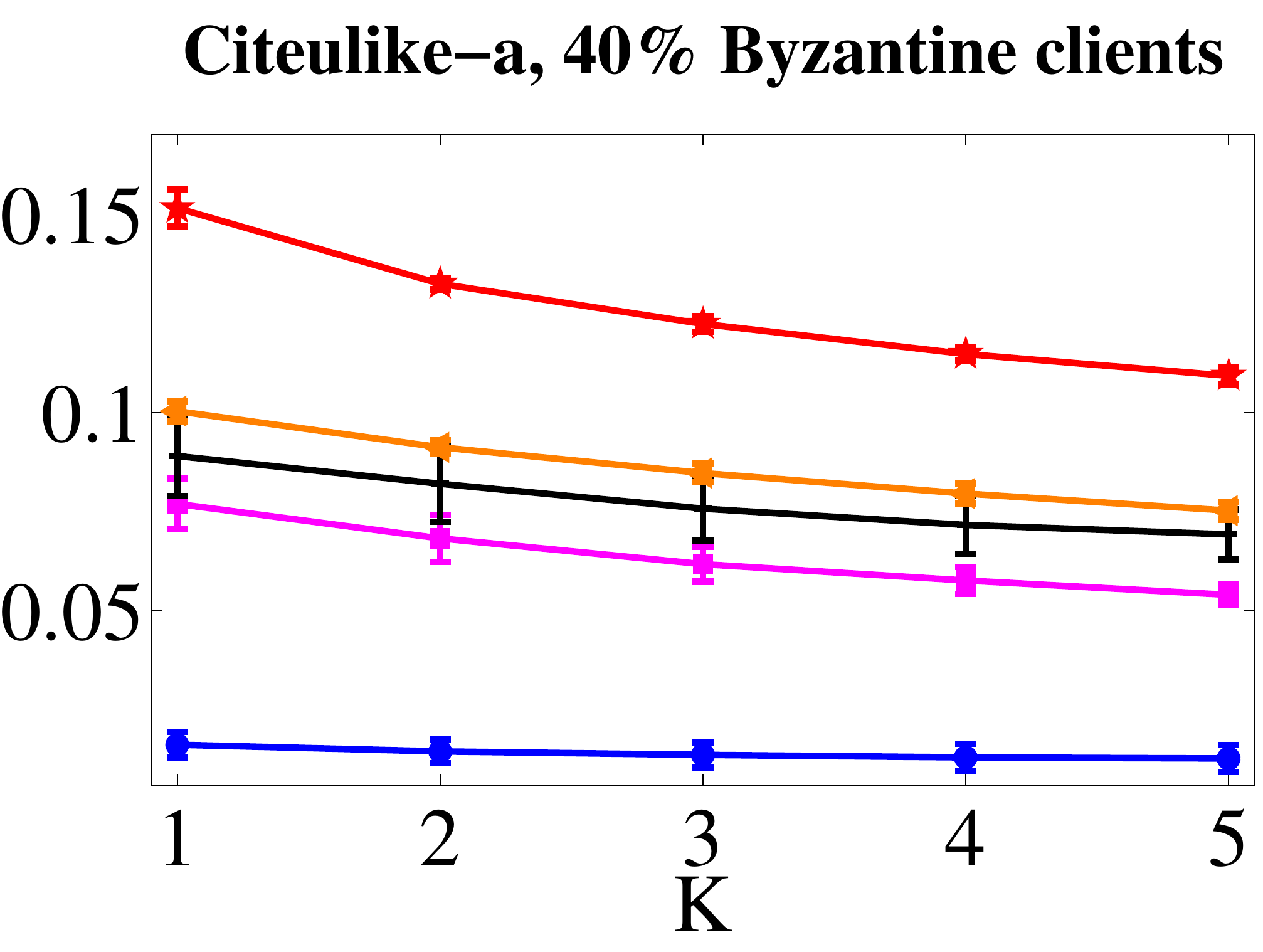}
	}
	\subfigure{
		\includegraphics[width=0.22\linewidth]{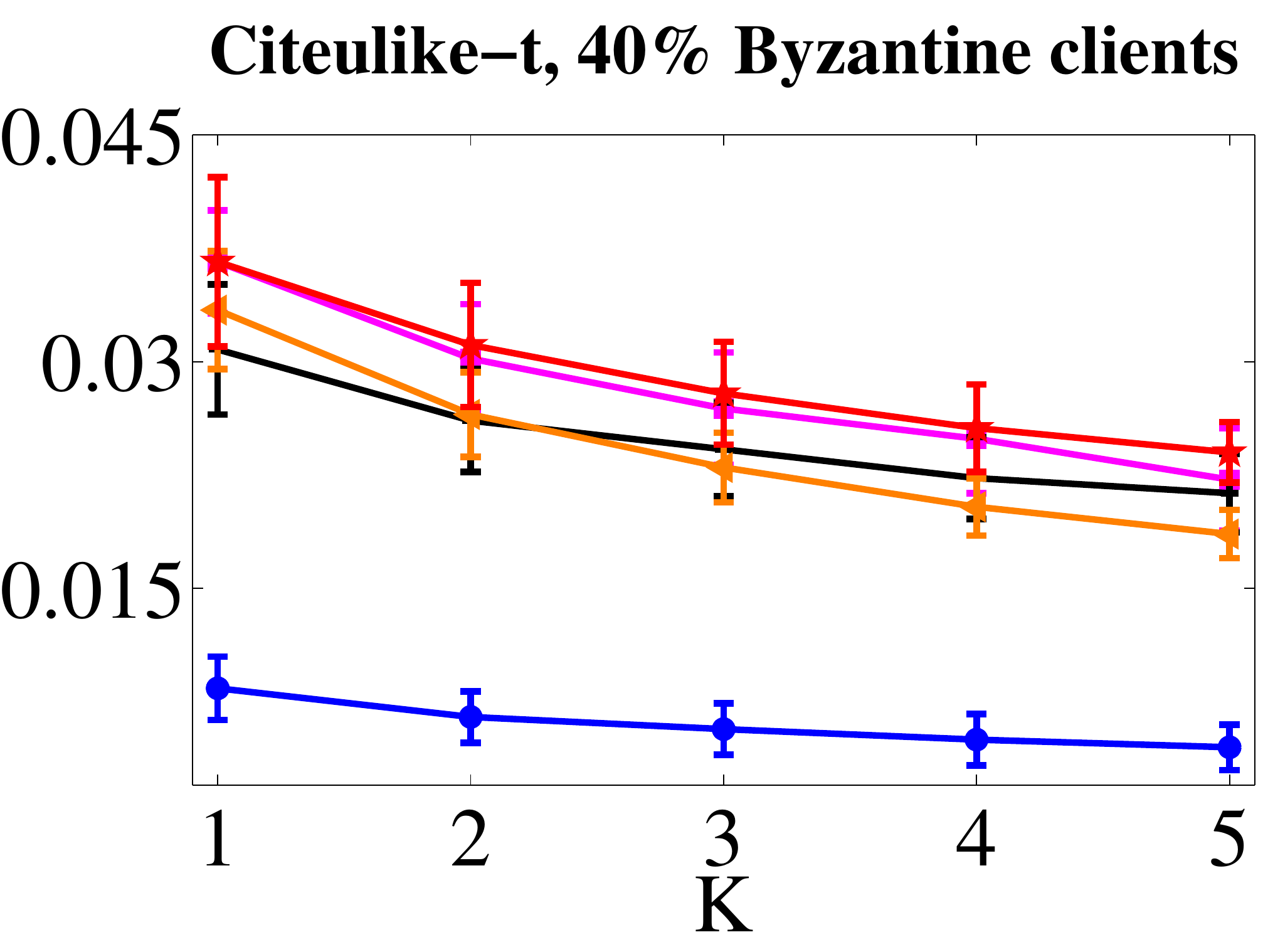}
	}
	\caption{Precision@$K$ (mean and standard deviation) of all methods on 4 datasets and 3 different numbers of Byzantine clients. A-RFRS (red line) is our proposed defense method.}
	\label{fig:performance2}
\end{figure*}
\paragraph{Robust federated recommendation system.} In the second part of our experiments, we demonstrate the efficacy of our A-RFRS (Option $\mathbf{II}$ in Algorithm~\ref{alg:rfrs}) under poisoning attacks when clients use Adam optimizer. We compare A-RFRS with three defense baselines: Krum \cite{conf/nips/BlanchardMGS17}, RFA \cite{journals/corr/abs-1912-13445}, and Trmean \cite{journals/corr/abs-1903-06996}. The detailed descriptions of these baselines are in Appendix
G.

In our setting, the Byzantine clients employ the gradient ascent attack \cite{conf/nips/BlanchardMGS17}. The Byzantine clients firstly use their own data to compute the gradient $\mathbf{g}$ normally, but use $-\mathbf{g}$ to calculate $\widetilde{\mathbf{m}}, \widetilde{\mathbf{v}}$ and $\widetilde{\boldsymbol{\theta}}$ and send them back to the server.
We have also tried additive noise attack \cite{journals/corr/abs-2002-00211}, where Byzantine clients add Gaussian noise to their model parameters, but the additive noise attack has minor poisoning effects on the global model. Thus, we do not consider the additive noise attack. In addition, we also do not consider the camouflage attack (detailed in Section~\ref{sec:camouflage}) in this section since the existing defense methods only examine the model parameters, therefore they cannot defend camouflage attack at all.

In Figure \ref{fig:performance2}, we set the number of Byzantine clients $\widetilde{n}=0.2n, 0.3n, 0.4n$, and evaluate and compare our A-RFRS with RFA, Trmean, Krum and No defense model (our non-robust A-FRS) on four different datasets.
For each client, we randomly select 80\% of the local data as the training set, but we evaluate the global model with the remaining 20\% data of benign clients.
We run our experiment with five repeated trials and report the average Precision@$K$ (the ranking position $K$ ranges from $1$ to $5$) and its standard deviation. We also report the average Recall@$K$ and its standard deviation in Appendix H.1.

Figure \ref{fig:performance2} demonstrates that
our proposed A-RFRS (red line) outperforms all other defense baselines on the four datasets with different portions of Byzantine clients. For example, on Last.fm dataset with 40\% Byzantine clients, A-RFRS improves the best results of baselines by 33.7\% on Precision@$1$. This shows that our proposed A-RFRS achieves superior performance.
In addition, our A-RFRS outperforms Krum on all four datasets where the original Krum~\cite{conf/nips/BlanchardMGS17} detects Byzantine clients using model parameters. This shows that gradients are more suitable than model parameters for detecting Byzantine clients.
To conclude, besides the theoretical guarantee in Section 4.3, we empirically show the efficacy of our robust learning strategy against poisoning attacks.

\section{Conclusion}
This paper proposes a novel robust learning strategy to defend poisoning attacks in momentum-based federated recommendation system.
We first show that Byzantine clients can camouflage the model parameters and elude existing defense methods by launching an effective camouflage attack.
To solve this problem, we propose to use gradients (instead of model parameters) to filter out Byzantine clients in momentum-based federated recommendation system. We theoretically proved that our proposed robust learning strategy is Byzantine resilient and empirically verified its efficacy.
A promising future direction is extending the current work to defend targeted poisoning attacks.

\clearpage

\bibliographystyle{plain}
\bibliography{arxiv}
\clearpage
\appendix
\section{Notations}
\begin{table}[h]
\scriptsize 
\centering
\normalsize
\caption{Notation table.}
\label{tbl:notations}
\begin{tabular}{|c|c|}
\hline
Notation & Description \\
\hline
$n$ & Total number of clients \\
\hline
$\widetilde{n}$ & Total number of Byzantine clients \\
\hline
$\mathcal{F}_t$ & Set of selected clients for aggregation at round $t$ \\
\hline
$\overline{\mathbf{m}}^{t-1}$ & Aggregated first moment at round $t-1$ \\
\hline
$\overline{\mathbf{v}}^{t-1}$ &  Aggregated second moment at round $t-1$\\
\hline
$\overline{\mathbf{r}}^{t-1}$ &  Aggregated squared gradient at round $t-1$\\
\hline
$\overline{\boldsymbol{\theta}}^{t-1}$ &  Aggregated model parameter at round $t-1$\\
\hline
$\hat{\mathbf{g}}^{ti}$ & Gradient of $i$-th client in $\mathcal{F}_t$\\
\hline
$\hat{\mathbf{m}}^{ti}$ & First moment of $i$-th client in $\mathcal{F}_t$\\
\hline
$\hat{\mathbf{v}}^{ti}$ & Second moment of $i$-th client in $\mathcal{F}_t$\\
\hline
$\hat{\mathbf{r}}^{ti}$ & Squared gradient of a client in $\mathcal{F}_t$\\
\hline
$\hat{\boldsymbol{\theta}}^{ti}$ & Model parameter of $i$-th client in $\mathcal{F}_t$\\
\hline
$\mathbf{g}^{ti}$ & Gradient of $i$-th benign client at round $t$\\
\hline
$\mathbf{m}^{ti}$ & First moment of $i$-th benign client at round $t$\\
\hline
$\mathbf{v}^{ti}$ & Second moment of $i$-th benign client at round $t$\\
\hline
$\mathbf{r}^{ti}$ & Squared gradient of $i$-th benign client at round $t$\\
\hline
$\boldsymbol{\theta}^{ti}$ & Model parameter of $i$-th benign client at round $t$\\
\hline
$\widetilde{\mathbf{g}}^{ti}$ & Gradient of $i$-th Byzantine client at round $t$\\
\hline
$\widetilde{\mathbf{m}}^{ti}$ & First moment of $i$-th Byzantine client at round $t$\\
\hline
$\widetilde{\mathbf{v}}^{ti}$ & Second moment of $i$-th Byzantine client at round $t$\\
\hline
$\widetilde{\mathbf{r}}^{ti}$ & Squared gradient of $i$-th benign Byzantine at round $t$\\
\hline
$\widetilde{\boldsymbol{\theta}}^{ti}$ & Model parameter of $i$-th benign Byzantine at round $t$\\
\hline
$\beta_1$ & Hyperparameter of Adam, $0<\beta_1<1$ \\
\hline
$\beta_2$ & Hyperparameter of Adam, $0<\beta_2<1$ \\
\hline
$\beta_3$ & Hyperparameter of SGD with momentum, $0<\beta_3<1$ \\
\hline
$\beta_4$ & Hyperparameter of RMSProp, $0<\beta_4<1$ \\
\hline
$\odot$ & Element-wise multiplication \\
\hline
\end{tabular}
\end{table}
\section{Proof of Theorem 1}
In this section, we show the proof of Theorem 1.

Suppose $\widetilde{n}$ out of $n$ clients are Byzantine.
Let $G^t=\{\mathbf{g}^{ti}|i\in\{1,...,n-\widetilde{n}\}\}$ be the gradient set of $n-\widetilde{n}$ benign clients at round $t$. Let  $\widetilde{G}^t=\{\widetilde{\mathbf{g}}^{ti}|i\in\{1,...,\widetilde{n}\}\}$ be the gradient set of $\widetilde{n}$ Byzantine clients at round $t$.
Let $\mathcal{F}_t$ be the set of selected clients for aggregation.
Let $\hat{G}^t=\{\hat{\mathbf{g}}^{ti}|\text{client} \, i\in\mathcal{F}_t\}$ be the gradient set of clients in $\mathcal{F}_t$.
We define Adam-Byzantine resilience as follows:

\textbf{Definition 1} Adam-Byzantine Resilience.
For client $i$ in $\mathcal{F}_t$ at training round $t$, we denote its first moment, second moment and model parameter as $\hat{\mathbf{m}}^{ti}$, $\hat{\mathbf{v}}^{ti}$ and $\hat{\boldsymbol{\theta}}^{ti}$.  For any benign client $j$ at training round $t$, we denote its first moment, second moment, and model parameter as $\mathbf{m}^{tj}$, $\mathbf{v}^{tj}$ and $\boldsymbol{\theta}^{tj}$.
A defense method is Adam-Byzantine resilient, if for the round $T$ there exist positive constant numbers $C_m$, $C_v$ and $C_\theta$, such that
\begin{enumerate}
    \item $\sum\limits_{t=1}^T\sum\limits_{\text{client} \, i \in \mathcal{F}_t}\sum\limits_{\text{benign client} \, j}\norm{\hat{\mathbf{m}}^{ti}-\mathbf{m}^{tj}}\leq C_m$;
    \item $\sum\limits_{t=1}^T\sum\limits_{\text{client} \, i \in \mathcal{F}_t}\sum\limits_{\text{benign client} \, j}\norm{\hat{\mathbf{v}}^{ti}-\mathbf{v}^{tj}}\leq C_v$;
    \item $\sum\limits_{t=1}^T\sum\limits_{\text{client} \, i \in \mathcal{F}_t}\sum\limits_{\text{benign client} \, j}\norm{\hat{\boldsymbol{\theta}}^{ti}-\boldsymbol{\theta}^{tj}}\leq C_\theta$.
\end{enumerate}

\textbf{Assumption 1}. For any gradient $\mathbf{g}$, its norm is upper bounded by a positive constant number $g_{max}$.
Formally, $\norm{\mathbf{g}}\leq g_{max}$ with $\mathbf{g}\in G^t\cup \widetilde{G}^t, t\in \mathbb{N}^*$.

\textbf{Assumption 2}. After $T'$ rounds of training, each component of $\overline{\mathbf{v}}^{t-1}$ is lower bounded by a positive constant  number $v_{min}$. Formally, for any round $t$ with $t>T'$, $\overline{v}^{t-1}_k\geq v_{min}$, where $\overline{v}^{t-1}_k$ denotes the $k$-th component of $\overline{\mathbf{v}}^{t-1}$.

\textbf{Lemma 1}. Let $k$ be any positive integer.
if $\sum\limits_{i=1}^k a_i\leq A$ and $\sum\limits_{i=1}^k b_i\leq B$ with $a_i,b_i,A,B\geq 0$. Then, $\sum\limits_{i=1}^k a_ib_i\leq AB$.

\begin{proof}
\begin{equation}\label{eq:lemma1}
\begin{split}
AB&\geq\left(\sum\limits_{i=1}^k a_t\right)\left(\sum\limits_{i=1}^k b_t\right) \\
&=\sum\limits_{i=1}^k\sum\limits_{j=1}^{k} a_i  b_j \\
&\geq\sum\limits_{i=1}^k a_ib_i.
\end{split}
\end{equation}
Thus,$\sum\limits_{i=1}^k a_ib_i\leq AB$ holds.
\end{proof}
\textbf{Lemma 2}. If Assumption 1 holds, then for any client $i$ with first moment $\mathbf{m}^{ti}$ at round $t$, the norm of its first moment is upper bound by $g_{max}$. Formally, $\norm{\mathbf{m}^{ti}}\leq g_{max}$.
\begin{proof}
We prove the lemma by mathematical induction.

When $t=1$, since $\overline{\mathbf{m}}^{t-1}=\overline{\mathbf{m}}^0=\mathbf{0}$ and according to Assumption 1,
\begin{equation}
\begin{split}
\norm{\mathbf{m}^{1i}}&=\norm{\beta_1  \overline{\mathbf{m}}^0+(1-\beta_1)  \mathbf{g}^{1i}} \\
&= \norm{(1-\beta_1)  \mathbf{g}^{1i}} \\
&\leq (1-\beta_1)  g_{max} \\
&\leq g_{max}.
\end{split}
\end{equation}
The first equality is due to the definition of first moment. The first inequality is due to Assumption 1. Thus, $\norm{\mathbf{m}^{1i}}\leq g_{max}$ holds.

Suppose when $t=k$, $\norm{\mathbf{m}^{ti}}=\norm{\mathbf{m}^{ki}}\leq g_{max}$ holds.

When $t=k+1$, firstly, we focus on $\overline{\mathbf{m}}^k$. Recall that the aggregation rule is weighted aggregation of all clients in $\mathcal{F}_k$. Formally,
\begin{equation}
\begin{split}
\overline{\mathbf{m}}^k=\frac{\sum\limits_{i\in \mathcal{F}_k} \lambda_i  \mathbf{m}^{ki}}{\sum\limits_{i\in \mathcal{F}_k}\lambda_i},
\end{split}
\end{equation}
where $\lambda_i$ is the weight for the $i$-th first moment $\mathbf{m}^{ki}$.
When $t=k$, $\norm{\mathbf{m}^{ki}}\leq g_{max}$ holds.
Thus,
\begin{equation}\label{eq:lemma2}
\begin{split}
\norm{\overline{\mathbf{m}}^k}&=\norm{\frac{\sum\limits_{i\in \mathcal{F}_k} \lambda_i  \mathbf{m}^{ki}}{\sum\limits_{i\in \mathcal{F}_k}\lambda_i}} \\
&=\frac{\sum\limits_{i\in \mathcal{F}_k} \lambda_i  \norm{\mathbf{m}^{ki}}}{\sum\limits_{i\in \mathcal{F}_k}\lambda_i} \\
&\leq\frac{\sum\limits_{i\in \mathcal{F}_k} \lambda_i  g_{max}}{\sum\limits_{i\in \mathcal{F}_k}\lambda_i}\\
&= g_{max}.
\end{split}
\end{equation}
The second equality is due to absolutely homogeneous of matrix norm.
Now we draw our attention to $\mathbf{m}^{(k+1)i}$:
\begin{equation}
\begin{split}
\norm{\mathbf{m}^{(k+1)i}}&=\norm{\beta_1 \overline{\mathbf{m}}^k+(1-\beta_1)  \mathbf{g}^{(k+1)i}}\\
&\leq\beta_1 \norm{\overline{\mathbf{m}}^k}+(1-\beta_1) \norm{ \mathbf{g}^{(k+1)i}} \\
&\leq \beta_1  g_{max}+(1-\beta_1)  g_{max}\\
&= g_{max}.
\end{split}
\end{equation}
The first inequality is due to triangle inequality and the second inequality is due to Eq. (\ref{eq:lemma2}) and Assumption 1.
Thus, when $t=k+1$, $\norm{\mathbf{m}^{ti}}=\norm{\mathbf{m}^{(k+1)i}}\leq g_{max}$ holds.

Therefore, for any $t\in \mathbb{N}^*$ , $\norm{\mathbf{m}^{ti}}\leq g_{max}$ holds.
\end{proof}

\textbf{Theorem 1}. A-RFRS is Adam-Byzantine resilient, if Assumption 1 and Assumption 2 hold, and for any client $i$ in $\mathcal{F}_t$ with gradient $\hat{\mathbf{g}}^{ti} \in \hat{G}^t$, for any benign client $j$ with gradient $\mathbf{g}^{tj}\in G^t$, and for training round $T\in\mathbb{N}^*$, there exist a positive constant number $C_g$, such that
$$\sum\limits_{t=1}^T\sum\limits_{\text{client} \, i \in \mathcal{F}_t}\sum\limits_{\text{benign client} \, j}\norm{\hat{\mathbf{g}}^{ti}-\mathbf{g}^{tj}}\leq C_g.$$

\begin{proof}

Intuitively, since $\hat{\mathbf{g}}^{ti}$ is not far from $\mathbf{g}^{tj}$, $\hat{\mathbf{m}}^{ti}$, $\hat{\mathbf{v}}^{ti}$ and $\hat{\boldsymbol{\theta}}^{ti}$ should also be close to $\mathbf{m}^{tj}$, $\mathbf{v}^{tj}$ and $\boldsymbol{\theta}^{tj}$. We will prove them step by step below.

The first step is to prove $\hat{\mathbf{m}}^{ti}$ is close to $\mathbf{m}^{tj}$.

According to the definition of first moment,
\begin{equation}\label{eq:m1}
\begin{split}
\norm{\hat{\mathbf{m}}^{ti}-\mathbf{m}^{tj}}&=\norm{\beta_1  \overline{\mathbf{m}}^{t-1}+(1-\beta_1)  \hat{\mathbf{g}}^{ti}-(\beta_1  \overline{\mathbf{m}}^{t-1}+(1-\beta_1)  \mathbf{g}^{tj})} \\
&=(1-\beta_1) \norm{\hat{\mathbf{g}}^{ti}-\mathbf{g}^{tj}}.
\end{split}
\end{equation}
The second equality is due to absolutely homogeneous of matrix norm.
By adding round $1$ to round $T$, all clients in $\mathcal{F}_t$ and all benign clients together,
\begin{equation}\label{eq:m3}
\begin{split}
\sum\limits_{t=1}^T\sum\limits_{\text{client} \, i \in \mathcal{F}_t}\sum\limits_{\text{benign client} \, j}
\norm{\hat{\mathbf{m}}^{ti}-\mathbf{m}^{tj}}
&=\sum\limits_{t=1}^T\sum\limits_{\text{client} \, i \in \mathcal{F}_t}\sum\limits_{\text{benign client} \, j}
(1-\beta_1) \norm{\hat{\mathbf{g}}^{ti}-\mathbf{g}^{tj}}\\
&=(1-\beta_1) \sum\limits_{t=1}^T\sum\limits_{\text{client} \, i \in \mathcal{F}_t}\sum\limits_{\text{benign client} \, j}
\norm{\hat{\mathbf{g}}^{ti}-\mathbf{g}^{tj}} \\
&\leq (1-\beta_1)  C_g.
\end{split}
\end{equation}
The first equlity is due to Eq. (\ref{eq:m1}). The inequality is due to the assumption of Theorem 1.
Let $C_m=(1-\beta_1)  C_g$. Then $C_m$ is a positive constant number, and
\begin{equation}\label{eq:m4}
\begin{split}
\sum\limits_{t=1}^T\sum\limits_{\text{client} \, i \in \mathcal{F}_t}\sum\limits_{\text{benign client} \, j}\norm{\hat{\mathbf{m}}^{ti}-\mathbf{m}^{tj}}&\leq C_m.
\end{split}
\end{equation}
Thus, condition 1 of Adam-Byzantine resilience holds.

The second step is to prove $\hat{\mathbf{v}}^{ti}$ is close to $\mathbf{v}^{tj}$.
\begin{equation}\label{eq:v1}
\begin{split}
\norm{\hat{\mathbf{v}}^{ti}-\mathbf{v}^{tj}}&=\norm{\beta_2  \overline{\mathbf{v}}^{t-1}+(1-\beta_2)  \hat{\mathbf{g}}^{ti}\odot \hat{\mathbf{g}}^{ti}-(\beta_2  \overline{\mathbf{v}}^{t-1}+(1-\beta_2)  \mathbf{g}^{tj}\odot \mathbf{g}^{tj})} \\
&= (1-\beta_2) \norm{\hat{\mathbf{g}}^{ti}\odot\hat{\mathbf{g}}^{ti} -\mathbf{g}^{tj}\odot \mathbf{g}^{tj}} \\
&= (1-\beta_2)   \norm{(\hat{\mathbf{g}}^{ti}+\mathbf{g}^{tj})\odot(\hat{\mathbf{g}}^{ti}-\mathbf{g}^{tj})} \\
&\leq (1-\beta_2)   \norm{\hat{\mathbf{g}}^{ti}+\mathbf{g}^{tj}}  \norm{\hat{\mathbf{g}}^{ti}-\mathbf{g}^{tj}} \\
&= (1-\beta_2)   \norm{(\hat{\mathbf{g}}^{ti}-\mathbf{g}^{tj})+2  \mathbf{g}^{tj}}  \norm{\hat{\mathbf{g}}^{ti}-\mathbf{g}^{tj}}  \\
&\leq (1-\beta_2)   \norm{\hat{\mathbf{g}}^{ti}-\mathbf{g}^{tj}}\left(\norm{\hat{\mathbf{g}}^{ti}-\mathbf{g}^{tj}}+2\norm{\mathbf{g}^{tj}}\right) \\
&= (1-\beta_2) \norm{\hat{\mathbf{g}}^{ti}-\mathbf{g}^{tj}}^2
+2(1-\beta_2) \norm{\hat{\mathbf{g}}^{ti}-\mathbf{g}^{tj}}\norm{\mathbf{g}^{tj}}.
\end{split}
\end{equation}
The first equality is due to the definition of second moment. The second equality is due to absolutely homogeneous of matrix norm. The third equality follows from square of the difference formula. The first inequality is due to submultiplicativity of matrix norm. The second inequality is due to triangle inequality.
By adding round $1$ to round $T$, all clients in $\mathcal{F}_t$ and all benign clients together,
\begin{equation}\label{eq:v2}
\begin{split}
&\sum\limits_{t=1}^T\sum\limits_{\text{client} \, i \in \mathcal{F}_t}\sum\limits_{\text{benign client} \, j}
\norm{\hat{\mathbf{v}}^{ti}-\mathbf{v}^{tj}}\\
\leq&
\sum\limits_{t=1}^T\sum\limits_{\text{client} \, i \in \mathcal{F}_t}\sum\limits_{\text{benign client} \, j}
\left((1-\beta_2) \norm{\hat{\mathbf{g}}^{ti}-\mathbf{g}^{tj}}^2
+2(1-\beta_2) \norm{\hat{\mathbf{g}}^{ti}-\mathbf{g}^{tj}}\norm{\mathbf{g}^{tj}}\right)\\
\leq&
(1-\beta_2)\sum\limits_{t=1}^T\sum\limits_{\text{client} \, i \in \mathcal{F}_t}\sum\limits_{\text{benign client} \, j}
\norm{\hat{\mathbf{g}}^{ti}-\mathbf{g}^{tj}}^2\\
&+2(1-\beta_2)\sum\limits_{t=1}^T\sum\limits_{\text{client} \, i \in \mathcal{F}_t}\sum\limits_{\text{benign client} \, j}
\norm{\hat{\mathbf{g}}^{ti}-\mathbf{g}^{tj}}g_{max} \\
\leq& (1-\beta_2)C_g^2+2(1-\beta_2)g_{max}C_g.
\end{split}
\end{equation}
The first inequality is due to Eq. (\ref{eq:v1}). The second inequality follows from Assumption 1. The third inequality is due to the assumption of Theorem 1 and Lemma 1.
Let $C_v=(1-\beta_2)C_g^2+2(1-\beta_2)g_{max}C_g$. Then,  $C_v$ is a positive constant  number, and
\begin{equation}\label{eq:v3}
\begin{split}
\sum\limits_{t=1}^T\sum\limits_{\text{client} \, i \in \mathcal{F}_t}\sum\limits_{\text{benign client} \, j}
\norm{\hat{\mathbf{v}}^{ti}-\mathbf{v}^{tj}}&\leq C_v.
\end{split}
\end{equation}
Therefore, condition 2 of Adam-Byzantine resilience holds.

The third step is to prove $\hat{\boldsymbol{\theta}}^{ti}$ is close to $\boldsymbol{\theta}^{tj}$. According to the definition, $\hat{\boldsymbol{\theta}}^{ti}=\overline{\boldsymbol{\theta}}^{t-1}-\frac{\hat{\mathbf{m}}^{ti}}{\sqrt{\hat{\mathbf{v}}^{ti}}}$, $\boldsymbol{\theta}^{tj}=\overline{\boldsymbol{\theta}}^{t-1}-\frac{\mathbf{m}^{tj}}{\sqrt{\mathbf{v}^{tj}}}$ (for simplicity, we omit the learning rate).
We can decompose the proof into four parts:
\begin{enumerate}
    \item $\sqrt{\hat{\mathbf{v}}^{ti}}$ is close to $\sqrt{\mathbf{v}^{tj}}$
    \item $\frac{1}{\sqrt{\hat{\mathbf{v}}^{ti}}}$ is close to $\frac{1}{\sqrt{\mathbf{v}^{tj}}}$
    \item $\frac{\hat{\mathbf{m}}^{ti}}{\sqrt{\hat{\mathbf{v}}^{ti}}}$ is close to $\frac{\mathbf{m}^{tj}}{\sqrt{\mathbf{v}^{tj}}}$
    \item $\hat{\boldsymbol{\theta}}^{ti}$ is close to $\boldsymbol{\theta}^{tj}$.
\end{enumerate}
The first part is to prove $\sqrt{\hat{\mathbf{v}}^{ti}}$ is close to $\sqrt{\mathbf{v}^{tj}}$.
We focus on one component of $\sqrt{\hat{\mathbf{v}}^{ti}}$ and $\sqrt{\mathbf{v}^{tj}}$.
For any vector $\mathbf{v}$,
let $v_k$ denote the $k$-th component of $\mathbf{v}$.
\begin{equation}\label{eq:va1}
\begin{split}
\left|\sqrt{\hat{v}^{ti}_k}-\sqrt{v^{tj}_k}\right|
&= \left|\frac{\hat{v}^{ti}_k-v^{tj}_k}{\sqrt{\hat{v}^{ti}_k}+\sqrt{v^{tj}_k}}\right| \\
&= \left|\hat{v}^{ti}_k-v^{tj}_k\right| \left|\frac{1}{\sqrt{\hat{v}^{ti}_k}+\sqrt{v^{tj}_k}}\right|.
\end{split}
\end{equation}
Recall that $\hat{v}^{ti}_k=\beta_2 \overline{v}^{t-1}_k+(1-\beta_2) (\hat{g}^{ti}_k)^2  $ and $v^{tj}_k=\beta_2 \overline{v}^{t-1}_k+(1-\beta_2)  (g^{tj}_k)^2$.
When $t>T'$,
\begin{equation}\label{eq:va2}
\begin{split}
\left|\frac{1}{\sqrt{\hat{v}^{ti}_k}+\sqrt{v^{tj}_k}}\right|&= \frac{1}{\sqrt{\beta_2 \overline{v}^{t-1}_k+(1-\beta_2) (\hat{g}^{ti}_k)^2}+\sqrt{\beta_2 \overline{v}^{t-1}_k+(1-\beta_2)  (g^{tj}_k)^2}} \\
&\leq \frac{1}{\sqrt{\beta_2 \overline{v}^{t-1}_k}+\sqrt{\beta_2 \overline{v}^{t-1}_k}} \\
&\leq \frac{1}{2 \sqrt{\beta_2  v_{min}}}.
\end{split}
\end{equation}
The second inequality is due to Assumption 2.
By combining Eq. (\ref{eq:va1}) with Eq. (\ref{eq:va2}),
\begin{equation}\label{eq:va3}
\begin{split}
\left|\sqrt{\hat{v}^{ti}_k}-\sqrt{v^{tj}_k}\right|&\leq \frac{1}{2 \sqrt{ \beta_2  v_{min}}}   \left|\hat{v}^{ti}_k-v^{tj}_k\right|.
\end{split}
\end{equation}
We combine all the components together:
\begin{equation}\label{eq:va4}
\begin{split}
\norm{\sqrt{\hat{\mathbf{v}}^{ti}}-\sqrt{\mathbf{v}^{tj}}}
&\leq \frac{1}{2 \sqrt{\beta_2  v_{min}}}   \norm{\hat{\mathbf{v}}^{ti}-\mathbf{v}^{tj}}.
\end{split}
\end{equation}
By adding round $1$ to round $T$, all clients in $\mathcal{F}_t$ and all benign clients together, when $T>T'$,
\begin{equation}\label{eq:va5}
\begin{split}
&\sum\limits_{t=1}^T\sum\limits_{\text{client} \, i \in \mathcal{F}_t}\sum\limits_{\text{benign client} \, j}
\norm{\sqrt{\hat{\mathbf{v}}^{ti}}-\sqrt{\mathbf{v}^{tj}}}\\
=& \sum\limits_{t=1}^{T'}\sum\limits_{\text{client} \, i \in \mathcal{F}_t}\sum\limits_{\text{benign client} \, j}
\norm{\sqrt{\hat{\mathbf{v}}^{ti}}-\sqrt{\mathbf{v}^{tj}}}+
\sum\limits_{t=T'+1}^T\sum\limits_{\text{client} \, i \in \mathcal{F}_t}\sum\limits_{\text{benign client} \, j}
\norm{\sqrt{\hat{\mathbf{v}}^{ti}}-\sqrt{\mathbf{v}^{tj}}} \\
\leq& \sum\limits_{t=1}^{T'}\sum\limits_{\text{client} \, i \in \mathcal{F}_t}\sum\limits_{\text{benign client} \, j}
\norm{\sqrt{\hat{\mathbf{v}}^{ti}}-\sqrt{\mathbf{v}^{tj}}}+\sum\limits_{t=T'+1}^T\sum\limits_{\text{client} \, i \in \mathcal{F}_t}\sum\limits_{\text{benign client} \, j}
\frac{1}{2  \sqrt{\beta_2  v_{min}}}  \norm{\hat{\mathbf{v}}^{ti}-\mathbf{v}^{tj}} \\
=& \sum\limits_{t=1}^{T'}\sum\limits_{\text{client} \, i \in \mathcal{F}_t}\sum\limits_{\text{benign client} \, j}
\norm{\sqrt{\hat{\mathbf{v}}^{ti}}-\sqrt{\mathbf{v}^{tj}}}+\frac{1}{2  \sqrt{\beta_2  v_{min}}}  \sum\limits_{t=T'+1}^T\sum\limits_{\text{client} \, i \in \mathcal{F}_t}\sum\limits_{\text{benign client} \, j}
\norm{\hat{\mathbf{v}}^{ti}-\mathbf{v}^{tj}} \\
\leq& \sum\limits_{t=1}^{T'}\sum\limits_{\text{client} \, i \in \mathcal{F}_t}\sum\limits_{\text{benign client} \, j}
\norm{\sqrt{\hat{\mathbf{v}}^{ti}}-\sqrt{\mathbf{v}^{tj}}}+\frac{C_v}{2  \sqrt{\beta_2  v_{min}}}.
\end{split}
\end{equation}
The first inequality is due to Eq. (\ref{eq:va4}). The second inequality is due to Eq. (\ref{eq:v3}).
Let $C'_v=\sum\limits_{t=1}^{T'}\sum\limits_{\text{client} \, i \in \mathcal{F}_t}\sum\limits_{\text{benign client} \, j}
\norm{\sqrt{\hat{\mathbf{v}}^{ti}}-\sqrt{\mathbf{v}^{tj}}}+\frac{C_v}{2  \sqrt{\beta_2  v_{min}}}$. Then $C'_v$ is a positive constant  number, and
\begin{equation}\label{eq:va7}
\begin{split}
\sum\limits_{t=1}^T\sum\limits_{\text{client} \, i \in \mathcal{F}_t}\sum\limits_{\text{benign client} \, j}
\norm{\sqrt{\hat{\mathbf{v}}^{ti}}-\sqrt{\mathbf{v}^{tj}}}&\leq C'_v.
\end{split}
\end{equation}
When $T\leq T'$,
\begin{equation}\label{eq:va8}
\begin{split}
\sum\limits_{t=1}^{T}\sum\limits_{\text{client} \, i \in \mathcal{F}_t}\sum\limits_{\text{benign client} \, j}
\norm{\sqrt{\hat{\mathbf{v}}^{ti}}-\sqrt{\mathbf{v}^{tj}}}
\leq&\sum\limits_{t=1}^{T'}\sum\limits_{\text{client} \, i \in \mathcal{F}_t}\sum\limits_{\text{benign client} \, j}
\norm{\sqrt{\hat{\mathbf{v}}^{ti}}-\sqrt{\mathbf{v}^{tj}}}\\
\leq& C'_v.
\end{split}
\end{equation}
Thus Eq. (\ref{eq:va7}) still holds.

The second part is to prove $\frac{1}{\sqrt{\hat{\mathbf{v}}^{ti}}}$ is close to $\frac{1}{\sqrt{\mathbf{v}^{tj}}}$. Similar to $\sqrt{\hat{\mathbf{v}}^{ti}}$ and $\sqrt{\mathbf{v}^{tj}}$, we also focus on one components of $\frac{1}{\sqrt{\hat{\mathbf{v}}^{ti}}}$ and $\frac{1}{\sqrt{\mathbf{v}^{tj}}}$.
\begin{equation}\label{eq:vb1}
\begin{split}
\left|\frac{1}{\sqrt{\hat{v}^{ti}_k}}-\frac{1}{\sqrt{v^{tj}_k}}\right|
&= \left|\frac{\sqrt{v^{tj}_k}-\sqrt{\hat{v}^{ti}_k}}{\sqrt{\hat{v}^{ti}_k}  \sqrt{v^{tj}_k}}\right| \\
&= \left|\sqrt{\hat{v}^{ti}_k}-\sqrt{v^{tj}_k}\right| \left|\frac{1}{\sqrt{\hat{v}^{ti}_k}  \sqrt{v^{tj}_k}}\right|.
\end{split}
\end{equation}
Recall that $\sqrt{\hat{v}^{ti}_k}=\sqrt{\beta_2 \overline{v}^{t-1}_k+(1-\beta_2) (\hat{g}^{ti}_k)^2}$ and $\sqrt{v^{tj}_k}=\sqrt{\beta_2 \overline{v}^{t-1}_k+(1-\beta_2)  (g^{tj}_k)^2}$.
When $t>T'$,
\begin{equation}\label{eq:vb2}
\begin{split}
\left|\frac{1}{\sqrt{\hat{v}^{ti}_k}  \sqrt{v^{tj}_k}}\right| &= \frac{1}{\sqrt{\beta_2 \overline{v}^{t-1}_k+(1-\beta_2) (\hat{g}^{ti}_k)^2} \sqrt{\beta_2 \overline{v}^{t-1}_k+(1-\beta_2)  (g^{tj}_k)^2}} \\
&\leq \frac{1}{\sqrt{\beta_2 \overline{v}^{t-1}_k} \sqrt{\beta_2 \overline{v}^{t-1}_k}} \\
&\leq \frac{1}{\sqrt{\beta_2  v_{min}} \sqrt{\beta_2  v_{min}}} \\
&= \frac{1}{\beta_2  v_{min}}.
\end{split}
\end{equation}
The second inequality is due to Assumption 2.
By combining Eq. (\ref{eq:vb1}) with Eq. (\ref{eq:vb2}),
\begin{equation}\label{eq:vb3}
\begin{split}
\left|\frac{1}{\sqrt{\hat{v}^{ti}_k}}-\frac{1}{\sqrt{v^{tj}_k}}\right| &\leq \frac{1}{\beta_2  v_{min}}  \left|\sqrt{\hat{v}^{ti}_k}-\sqrt{v^{tj}_k}\right|.
\end{split}
\end{equation}
We combine all the components together:
\begin{equation}\label{eq:vb4}
\begin{split}
\norm{\frac{1}{\sqrt{\hat{\mathbf{v}}^{ti}}}-\frac{1}{\sqrt{\mathbf{v}^{tj}}}}&\leq \frac{1}{\beta_2  v_{min}}   \norm{\sqrt{\hat{\mathbf{v}}^{ti}}-\sqrt{\mathbf{v}^{tj}}}.
\end{split}
\end{equation}
By adding round $1$ to round $T$, all clients in $\mathcal{F}_t$ and all benign clients together, when $T>T'$,
\begin{equation}\label{eq:vb5}
\begin{split}
&\sum\limits_{t=1}^T\sum\limits_{\text{client} \, i \in \mathcal{F}_t}\sum\limits_{\text{benign client} \, j}
\norm{\frac{1}{\sqrt{\hat{\mathbf{v}}^{ti}}}-\frac{1}{\sqrt{\mathbf{v}^{tj}}}}\\
=&\sum\limits_{t=1}^{T'}\sum\limits_{\text{client} \, i \in \mathcal{F}_t}\sum\limits_{\text{benign client} \, j}
\norm{\frac{1}{\sqrt{\hat{\mathbf{v}}^{ti}}}-\frac{1}{\sqrt{\mathbf{v}^{tj}}}}
+\sum\limits_{t=T'+1}^T\sum\limits_{\text{client} \, i \in \mathcal{F}_t}\sum\limits_{\text{benign client} \, j}
\norm{\frac{1}{\sqrt{\hat{\mathbf{v}}^{ti}}}-\frac{1}{\sqrt{\mathbf{v}^{tj}}}} \\
\leq& \sum\limits_{t=1}^{T'}\sum\limits_{\text{client} \, i \in \mathcal{F}_t}\sum\limits_{\text{benign client} \, j}
\norm{\frac{1}{\sqrt{\hat{\mathbf{v}}^{ti}}}-\frac{1}{\sqrt{\mathbf{v}^{tj}}}}
+\sum\limits_{t=T'+1}^T\sum\limits_{\text{client} \, i \in \mathcal{F}_t}\sum\limits_{\text{benign client} \, j}
\frac{1}{\beta_2  v_{min}}   \norm{\sqrt{\hat{\mathbf{v}}^{ti}}-\sqrt{\mathbf{v}^{tj}}} \\
=&\sum\limits_{t=1}^{T'}\sum\limits_{\text{client} \, i \in \mathcal{F}_t}\sum\limits_{\text{benign client} \, j}
\norm{\frac{1}{\sqrt{\hat{\mathbf{v}}^{ti}}}-\frac{1}{\sqrt{\mathbf{v}^{tj}}}}+\frac{1}{\beta_2  v_{min}}
 \sum\limits_{t=T'+1}^T\sum\limits_{\text{client} \, i \in \mathcal{F}_t}\sum\limits_{\text{benign client} \, j}
\norm{\sqrt{\hat{\mathbf{v}}^{ti}}-\sqrt{\mathbf{v}^{tj}}} \\
\leq& \sum\limits_{t=1}^{T'}\sum\limits_{\text{client} \, i \in \mathcal{F}_t}\sum\limits_{\text{benign client} \, j}
\norm{\frac{1}{\sqrt{\hat{\mathbf{v}}^{ti}}}-\frac{1}{\sqrt{\mathbf{v}^{tj}}}}+\frac{C'_v}{\beta_2  v_{min}}.
\end{split}
\end{equation}
The first inequality is due to Eq. (\ref{eq:vb4}). The second inequality is due to Eq. (\ref{eq:va7}).
Let $C''_v=\sum\limits_{t=1}^{T'}\sum\limits_{\text{client} \, i \in \mathcal{F}_t}\sum\limits_{\text{benign client} \, j}
\norm{\frac{1}{\sqrt{\hat{\mathbf{v}}^{ti}}}-\frac{1}{\sqrt{\mathbf{v}^{tj}}}}+\frac{C'_v}{\beta_2  v_{min}}$. Then $C''_v$ is a positive constant  number, and
\begin{equation}\label{eq:vb6}
\begin{split}
\sum\limits_{t=1}^T\sum\limits_{\text{client} \, i \in \mathcal{F}_t}\sum\limits_{\text{benign client} \, j}
\norm{\frac{1}{\sqrt{\hat{\mathbf{v}}^{ti}}}-\frac{1}{\sqrt{\mathbf{v}^{tj}}}}&\leq C''_v.
\end{split}
\end{equation}
When $T\leq T'$,
\begin{equation}\label{eq:vb7}
\begin{split}
\sum\limits_{t=1}^{T}\sum\limits_{\text{client} \, i \in \mathcal{F}_t}\sum\limits_{\text{benign client} \, j}
\norm{\frac{1}{\sqrt{\hat{\mathbf{v}}^{ti}}}-\frac{1}{\sqrt{\mathbf{v}^{tj}}}}
&\leq\sum\limits_{t=1}^{T'}\sum\limits_{\text{client} \, i \in \mathcal{F}_t}\sum\limits_{\text{benign client} \, j}
\norm{\frac{1}{\sqrt{\hat{\mathbf{v}}^{ti}}}-\frac{1}{\sqrt{\mathbf{v}^{tj}}}}\\
&\leq C''_v.
\end{split}
\end{equation}
Thus Eq. (\ref{eq:vb6}) still holds.

The third part is to prove $\frac{\hat{\mathbf{m}}^{ti}}{\sqrt{\hat{\mathbf{v}}^{ti}}}$ is close to $\frac{\mathbf{m}^{tj}}{\sqrt{\mathbf{v}^{tj}}}$. We also focus on one component of $\frac{\hat{\mathbf{m}}^{ti}}{\sqrt{\hat{\mathbf{v}}^{ti}}}$ and $\frac{\mathbf{m}^{tj}}{\sqrt{\mathbf{v}^{tj}}}$, i.e., $\frac{\hat{m}^{ti}_k}{\sqrt{\hat{v}^{ti}_k}}$ and $\frac{m^{tj}_k }{\sqrt{v^{tj}_k}}$.
Let $\triangle m^t_k=\hat{m}^{ti}_k-m^{tj}_k$ and $\triangle v^t_k=\frac{1}{\sqrt{\hat{v}^{ti}_k}}-\frac{1}{\sqrt{v^{tj}_k}}$.
\begin{equation}\label{eq:u2}
\begin{split}
\left|\frac{\hat{m}^{ti}_k}{\sqrt{\hat{v}^{ti}_k}}-\frac{m^{tj}_k }{\sqrt{v^{tj}_k}}\right|
&=\left|\hat{m}^{ti}_k\frac{1}{\sqrt{\hat{v}^{ti}_k}}-m^{tj}_k\frac{1}{\sqrt{v^{tj}_k}}\right|\\
&= \left|(m^{tj}_k+\triangle m^t_k) (\frac{1}{\sqrt{v^{tj}_k}}+\triangle v^t_k)-m^{tj}_k  \frac{1}{\sqrt{v^{tj}_k}}\right| \\
&= \left|m^{tj}_k  \frac{1}{\sqrt{v^{tj}_k}}+m^{tj}_k \triangle v^t_k+\frac{1}{\sqrt{v^{tj}_k}} \triangle m^t_k+\triangle m^t_k \triangle v^t_k-m^{tj}_k  \frac{1}{\sqrt{v^{tj}_k}}\right|\\
&= \left|m^{tj}_k \triangle v^t_k+\frac{1}{\sqrt{v^{tj}_k}} \triangle m^t_k+\triangle m^t_k \triangle v^t_k\right| \\
&\leq \left|m^{tj}_k\right| \left|\triangle v^t_k\right|+\left|\frac{1}{\sqrt{v^{tj}_k}}\right| \left|\triangle m^t_k\right|+\left|\triangle m^t_k\right| \left|\triangle v^t_k\right|.
\end{split}
\end{equation}
When $t>T'$,
\begin{equation}\label{eq:u3}
\begin{split}
\left|\frac{1}{\sqrt{v^{tj}_k}}\right|
&= \frac{1}{\sqrt{v^{tj}_k}}\\
&= \frac{1}{\sqrt{\beta_2 \overline{v}^{t-1}_k+(1-\beta_2)  (g^{tj}_k)^2}}\\
&\leq \frac{1}{\sqrt{\beta_2 \overline{v}^{t-1}_k}} \\
&\leq \frac{1}{\sqrt{\beta_2  v_{min}}}.
\end{split}
\end{equation}
The second equality is due to the definition of $\frac{1}{\sqrt{v^{tj}_k}}$. The second inequality is due to Assumption 2.
By combining Eq. (\ref{eq:u2}) with Eq. (\ref{eq:u3}),
\begin{equation}\label{eq:u4}
\begin{split}
\left|\hat{m}^{ti}_k  \frac{1}{\sqrt{\hat{v}^{ti}_k}}-m^{tj}_k  \frac{1}{\sqrt{v^{tj}_k}}\right|\leq& \left|m^{tj}_k\right| \left|\triangle v^t_k\right|+\frac{1}{\sqrt{\beta_2  v_{min}}} \left|\triangle m^t_k\right|+\left|\triangle m^t_k\right| \left|\triangle v^t_k\right|\\
=& \left|m^{tj}_k\right| \left|\frac{1}{\sqrt{\hat{v}^{ti}_k}}-\frac{1}{\sqrt{v^{tj}_k}}\right|\\
&+\frac{1}{\sqrt{\beta_2  v_{min}}} \left|\hat{m}^{ti}_k-m^{tj}_k\right|+
\left|\hat{m}^{ti}_k-m^{tj}_k\right| \left|\frac{1}{\sqrt{\hat{v}^{ti}_k}}-\frac{1}{\sqrt{v^{tj}_k}}\right|.
\end{split}
\end{equation}
We combine all the components together:
\begin{equation}\label{eq:u5}
\begin{split}
&\norm{\frac{\hat{\mathbf{m}}^{ti}}{\sqrt{\hat{\mathbf{v}}^{ti}}}- \frac{\mathbf{m}^{tj}}{\sqrt{\mathbf{v}^{tj}}}}\\
\leq& \norm{\mathbf{m}^{tj}} \norm{\frac{1}{\sqrt{\hat{\mathbf{v}}^{ti}}}-\frac{1}{\sqrt{\mathbf{v}^{tj}}}}+
\frac{1}{\sqrt{\beta_2  v_{min}}} \norm{\hat{\mathbf{m}}^{ti}-\mathbf{m}^{tj}}+
\norm{\hat{\mathbf{m}}^{ti}-\mathbf{m}^{tj}} \norm{\frac{1}{\sqrt{\hat{\mathbf{v}}^{ti}}}-\frac{1}{\sqrt{\mathbf{v}^{tj}}}}.
\end{split}
\end{equation}
By adding round $1$ to round $T$, all clients in $\mathcal{F}_t$ and all benign clients together, when $T>T'$,
\begin{equation}\label{eq:u9}
\begin{split}
&\sum\limits_{t=1}^T\sum\limits_{\text{client} \, i \in \mathcal{F}_t}\sum\limits_{\text{benign client} \, j}
\norm{\frac{\hat{\mathbf{m}}^{ti}}{\sqrt{\hat{\mathbf{v}}^{ti}}}- \frac{\mathbf{m}^{tj}}{\sqrt{\mathbf{v}^{tj}}}}
\\
=& \sum\limits_{t=1}^{T'}\sum\limits_{\text{client} \, i \in \mathcal{F}_t}\sum\limits_{\text{benign client} \, j}
\norm{\frac{\hat{\mathbf{m}}^{ti}}{\sqrt{\hat{\mathbf{v}}^{ti}}}-\frac{\mathbf{m}^{tj}}{\sqrt{\mathbf{v}^{tj}}}}
+\sum\limits_{t=T'+1}^T\sum\limits_{\text{client} \, i \in \mathcal{F}_t}\sum\limits_{\text{benign client} \, j}
\norm{\frac{\hat{\mathbf{m}}^{ti}}{\sqrt{\hat{\mathbf{v}}^{ti}}}-\frac{\mathbf{m}^{tj}}{\sqrt{\mathbf{v}^{tj}}}}\\
\leq& \sum\limits_{t=1}^{T'}\sum\limits_{\text{client} \, i \in \mathcal{F}_t}\sum\limits_{\text{benign client} \, j}
\norm{\frac{\hat{\mathbf{m}}^{ti}}{\sqrt{\hat{\mathbf{v}}^{ti}}}-\frac{\mathbf{m}^{tj}}{\sqrt{\mathbf{v}^{tj}}}}\\
&+\sum\limits_{t=T'+1}^T\sum\limits_{\text{client} \, i \in \mathcal{F}_t}\sum\limits_{\text{benign client} \, j}
\left(\norm{\mathbf{m}^{tj}} \norm{\frac{1}{\sqrt{\hat{\mathbf{v}}^{ti}}}-\frac{1}{\sqrt{\mathbf{v}^{tj}}}}\right.\\
&\left.+\frac{1}{\sqrt{\beta_2  v_{min}}} \norm{\hat{\mathbf{m}}^{ti}-\mathbf{m}^{tj}}+
\norm{\hat{\mathbf{m}}^{ti}-\mathbf{m}^{tj}} \norm{\frac{1}{\sqrt{\hat{\mathbf{v}}^{ti}}}-\frac{1}{\sqrt{\mathbf{v}^{tj}}}}\right)\\
\leq& \sum\limits_{t=1}^{T'}\sum\limits_{\text{client} \, i \in \mathcal{F}_t}\sum\limits_{\text{benign client} \, j}
\norm{\frac{\hat{\mathbf{m}}^{ti}}{\sqrt{\hat{\mathbf{v}}^{ti}}}-\frac{\mathbf{m}^{tj}}{\sqrt{\mathbf{v}^{tj}}}}
+\sum\limits_{t=T'+1}^T\sum\limits_{\text{client} \, i \in \mathcal{F}_t}\sum\limits_{\text{benign client} \, j}
g_{max} \norm{\frac{1}{\sqrt{\hat{\mathbf{v}}^{ti}}}-\frac{1}{\sqrt{\mathbf{v}^{tj}}}} \\
&+\sum\limits_{t=T'+1}^T\sum\limits_{\text{client} \, i \in \mathcal{F}_t}\sum\limits_{\text{benign client} \, j}
\frac{1}{\sqrt{\beta_2  v_{min}}} \norm{\hat{\mathbf{m}}^{ti}-\mathbf{m}^{tj}}\\
&+\sum\limits_{t=T'+1}^T\sum\limits_{\text{client} \, i \in \mathcal{F}_t}\sum\limits_{\text{benign client} \, j}
\norm{\hat{\mathbf{m}}^{ti}-\mathbf{m}^{tj}} \norm{\frac{1}{\sqrt{\hat{\mathbf{v}}^{ti}}}-\frac{1}{\sqrt{\mathbf{v}^{tj}}}}.
\end{split}
\end{equation}
The first inequality is due to Eq. (\ref{eq:u5}). The second inequality follows from Lemma 2.
Since $\sum\limits_{t=T'+1}^T\sum\limits_{\text{client} \, i \in \mathcal{F}_t}\sum\limits_{\text{benign client} \, j}
\norm{\hat{\mathbf{m}}^{ti}-\mathbf{m}^{tj}}\leq C_m$ (Eq. (\ref{eq:m4})) and $\sum\limits_{t=T'+1}^T\sum\limits_{\text{client} \, i \in \mathcal{F}_t}\sum\limits_{\text{benign client} \, j}\norm{\frac{1}{\sqrt{\hat{\mathbf{v}}^{ti}}}-\frac{1}{\sqrt{\mathbf{v}^{tj}}}}\leq C''_v$ (Eq. (\ref{eq:vb6})), then
\begin{equation}\label{eq:u10}
\begin{split}
\sum\limits_{t=T'+1}^T\sum\limits_{\text{client} \, i \in \mathcal{F}_t}\sum\limits_{\text{benign client} \, j}
\norm{\hat{\mathbf{m}}^{ti}-\mathbf{m}^{tj}} \norm{\frac{1}{\sqrt{\hat{\mathbf{v}}^{ti}}}-\frac{1}{\sqrt{\mathbf{v}^{tj}}}}
\leq C_mC''_v.
\end{split}
\end{equation}
The inequality follows from Lemma 1.
We draw our attention back to Eq. (\ref{eq:u9}):
\begin{equation}\label{eq:u11}
\begin{split}
&\sum\limits_{t=1}^T\sum\limits_{\text{client} \, i \in \mathcal{F}_t}\sum\limits_{\text{benign client} \, j}
\norm{\frac{\hat{\mathbf{m}}^{ti}}{\sqrt{\hat{\mathbf{v}}^{ti}}}-\frac{\mathbf{m}^{tj}}{\sqrt{\mathbf{v}^{tj}}}}\\
\leq& \sum\limits_{t=1}^{T'}\sum\limits_{\text{client} \, i \in \mathcal{F}_t}\sum\limits_{\text{benign client} \, j}
\norm{\frac{\hat{\mathbf{m}}^{ti}}{\sqrt{\hat{\mathbf{v}}^{ti}}}-\frac{\mathbf{m}^{tj}}{\sqrt{\mathbf{v}^{tj}}}}
+\sum\limits_{t=T'+1}^T\sum\limits_{\text{client} \, i \in \mathcal{F}_t}\sum\limits_{\text{benign client} \, j}g_{max} \norm{\frac{1}{\sqrt{\hat{\mathbf{v}}^{ti}}}-\frac{1}{\sqrt{\mathbf{v}^{tj}}}}\\
&+\sum\limits_{t=T'+1}^T\sum\limits_{\text{client} \, i \in \mathcal{F}_t}\sum\limits_{\text{benign client} \, j}\frac{1}{\sqrt{\beta_2  v_{min}}} \norm{\hat{\mathbf{m}}^{ti}-\mathbf{m}^{tj}} \\
&+\sum\limits_{t=T'+1}^T\sum\limits_{\text{client} \, i \in \mathcal{F}_t}\sum\limits_{\text{benign client} \, j}\norm{\hat{\mathbf{m}}^{ti}-\mathbf{m}^{tj}} \norm{\frac{1}{\sqrt{\hat{\mathbf{v}}^{ti}}}-\frac{1}{\sqrt{\mathbf{v}^{tj}}}}\\
\leq& \sum\limits_{t=1}^{T'}\sum\limits_{\text{client} \, i \in \mathcal{F}_t}\sum\limits_{\text{benign client} \, j}
\norm{\frac{\hat{\mathbf{m}}^{ti}}{\sqrt{\hat{\mathbf{v}}^{ti}}}-\frac{\mathbf{m}^{tj}}{\sqrt{\mathbf{v}^{tj}}}}
+\sum\limits_{t=T'+1}^T\sum\limits_{\text{client} \, i \in \mathcal{F}_t}\sum\limits_{\text{benign client} \, j}g_{max} \norm{\frac{1}{\sqrt{\hat{\mathbf{v}}^{ti}}}-\frac{1}{\sqrt{\mathbf{v}^{tj}}}}\\
&+\sum\limits_{t=T'+1}^T\sum\limits_{\text{client} \, i \in \mathcal{F}_t}\sum\limits_{\text{benign client} \, j}\frac{1}{\sqrt{\beta_2  v_{min}}} \norm{\hat{\mathbf{m}}^{ti}-\mathbf{m}^{tj}}
+C_mC''_v\\
\leq& \sum\limits_{t=1}^{T'}\sum\limits_{\text{client} \, i \in \mathcal{F}_t}\sum\limits_{\text{benign client} \, j}
\norm{\frac{\hat{\mathbf{m}}^{ti}}{\sqrt{\hat{\mathbf{v}}^{ti}}}-\frac{\mathbf{m}^{tj}}{\sqrt{\mathbf{v}^{tj}}}}
+\frac{C_m}{\sqrt{\beta_2  v_{min}}}+g_{max}C''_v+C_mC''_v\\
\end{split}
\end{equation}
The first inequality is due to Eq. (\ref{eq:u9}). The second inequality follows from Eq. (\ref{eq:u10}). The third inequality is due to Eq. (\ref{eq:m4}) and Eq. (\ref{eq:vb6}).
Let $C_\theta=\sum\limits_{t=1}^{T'}\sum\limits_{\text{client} \, i \in \mathcal{F}_t}\sum\limits_{\text{benign client} \, j}
\norm{\frac{\hat{\mathbf{m}}^{ti}}{\sqrt{\hat{\mathbf{v}}^{ti}}}-\frac{\mathbf{m}^{tj}}{\sqrt{\mathbf{v}^{tj}}}}
+\frac{C_m}{\sqrt{\beta_2  v_{min}}}+g_{max}C''_v+C_mC''_v$. Then $C_\theta$ is a positive constant  number, and  \begin{equation}\label{eq:u12}
\begin{split}
\sum\limits_{t=1}^T\sum\limits_{\text{client} \, i \in \mathcal{F}_t}\sum\limits_{\text{benign client} \, j}
\norm{\frac{\hat{\mathbf{m}}^{ti}}{\sqrt{\hat{\mathbf{v}}^{ti}}}- \frac{\mathbf{m}^{tj}}{\sqrt{\mathbf{v}^{tj}}}}\leq C_\theta.
\end{split}
\end{equation}
When $T\leq T'$,
\begin{equation}\label{eq:u13}
\begin{split}
\sum\limits_{t=1}^T\sum\limits_{\text{client} \, i \in \mathcal{F}_t}\sum\limits_{\text{benign client} \, j}
\norm{\frac{\hat{\mathbf{m}}^{ti}}{\sqrt{\hat{\mathbf{v}}^{ti}}}- \frac{\mathbf{m}^{tj}}{\sqrt{\mathbf{v}^{tj}}}}
&\leq\sum\limits_{t=1}^{T'}\sum\limits_{\text{client} \, i \in \mathcal{F}_t}\sum\limits_{\text{benign client} \, j}
\norm{\frac{\hat{\mathbf{m}}^{ti}}{\sqrt{\hat{\mathbf{v}}^{ti}}}- \frac{\mathbf{m}^{tj}}{\sqrt{\mathbf{v}^{tj}}}}\\
&\leq C_\theta.
\end{split}
\end{equation}
Thus Eq. (\ref{eq:u12}) still holds.

The fourth part is to prove $\hat{\boldsymbol{\theta}}^{ti}$ is close to $\boldsymbol{\theta}^{tj}$. According to the definition of $\hat{\boldsymbol{\theta}}^{ti}$ and $\boldsymbol{\theta}^{tj}$,
\begin{equation}\label{eq:theta1}
\begin{split}
\sum\limits_{t=1}^T\sum\limits_{\text{client} \, i \in \mathcal{F}_t}\sum\limits_{\text{benign client} \, j}
\norm{\hat{\boldsymbol{\theta}}^{ti}-\boldsymbol{\theta}^{tj}}
&=\sum\limits_{t=1}^T\sum\limits_{\text{client} \, i \in \mathcal{F}_t}\sum\limits_{\text{benign client} \, j}
\norm{\overline{\boldsymbol{\theta}}^{t-1}-\frac{\hat{\mathbf{m}}^{ti}}{\sqrt{\hat{\mathbf{v}}^{ti}}}-\overline{\boldsymbol{\theta}}^{t-1}+\frac{\mathbf{m}^{tj}}{\sqrt{\mathbf{v}^{tj}}}}\\
&=\sum\limits_{t=1}^T\sum\limits_{\text{client} \, i \in \mathcal{F}_t}\sum\limits_{\text{benign client} \, j}
\norm{\frac{\hat{\mathbf{m}}^{ti}}{\sqrt{\hat{\mathbf{v}}^{ti}}}- \frac{\mathbf{m}^{tj}}{\sqrt{\mathbf{v}^{tj}}}} \\
&\leq C_\theta.
\end{split}
\end{equation}
The inequality follows from Eq. (\ref{eq:u12}).
Therefore, condition 3 of Adam-Byzantine resilience holds.

Since condition 1,2 and 3 of Adam-Byzantine resilience all hold, A-RFRS is Adam-Byzantine resilient.
\end{proof}

\section{FRS based on SGD with momentum}
In this section, we propose the definition of SGD with momentum-Byzantine resilience and show that our robust learning strategy is suitable in FRS based on SGD with momentum optimizer~\cite{conf/icml/SutskeverMDH13} with theoretical guarantee.

The algorithm of SGD with momentum of client $i$ at round $t$ is:
\begin{equation}\label{eq:mom_momentum}
\begin{split}
\mathbf{m}^{ti}&=\beta_3 \overline{\mathbf{m}}^{t-1}+\mathbf{g}^{ti} \\
\boldsymbol{\theta}^{ti}&=\overline{\boldsymbol{\theta}}^{t-1}-\eta \mathbf{m}^{ti}
\end{split}
\end{equation}
where $\mathbf{g}^{ti}$, $\mathbf{m}^{ti}$ and $\boldsymbol{\theta}^{ti}$ are gradient, momentum and model parameter of client $i$ at round $t$. $\eta$ is the learning rate. $\beta_3$ is a hyperparameter controlling the weight of momentum. $\overline{\mathbf{m}}^{t-1}$ and $\overline{\boldsymbol{\theta}}^{t-1}$ represent aggregated momentum and aggregated model parameter at round ($t-1$).

Suppose $\widetilde{n}$ out of $n$ clients are Byzantine.
Let $G^t=\{\mathbf{g}^{ti}|i\in\{1,...,n-\widetilde{n}\}\}$ be the gradient set of $n-\widetilde{n}$ benign clients at round $t$. Let  $\widetilde{G}^t=\{\widetilde{\mathbf{g}}^{ti}|i\in\{1,...,\widetilde{n}\}\}$ be the gradient set of $\widetilde{n}$ Byzantine clients at round $t$.
Let $\mathcal{F}_t$ be the set of selected clients for aggregation.
Let $\hat{G}^t=\{\hat{\mathbf{g}}^{ti}|\text{client} \, i\in\mathcal{F}_t\}$ be the gradient set of clients in $\mathcal{F}_t$.
We define SGD with momentum-Byzantine resilience as follows:

\textbf{Definition 2} SGD with momentum-Byzantine Resilience. For any client $i$ in $\mathcal{F}_t$, we denote its momentum and model parameter as $\hat{\mathbf{m}}^{ti}$ and $\hat{\boldsymbol{\theta}}^{ti}$. For any benign client $j$, we denote its momentum and model parameter as $\mathbf{m}^{tj}$ and $\boldsymbol{\theta}^{tj}$. A defense method is SGD with momentum-Byzantine resilient, if for the round $T$ there exists positive constant numbers $C_m$ and $C_\theta$, such that:
\begin{enumerate}
    \item $\sum\limits_{t=1}^T\sum\limits_{\text{client} \, i \in \mathcal{F}_t}\sum\limits_{\text{benign client} \, j}\norm{\hat{\mathbf{m}}^{ti}-\mathbf{m}^{tj}}\leq C_m$ ;
    \item $\sum\limits_{t=1}^T\sum\limits_{\text{client} \, i \in \mathcal{F}_t}\sum\limits_{\text{benign client} \, j}\norm{\hat{\boldsymbol{\theta}}^{ti}-\boldsymbol{\theta}^{tj}}\leq C_\theta$.
\end{enumerate}

We propose SGD with momentum-based robust federated recommendation system (S-RFRS), which utilize gradients to detect Byzantine clients in FRS based on SGD with momentum optimizer. The algorithm of S-RFRS is the same to A-RFRS except the former uses SGD with momentum to learn the model while the latter uses Adam to learn the model.

\textbf{Theorem 2}. S-RFRS is SGD with momentum-Byzantine resilient, if for any client $i$ in $\mathcal{F}_t$ with gradient $\hat{\mathbf{g}}^{ti} \in \hat{G}^t$, for any benign client $j$ with gradient $\mathbf{g}^{tj}\in G^t$, and for the round $T$, there exist a positive constant number $C_g$, such that
$$\sum\limits_{t=1}^T\sum\limits_{\text{client} \, i \in \mathcal{F}_t}\sum\limits_{\text{benign client} \, j}\norm{\hat{\mathbf{g}}^{ti}-\mathbf{g}^{tj}}\leq C_g.$$

\begin{proof}
Intuitively, since $\hat{\mathbf{g}}^{ti}$ is not far from $\mathbf{g}^{tj}$, $\hat{\mathbf{m}}^{ti}$ and $\hat{\boldsymbol{\theta}}^{ti}$ should also be close to $\mathbf{m}^{tj}$ and $\boldsymbol{\theta}^{tj}$. We will prove them step by step below.

The first step is to prove $\hat{\mathbf{m}}^{ti}$ is close to $\mathbf{m}^{tj}$.

According to the definition of momentum,
\begin{equation}\label{eq:mom_m1}
\begin{split}
\norm{\hat{\mathbf{m}}^{ti}-\mathbf{m}^{tj}}&=\norm{\beta_3  \overline{\mathbf{m}}^{t-1}+\hat{\mathbf{g}}^{ti}-(\beta_3  \overline{\mathbf{m}}^{t-1}+ \mathbf{g}^{tj})} \\
&=\norm{\hat{\mathbf{g}}^{ti}-\mathbf{g}^{tj}}.
\end{split}
\end{equation}
By adding round $1$ to round $T$, all clients in $\mathcal{F}_t$ and all benign clients together,
\begin{equation}\label{eq:mom_m3}
\begin{split}
\sum\limits_{t=1}^T\sum\limits_{\text{client} \, i \in \mathcal{F}_t}\sum\limits_{\text{benign client} \, j}
\norm{\hat{\mathbf{m}}^{ti}-\mathbf{m}^{tj}}
&=\sum\limits_{t=1}^T\sum\limits_{\text{client} \, i \in \mathcal{F}_t}\sum\limits_{\text{benign client} \, j}
\norm{\hat{\mathbf{g}}^{ti}-\mathbf{g}^{tj}}\\
&=\sum\limits_{t=1}^T\sum\limits_{\text{client} \, i \in \mathcal{F}_t}\sum\limits_{\text{benign client} \, j}
\norm{\hat{\mathbf{g}}^{ti}-\mathbf{g}^{tj}} \\
&\leq C_g.
\end{split}
\end{equation}
The first equlity is due to Eq. (\ref{eq:mom_m1}). The inequality is due to the assumption of Theorem 2.
Let $C_m=C_g$. Then $C_m$ is a positive constant  number, and
\begin{equation}\label{eq:mom_m4}
\begin{split}
\sum\limits_{t=1}^T\sum\limits_{\text{client} \, i \in \mathcal{F}_t}\sum\limits_{\text{benign client} \, j}\norm{\hat{\mathbf{m}}^{ti}-\mathbf{m}^{tj}}&\leq C_m.
\end{split}
\end{equation}
Thus, condition 1 of SGD with momentum-Byzantine resilience holds.

The second step is to prove $\hat{\boldsymbol{\theta}}^{ti}$ is close to $\boldsymbol{\theta}^{tj}$.

According to the definition of model parameter (for simplicity, we omit the learning rate),
\begin{equation}\label{eq:mom_theta1}
\begin{split}
\sum\limits_{t=1}^T\sum\limits_{\text{client} \, i \in \mathcal{F}_t}\sum\limits_{\text{benign client} \, j}
\norm{\hat{\boldsymbol{\theta}}^{ti}-\boldsymbol{\theta}^{tj}}
&=\sum\limits_{t=1}^T\sum\limits_{\text{client} \, i \in \mathcal{F}_t}\sum\limits_{\text{benign client} \, j}
\norm{\overline{\boldsymbol{\theta}}^{t-1}-\hat{\mathbf{m}}^{ti}-\overline{\boldsymbol{\theta}}^{t-1}+\mathbf{m}^{tj}} \\
&=\sum\limits_{t=1}^T\sum\limits_{\text{client} \, i \in \mathcal{F}_t}\sum\limits_{\text{benign client} \, j}\norm{\hat{\mathbf{m}}^{ti}-\mathbf{m}^{tj}}\\
&\leq C_m.
\end{split}
\end{equation}
The inequality is due to Eq. (\ref{eq:mom_m4}). Let $C_\theta=C_m$. Then $C_\theta$ is a positive constant  number, and
\begin{equation}\label{eq:mom_theta2}
\begin{split}
\sum\limits_{t=1}^T\sum\limits_{\text{client} \, i \in \mathcal{F}_t}\sum\limits_{\text{benign client} \, j}
\norm{\hat{\boldsymbol{\theta}}^{ti}-\boldsymbol{\theta}^{tj}}\leq C_\theta.
\end{split}
\end{equation}
Thus, condition 2 of SGD with momentum-Byzantine resilience holds.

Since condition 1 and 2 of SGD with momentum-Byzantine resilience all hold, S-RFRS is SGD with momentum-Byzantine resilient.
\end{proof}

\section{FRS based on AdaGrad}
In this section, we propose the definition of AdaGrad-Byzantine resilience and show that our robust learning strategy is suitable in FRS based on AdaGrad optimizer~\cite{journals/jmlr/DuchiHS11} with theoretical guarantee.

The algorithm of AdaGrad of client $i$ at round $t$ is:
\begin{equation}\label{eq:ag_adagrad}
\begin{split}
\mathbf{r}^{ti}&=\overline{\mathbf{r}}^{t-1}+\mathbf{g}^{ti}\odot \mathbf{g}^{ti}\\
\mathbf{u}^{ti}&=\frac{\mathbf{g}^{ti}}{\sqrt{\mathbf{r}^{ti}}+\epsilon}\\
\boldsymbol{\theta}^{ti}&=\overline{\boldsymbol{\theta}}^{t-1}-\eta  \mathbf{u}^{ti}
\end{split}
\end{equation}
where $\mathbf{g}^{ti}$, $\mathbf{r}^{ti}$, $\mathbf{u}^{ti}$ and $\boldsymbol{\theta}^{ti}$ are gradient, squared gradient, update, and model parameter of client $i$ at round $t$. $\eta$ is the learning rate. $\overline{\mathbf{r}}^{t-1}$ and $\overline{\boldsymbol{\theta}}^{t-1}$ represent aggregated squared gradient and aggregated model parameter at round ($t-1$). $\epsilon$ is a small constant for numerical stability.

Suppose $\widetilde{n}$ out of $n$ clients are Byzantine.
Let $G^t=\{\mathbf{g}^{ti}|i\in\{1,...,n-\widetilde{n}\}\}$ be the gradient set of $n-\widetilde{n}$ benign clients at round $t$. Let  $\widetilde{G}^t=\{\widetilde{\mathbf{g}}^{ti}|i\in\{1,...,\widetilde{n}\}\}$ be the gradient set of $\widetilde{n}$ Byzantine clients at round $t$.
Let $\mathcal{F}_t$ be the set of selected clients for aggregation.
Let $\hat{G}^t=\{\hat{\mathbf{g}}^{ti}|\text{client} \, i\in\mathcal{F}_t\}$ be the gradient set of clients in $\mathcal{F}_t$.
We define AdaGrad-Byzantine resilience as follows:

\textbf{Definition 3} AdaGrad-Byzantine Resilience.  For any client $i$ in $\mathcal{F}_t$, we denote its squared gradient and model parameter as $\hat{\mathbf{r}}^{ti}$ and $\hat{\boldsymbol{\theta}}^{ti}$. For any benign client $j$, we denote its squared gradient and model parameter as $\mathbf{r}^{tj}$ and $\boldsymbol{\theta}^{tj}$. A defense method is AdaGrad-Byzantine resilient, if for the round $T$ there exists positive constant numbers $C_r$ and $C_\theta$, such that:
\begin{enumerate}
    \item $\sum\limits_{t=1}^T\sum\limits_{\text{client} \, i \in \mathcal{F}_t}\sum\limits_{\text{benign client} \, j}\norm{\hat{\mathbf{r}}^{ti}-\mathbf{r}^{tj}}\leq C_r$ ;
    \item $\sum\limits_{t=1}^T\sum\limits_{\text{client} \, i \in \mathcal{F}_t}\sum\limits_{\text{benign client} \, j}\norm{\hat{\boldsymbol{\theta}}^{ti}-\boldsymbol{\theta}^{tj}}\leq C_\theta$.
\end{enumerate}

We propose AdaGrad-based robust federated recommendation system (AG-RFRS), which utilize gradients to detect Byzantine clients. The algorithm of AG-RFRS is the same to A-RFRS except the former uses AdaGrad to learn the model while the latter uses Adam to learn the model.

\textbf{Assumption 3}. For any gradient $\mathbf{g}$, its norm is upper bounded by a positive constant number $g_{max}$.
Formally, $\norm{\mathbf{g}}\leq g_{max}$ with $\mathbf{g}\in G^t\cup \widetilde{G}^t, t\in \mathbb{N}^*$.

\textbf{Assumption 4}. After $T'$ rounds of training, each component of $\overline{\mathbf{r}}^{t-1}$ is lower bounded by a positive constant number $r_{min}$. Formally, for any round $t$ with $t>T'$, $\overline{r}^{t-1}_k\geq r_{min}$, where $\overline{r}^{t-1}_k$ denotes the $k$-th component of $\overline{\mathbf{r}}^{t-1}$.

\textbf{Theorem 3}. AG-RFRS is AdaGrad-Byzantine resilient, if Assumption 3 and Assumption 4 hold, and for any client $i$ in $\mathcal{F}_t$ with gradient $\hat{\mathbf{g}}^{ti} \in \hat{G}^t$, for any benign client $j$ with gradient $\mathbf{g}^{tj}\in G^t$, and for the round $T$, there exist a positive constant number $C_g$, such that
$$\sum\limits_{t=1}^T\sum\limits_{\text{client} \, i \in \mathcal{F}_t}\sum\limits_{\text{benign client} \, j}\norm{\hat{\mathbf{g}}^{ti}-\mathbf{g}^{tj}}\leq C_g.$$

\begin{proof}
Intuitively, since $\hat{\mathbf{g}}^{ti}$ is not far from $\mathbf{g}^{tj}$, $\hat{\mathbf{r}}^{ti}$ and $\hat{\boldsymbol{\theta}}^{ti}$ should also be close to $\mathbf{r}^{tj}$ and $\boldsymbol{\theta}^{tj}$. We will prove them step by step below.

The first step is to prove $\hat{\mathbf{r}}^{ti}$ is close to $\mathbf{r}^{tj}$.
\begin{equation}\label{eq:ag_v1}
\begin{split}
\norm{\hat{\mathbf{r}}^{ti}-\mathbf{r}^{tj}}&=\norm{  \overline{\mathbf{r}}^{t-1}+  \hat{\mathbf{g}}^{ti}\odot \hat{\mathbf{g}}^{ti}-(  \overline{\mathbf{r}}^{t-1}+  \mathbf{g}^{tj}\odot \mathbf{g}^{tj})} \\
&=  \norm{\hat{\mathbf{g}}^{ti}\odot\hat{\mathbf{g}}^{ti} -\mathbf{g}^{tj}\odot \mathbf{g}^{tj}} \\
&=    \norm{(\hat{\mathbf{g}}^{ti}+\mathbf{g}^{tj})\odot(\hat{\mathbf{g}}^{ti}-\mathbf{g}^{tj})} \\
&\leq    \norm{\hat{\mathbf{g}}^{ti}+\mathbf{g}^{tj}}  \norm{\hat{\mathbf{g}}^{ti}-\mathbf{g}^{tj}} \\
&=    \norm{(\hat{\mathbf{g}}^{ti}-\mathbf{g}^{tj})+2  \mathbf{g}^{tj}}  \norm{\hat{\mathbf{g}}^{ti}-\mathbf{g}^{tj}}  \\
&\leq    \norm{\hat{\mathbf{g}}^{ti}-\mathbf{g}^{tj}}\left(\norm{\hat{\mathbf{g}}^{ti}-\mathbf{g}^{tj}}+2\norm{\mathbf{g}^{tj}}\right) \\
&=  \norm{\hat{\mathbf{g}}^{ti}-\mathbf{g}^{tj}}^2
+2 \norm{\hat{\mathbf{g}}^{ti}-\mathbf{g}^{tj}}\norm{\mathbf{g}^{tj}}.
\end{split}
\end{equation}
The first equality is due to the definition of squared gradient. The third equality follows from square of the difference formula. The first inequality is due to submultiplicativity of matrix norm. The second inequality is due to triangle inequality.
By adding round $1$ to round $T$, all clients in $\mathcal{F}_t$ and all benign clients together,
\begin{equation}\label{eq:ag_v2}
\begin{split}
&\sum\limits_{t=1}^T\sum\limits_{\text{client} \, i \in \mathcal{F}_t}\sum\limits_{\text{benign client} \, j}
\norm{\hat{\mathbf{r}}^{ti}-\mathbf{r}^{tj}}\\
\leq&
\sum\limits_{t=1}^T\sum\limits_{\text{client} \, i \in \mathcal{F}_t}\sum\limits_{\text{benign client} \, j}
\left( \norm{\hat{\mathbf{g}}^{ti}-\mathbf{g}^{tj}}^2
+2 \norm{\hat{\mathbf{g}}^{ti}-\mathbf{g}^{tj}}\norm{\mathbf{g}^{tj}}\right)\\
\leq&
\sum\limits_{t=1}^T\sum\limits_{\text{client} \, i \in \mathcal{F}_t}\sum\limits_{\text{benign client} \, j}
\norm{\hat{\mathbf{g}}^{ti}-\mathbf{g}^{tj}}^2\\
&+2\sum\limits_{t=1}^T\sum\limits_{\text{client} \, i \in \mathcal{F}_t}\sum\limits_{\text{benign client} \, j}
\norm{\hat{\mathbf{g}}^{ti}-\mathbf{g}^{tj}}g_{max} \\
\leq& C_g^2+2g_{max}C_g.
\end{split}
\end{equation}
The first inequality is due to Eq. (\ref{eq:ag_v1}). The second inequality follows from Assumption 3. The third inequality is due to the assumption of Theorem 3 and Lemma 1.
Let $C_r=C_g^2+2g_{max}C_g$. Then,  $C_r$ is a positive constant  number, and
\begin{equation}\label{eq:ag_v3}
\begin{split}
\sum\limits_{t=1}^T\sum\limits_{\text{client} \, i \in \mathcal{F}_t}\sum\limits_{\text{benign client} \, j}
\norm{\hat{\mathbf{r}}^{ti}-\mathbf{r}^{tj}}&\leq C_r.
\end{split}
\end{equation}
Therefore, condition 1 of AdaGrad-Byzantine resilience holds.

The second step is to prove $\hat{\boldsymbol{\theta}}^{ti}$ is close to $\boldsymbol{\theta}^{tj}$. According to the definition, $\hat{\boldsymbol{\theta}}^{ti}=\overline{\boldsymbol{\theta}}^{t-1}-\frac{\hat{\mathbf{g}}^{ti}}{\sqrt{\hat{\mathbf{r}}^{ti}}}$, $\boldsymbol{\theta}^{tj}=\overline{\boldsymbol{\theta}}^{t-1}-\frac{\mathbf{g}^{tj}}{\sqrt{\mathbf{r}^{tj}}}$ (for simplicity, we omit the learning rate).
We can decompose the proof into four parts:
\begin{enumerate}
    \item $\sqrt{\hat{\mathbf{r}}^{ti}}$ is close to $\sqrt{\mathbf{r}^{tj}}$
    \item $\frac{1}{\sqrt{\hat{\mathbf{r}}^{ti}}}$ is close to $\frac{1}{\sqrt{\mathbf{r}^{tj}}}$
    \item $\frac{\hat{\mathbf{g}}^{ti}}{\sqrt{\hat{\mathbf{r}}^{ti}}}$ is close to $\frac{\mathbf{g}^{tj}}{\sqrt{\mathbf{r}^{tj}}}$
    \item $\hat{\boldsymbol{\theta}}^{ti}$ is close to $\boldsymbol{\theta}^{tj}$.
\end{enumerate}
The first part is to prove $\sqrt{\hat{\mathbf{r}}^{ti}}$ is close to $\sqrt{\mathbf{r}^{tj}}$.
We focus on one component of $\sqrt{\hat{\mathbf{r}}^{ti}}$ and $\sqrt{\mathbf{r}^{tj}}$.
For any vector $\mathbf{r}$,
let $r_k$ denote the $k$-th component of $\mathbf{r}$.
\begin{equation}\label{eq:ag_va1}
\begin{split}
\left|\sqrt{\hat{r}^{ti}_k}-\sqrt{r^{tj}_k}\right|
&= \left|\frac{\hat{r}^{ti}_k-r^{tj}_k}{\sqrt{\hat{r}^{ti}_k}+\sqrt{r^{tj}_k}}\right| \\
&= \left|\hat{r}^{ti}_k-r^{tj}_k\right| \left|\frac{1}{\sqrt{\hat{r}^{ti}_k}+\sqrt{r^{tj}_k}}\right|.
\end{split}
\end{equation}
Recall that $\hat{r}^{ti}_k= \overline{r}^{t-1}_k+ (\hat{g}^{ti}_k)^2  $ and $r^{tj}_k= \overline{r}^{t-1}_k+  (g^{tj}_k)^2$.
When $t>T'$,
\begin{equation}\label{eq:ag_va2}
\begin{split}
\left|\frac{1}{\sqrt{\hat{r}^{ti}_k}+\sqrt{r^{tj}_k}}\right|&= \frac{1}{\sqrt{ \overline{r}^{t-1}_k+ (\hat{g}^{ti}_k)^2}+\sqrt{ \overline{r}^{t-1}_k+  (g^{tj}_k)^2}} \\
&\leq \frac{1}{\sqrt{ \overline{r}^{t-1}_k}+\sqrt{ \overline{r}^{t-1}_k}} \\
&\leq \frac{1}{2 \sqrt{  r_{min}}}.
\end{split}
\end{equation}
The second inequality is due to Assumption 4.
By combining Eq. (\ref{eq:ag_va1}) with Eq. (\ref{eq:ag_va2}),
\begin{equation}\label{eq:ag_va3}
\begin{split}
\left|\sqrt{\hat{r}^{ti}_k}-\sqrt{r^{tj}_k}\right|&\leq \frac{1}{2 \sqrt{   r_{min}}}   \left|\hat{r}^{ti}_k-r^{tj}_k\right|.
\end{split}
\end{equation}
We combine all the components together:
\begin{equation}\label{eq:ag_va4}
\begin{split}
\norm{\sqrt{\hat{\mathbf{r}}^{ti}}-\sqrt{\mathbf{r}^{tj}}}
&\leq \frac{1}{2 \sqrt{  r_{min}}}   \norm{\hat{\mathbf{r}}^{ti}-\mathbf{r}^{tj}}.
\end{split}
\end{equation}
By adding round $1$ to round $T$, all clients in $\mathcal{F}_t$ and all benign clients together, when $T>T'$,
\begin{equation}\label{eq:ag_va5}
\begin{split}
&\sum\limits_{t=1}^T\sum\limits_{\text{client} \, i \in \mathcal{F}_t}\sum\limits_{\text{benign client} \, j}
\norm{\sqrt{\hat{\mathbf{r}}^{ti}}-\sqrt{\mathbf{r}^{tj}}}\\
=& \sum\limits_{t=1}^{T'}\sum\limits_{\text{client} \, i \in \mathcal{F}_t}\sum\limits_{\text{benign client} \, j}
\norm{\sqrt{\hat{\mathbf{r}}^{ti}}-\sqrt{\mathbf{r}^{tj}}}+
\sum\limits_{t=T'+1}^T\sum\limits_{\text{client} \, i \in \mathcal{F}_t}\sum\limits_{\text{benign client} \, j}
\norm{\sqrt{\hat{\mathbf{r}}^{ti}}-\sqrt{\mathbf{r}^{tj}}} \\
\leq& \sum\limits_{t=1}^{T'}\sum\limits_{\text{client} \, i \in \mathcal{F}_t}\sum\limits_{\text{benign client} \, j}
\norm{\sqrt{\hat{\mathbf{r}}^{ti}}-\sqrt{\mathbf{r}^{tj}}}+\sum\limits_{t=T'+1}^T\sum\limits_{\text{client} \, i \in \mathcal{F}_t}\sum\limits_{\text{benign client} \, j}
\frac{1}{2  \sqrt{  r_{min}}}  \norm{\hat{\mathbf{r}}^{ti}-\mathbf{r}^{tj}} \\
=& \sum\limits_{t=1}^{T'}\sum\limits_{\text{client} \, i \in \mathcal{F}_t}\sum\limits_{\text{benign client} \, j}
\norm{\sqrt{\hat{\mathbf{r}}^{ti}}-\sqrt{\mathbf{r}^{tj}}}+\frac{1}{2  \sqrt{  r_{min}}}  \sum\limits_{t=T'+1}^T\sum\limits_{\text{client} \, i \in \mathcal{F}_t}\sum\limits_{\text{benign client} \, j}
\norm{\hat{\mathbf{r}}^{ti}-\mathbf{r}^{tj}} \\
\leq& \sum\limits_{t=1}^{T'}\sum\limits_{\text{client} \, i \in \mathcal{F}_t}\sum\limits_{\text{benign client} \, j}
\norm{\sqrt{\hat{\mathbf{r}}^{ti}}-\sqrt{\mathbf{r}^{tj}}}+\frac{C_r}{2  \sqrt{  r_{min}}}.
\end{split}
\end{equation}
The first inequality is due to Eq. (\ref{eq:ag_va4}). The second inequality is due to Eq. (\ref{eq:ag_v3}).
Let $C'_r=\sum\limits_{t=1}^{T'}\sum\limits_{\text{client} \, i \in \mathcal{F}_t}\sum\limits_{\text{benign client} \, j}
\norm{\sqrt{\hat{\mathbf{r}}^{ti}}-\sqrt{\mathbf{r}^{tj}}}+\frac{C_r}{2  \sqrt{  r_{min}}}$. Then $C'_r$ is a positive constant  number, and
\begin{equation}\label{eq:ag_va7}
\begin{split}
\sum\limits_{t=1}^T\sum\limits_{\text{client} \, i \in \mathcal{F}_t}\sum\limits_{\text{benign client} \, j}
\norm{\sqrt{\hat{\mathbf{r}}^{ti}}-\sqrt{\mathbf{r}^{tj}}}&\leq C'_r.
\end{split}
\end{equation}
When $T\leq T'$,
\begin{equation}\label{eq:ag_va8}
\begin{split}
\sum\limits_{t=1}^{T}\sum\limits_{\text{client} \, i \in \mathcal{F}_t}\sum\limits_{\text{benign client} \, j}
\norm{\sqrt{\hat{\mathbf{r}}^{ti}}-\sqrt{\mathbf{r}^{tj}}}
\leq&\sum\limits_{t=1}^{T'}\sum\limits_{\text{client} \, i \in \mathcal{F}_t}\sum\limits_{\text{benign client} \, j}
\norm{\sqrt{\hat{\mathbf{r}}^{ti}}-\sqrt{\mathbf{r}^{tj}}}\\
\leq& C'_r.
\end{split}
\end{equation}
Thus Eq. (\ref{eq:ag_va7}) still holds.

The second part is to prove $\frac{1}{\sqrt{\hat{\mathbf{r}}^{ti}}}$ is close to $\frac{1}{\sqrt{\mathbf{r}^{tj}}}$. Similar to $\sqrt{\hat{\mathbf{r}}^{ti}}$ and $\sqrt{\mathbf{r}^{tj}}$, we also focus on one components of $\frac{1}{\sqrt{\hat{\mathbf{r}}^{ti}}}$ and $\frac{1}{\sqrt{\mathbf{r}^{tj}}}$.
\begin{equation}\label{eq:ag_vb1}
\begin{split}
\left|\frac{1}{\sqrt{\hat{r}^{ti}_k}}-\frac{1}{\sqrt{r^{tj}_k}}\right|
&= \left|\frac{\sqrt{r^{tj}_k}-\sqrt{\hat{r}^{ti}_k}}{\sqrt{\hat{r}^{ti}_k}  \sqrt{r^{tj}_k}}\right| \\
&= \left|\sqrt{\hat{r}^{ti}_k}-\sqrt{r^{tj}_k}\right| \left|\frac{1}{\sqrt{\hat{r}^{ti}_k}  \sqrt{r^{tj}_k}}\right|.
\end{split}
\end{equation}
Recall that $\sqrt{\hat{r}^{ti}_k}=\sqrt{ \overline{r}^{t-1}_k+ (\hat{g}^{ti}_k)^2}$ and $\sqrt{r^{tj}_k}=\sqrt{ \overline{r}^{t-1}_k+  (g^{tj}_k)^2}$.
When $t>T'$,
\begin{equation}\label{eq:ag_vb2}
\begin{split}
\left|\frac{1}{\sqrt{\hat{r}^{ti}_k}  \sqrt{r^{tj}_k}}\right| &= \frac{1}{\sqrt{ \overline{r}^{t-1}_k+ (\hat{g}^{ti}_k)^2} \sqrt{ \overline{r}^{t-1}_k+  (g^{tj}_k)^2}} \\
&\leq \frac{1}{\sqrt{ \overline{r}^{t-1}_k} \sqrt{ \overline{r}^{t-1}_k}} \\
&\leq \frac{1}{\sqrt{  r_{min}} \sqrt{  r_{min}}} \\
&= \frac{1}{  r_{min}}.
\end{split}
\end{equation}
The second inequality is due to Assumption 4.
By combining Eq. (\ref{eq:ag_vb1}) with Eq. (\ref{eq:ag_vb2}),
\begin{equation}\label{eq:ag_vb3}
\begin{split}
\left|\frac{1}{\sqrt{\hat{r}^{ti}_k}}-\frac{1}{\sqrt{r^{tj}_k}}\right| &\leq \frac{1}{  r_{min}}  \left|\sqrt{\hat{r}^{ti}_k}-\sqrt{r^{tj}_k}\right|.
\end{split}
\end{equation}
We combine all the components together:
\begin{equation}\label{eq:ag_vb4}
\begin{split}
\norm{\frac{1}{\sqrt{\hat{\mathbf{r}}^{ti}}}-\frac{1}{\sqrt{\mathbf{r}^{tj}}}}&\leq \frac{1}{  r_{min}}   \norm{\sqrt{\hat{\mathbf{r}}^{ti}}-\sqrt{\mathbf{r}^{tj}}}.
\end{split}
\end{equation}
By adding round $1$ to round $T$, all clients in $\mathcal{F}_t$ and all benign clients together, when $T>T'$,
\begin{equation}\label{eq:ag_vb5}
\begin{split}
&\sum\limits_{t=1}^T\sum\limits_{\text{client} \, i \in \mathcal{F}_t}\sum\limits_{\text{benign client} \, j}
\norm{\frac{1}{\sqrt{\hat{\mathbf{r}}^{ti}}}-\frac{1}{\sqrt{\mathbf{r}^{tj}}}}\\
=&\sum\limits_{t=1}^{T'}\sum\limits_{\text{client} \, i \in \mathcal{F}_t}\sum\limits_{\text{benign client} \, j}
\norm{\frac{1}{\sqrt{\hat{\mathbf{r}}^{ti}}}-\frac{1}{\sqrt{\mathbf{r}^{tj}}}}
+\sum\limits_{t=T'+1}^T\sum\limits_{\text{client} \, i \in \mathcal{F}_t}\sum\limits_{\text{benign client} \, j}
\norm{\frac{1}{\sqrt{\hat{\mathbf{r}}^{ti}}}-\frac{1}{\sqrt{\mathbf{r}^{tj}}}} \\
\leq& \sum\limits_{t=1}^{T'}\sum\limits_{\text{client} \, i \in \mathcal{F}_t}\sum\limits_{\text{benign client} \, j}
\norm{\frac{1}{\sqrt{\hat{\mathbf{r}}^{ti}}}-\frac{1}{\sqrt{\mathbf{r}^{tj}}}}
+\sum\limits_{t=T'+1}^T\sum\limits_{\text{client} \, i \in \mathcal{F}_t}\sum\limits_{\text{benign client} \, j}
\frac{1}{  r_{min}}   \norm{\sqrt{\hat{\mathbf{r}}^{ti}}-\sqrt{\mathbf{r}^{tj}}} \\
=&\sum\limits_{t=1}^{T'}\sum\limits_{\text{client} \, i \in \mathcal{F}_t}\sum\limits_{\text{benign client} \, j}
\norm{\frac{1}{\sqrt{\hat{\mathbf{r}}^{ti}}}-\frac{1}{\sqrt{\mathbf{r}^{tj}}}}+\frac{1}{  r_{min}}
 \sum\limits_{t=T'+1}^T\sum\limits_{\text{client} \, i \in \mathcal{F}_t}\sum\limits_{\text{benign client} \, j}
\norm{\sqrt{\hat{\mathbf{r}}^{ti}}-\sqrt{\mathbf{r}^{tj}}} \\
\leq& \sum\limits_{t=1}^{T'}\sum\limits_{\text{client} \, i \in \mathcal{F}_t}\sum\limits_{\text{benign client} \, j}
\norm{\frac{1}{\sqrt{\hat{\mathbf{r}}^{ti}}}-\frac{1}{\sqrt{\mathbf{r}^{tj}}}}+\frac{C'_r}{  r_{min}}.
\end{split}
\end{equation}
The first inequality is due to Eq. (\ref{eq:ag_vb4}). The second inequality is due to Eq. (\ref{eq:ag_va7}).
Let $C''_r=\sum\limits_{t=1}^{T'}\sum\limits_{\text{client} \, i \in \mathcal{F}_t}\sum\limits_{\text{benign client} \, j}
\norm{\frac{1}{\sqrt{\hat{\mathbf{r}}^{ti}}}-\frac{1}{\sqrt{\mathbf{r}^{tj}}}}+\frac{C'_r}{  r_{min}}$. Then $C''_r$ is a positive constant  number, and
\begin{equation}\label{eq:ag_vb6}
\begin{split}
\sum\limits_{t=1}^T\sum\limits_{\text{client} \, i \in \mathcal{F}_t}\sum\limits_{\text{benign client} \, j}
\norm{\frac{1}{\sqrt{\hat{\mathbf{r}}^{ti}}}-\frac{1}{\sqrt{\mathbf{r}^{tj}}}}&\leq C''_r.
\end{split}
\end{equation}
When $T\leq T'$,
\begin{equation}\label{eq:ag_vb7}
\begin{split}
\sum\limits_{t=1}^{T}\sum\limits_{\text{client} \, i \in \mathcal{F}_t}\sum\limits_{\text{benign client} \, j}
\norm{\frac{1}{\sqrt{\hat{\mathbf{r}}^{ti}}}-\frac{1}{\sqrt{\mathbf{r}^{tj}}}}
&\leq\sum\limits_{t=1}^{T'}\sum\limits_{\text{client} \, i \in \mathcal{F}_t}\sum\limits_{\text{benign client} \, j}
\norm{\frac{1}{\sqrt{\hat{\mathbf{r}}^{ti}}}-\frac{1}{\sqrt{\mathbf{r}^{tj}}}}\\
&\leq C''_r.
\end{split}
\end{equation}
Thus Eq. (\ref{eq:ag_vb6}) still holds.

The third part is to prove $\frac{\hat{\mathbf{g}}^{ti}}{\sqrt{\hat{\mathbf{r}}^{ti}}}$ is close to $\frac{\mathbf{g}^{tj}}{\sqrt{\mathbf{r}^{tj}}}$. We also focus on one component of $\frac{\hat{\mathbf{g}}^{ti}}{\sqrt{\hat{\mathbf{r}}^{ti}}}$ and $\frac{\mathbf{g}^{tj}}{\sqrt{\mathbf{r}^{tj}}}$, i.e., $\frac{\hat{g}^{ti}_k}{\sqrt{\hat{r}^{ti}_k}}$ and $\frac{g^{tj}_k }{\sqrt{r^{tj}_k}}$.
Let $\triangle g^t_k=\hat{g}^{ti}_k-g^{tj}_k$ and $\triangle r^t_k=\frac{1}{\sqrt{\hat{r}^{ti}_k}}-\frac{1}{\sqrt{r^{tj}_k}}$.
\begin{equation}\label{eq:ag_u2}
\begin{split}
\left|\frac{\hat{g}^{ti}_k}{\sqrt{\hat{r}^{ti}_k}}-\frac{g^{tj}_k }{\sqrt{r^{tj}_k}}\right|
&=\left|\hat{g}^{ti}_k\frac{1}{\sqrt{\hat{r}^{ti}_k}}-g^{tj}_k\frac{1}{\sqrt{r^{tj}_k}}\right|\\
&= \left|(g^{tj}_k+\triangle g^t_k) (\frac{1}{\sqrt{r^{tj}_k}}+\triangle r^t_k)-g^{tj}_k  \frac{1}{\sqrt{r^{tj}_k}}\right| \\
&= \left|g^{tj}_k  \frac{1}{\sqrt{r^{tj}_k}}+g^{tj}_k \triangle r^t_k+\frac{1}{\sqrt{r^{tj}_k}} \triangle g^t_k+\triangle g^t_k \triangle r^t_k-g^{tj}_k  \frac{1}{\sqrt{r^{tj}_k}}\right|\\
&= \left|g^{tj}_k \triangle r^t_k+\frac{1}{\sqrt{r^{tj}_k}} \triangle g^t_k+\triangle g^t_k \triangle r^t_k\right| \\
&\leq \left|g^{tj}_k\right| \left|\triangle r^t_k\right|+\left|\frac{1}{\sqrt{r^{tj}_k}}\right| \left|\triangle g^t_k\right|+\left|\triangle g^t_k\right| \left|\triangle r^t_k\right|.
\end{split}
\end{equation}
When $t>T'$,
\begin{equation}\label{eq:ag_u3}
\begin{split}
\left|\frac{1}{\sqrt{r^{tj}_k}}\right|
&= \frac{1}{\sqrt{r^{tj}_k}}\\
&= \frac{1}{\sqrt{ \overline{r}^{t-1}_k+  (g^{tj}_k)^2}}\\
&\leq \frac{1}{\sqrt{ \overline{r}^{t-1}_k}} \\
&\leq \frac{1}{\sqrt{  r_{min}}}.
\end{split}
\end{equation}
The second equality is due to the definition of $\frac{1}{\sqrt{r^{tj}_k}}$. The second inequality is due to Assumption 4.
By combining Eq. (\ref{eq:ag_u2}) with Eq. (\ref{eq:ag_u3}),
\begin{equation}\label{eq:ag_u4}
\begin{split}
\left|\hat{g}^{ti}_k  \frac{1}{\sqrt{\hat{r}^{ti}_k}}-g^{tj}_k  \frac{1}{\sqrt{r^{tj}_k}}\right|\leq& \left|g^{tj}_k\right| \left|\triangle r^t_k\right|+\frac{1}{\sqrt{  r_{min}}} \left|\triangle g^t_k\right|+\left|\triangle g^t_k\right| \left|\triangle r^t_k\right|\\
=& \left|g^{tj}_k\right| \left|\frac{1}{\sqrt{\hat{r}^{ti}_k}}-\frac{1}{\sqrt{r^{tj}_k}}\right|\\
&+\frac{1}{\sqrt{  r_{min}}} \left|\hat{g}^{ti}_k-g^{tj}_k\right|+
\left|\hat{g}^{ti}_k-g^{tj}_k\right| \left|\frac{1}{\sqrt{\hat{r}^{ti}_k}}-\frac{1}{\sqrt{r^{tj}_k}}\right|.
\end{split}
\end{equation}
We combine all the components together:
\begin{equation}\label{eq:ag_u5}
\begin{split}
&\norm{\frac{\hat{\mathbf{g}}^{ti}}{\sqrt{\hat{\mathbf{r}}^{ti}}}- \frac{\mathbf{g}^{tj}}{\sqrt{\mathbf{r}^{tj}}}}\\
\leq& \norm{\mathbf{g}^{tj}} \norm{\frac{1}{\sqrt{\hat{\mathbf{r}}^{ti}}}-\frac{1}{\sqrt{\mathbf{r}^{tj}}}}+
\frac{1}{\sqrt{  r_{min}}} \norm{\hat{\mathbf{g}}^{ti}-\mathbf{g}^{tj}}+
\norm{\hat{\mathbf{g}}^{ti}-\mathbf{g}^{tj}} \norm{\frac{1}{\sqrt{\hat{\mathbf{r}}^{ti}}}-\frac{1}{\sqrt{\mathbf{r}^{tj}}}}.
\end{split}
\end{equation}
By adding round $1$ to round $T$, all clients in $\mathcal{F}_t$ and all benign clients together, when $T>T'$,
\begin{equation}\label{eq:ag_u9}
\begin{split}
&\sum\limits_{t=1}^T\sum\limits_{\text{client} \, i \in \mathcal{F}_t}\sum\limits_{\text{benign client} \, j}
\norm{\frac{\hat{\mathbf{g}}^{ti}}{\sqrt{\hat{\mathbf{r}}^{ti}}}- \frac{\mathbf{g}^{tj}}{\sqrt{\mathbf{r}^{tj}}}}
\\
=& \sum\limits_{t=1}^{T'}\sum\limits_{\text{client} \, i \in \mathcal{F}_t}\sum\limits_{\text{benign client} \, j}
\norm{\frac{\hat{\mathbf{g}}^{ti}}{\sqrt{\hat{\mathbf{r}}^{ti}}}-\frac{\mathbf{g}^{tj}}{\sqrt{\mathbf{r}^{tj}}}}
+\sum\limits_{t=T'+1}^T\sum\limits_{\text{client} \, i \in \mathcal{F}_t}\sum\limits_{\text{benign client} \, j}
\norm{\frac{\hat{\mathbf{g}}^{ti}}{\sqrt{\hat{\mathbf{r}}^{ti}}}-\frac{\mathbf{g}^{tj}}{\sqrt{\mathbf{r}^{tj}}}}\\
\leq& \sum\limits_{t=1}^{T'}\sum\limits_{\text{client} \, i \in \mathcal{F}_t}\sum\limits_{\text{benign client} \, j}
\norm{\frac{\hat{\mathbf{g}}^{ti}}{\sqrt{\hat{\mathbf{r}}^{ti}}}-\frac{\mathbf{g}^{tj}}{\sqrt{\mathbf{r}^{tj}}}}
+\sum\limits_{t=T'+1}^T\sum\limits_{\text{client} \, i \in \mathcal{F}_t}\sum\limits_{\text{benign client} \, j}
\left(\norm{\mathbf{g}^{tj}} \norm{\frac{1}{\sqrt{\hat{\mathbf{r}}^{ti}}}-\frac{1}{\sqrt{\mathbf{r}^{tj}}}}\right.\\
&\left.+\frac{1}{\sqrt{  r_{min}}} \norm{\hat{\mathbf{g}}^{ti}-\mathbf{g}^{tj}}+
\norm{\hat{\mathbf{g}}^{ti}-\mathbf{g}^{tj}} \norm{\frac{1}{\sqrt{\hat{\mathbf{r}}^{ti}}}-\frac{1}{\sqrt{\mathbf{r}^{tj}}}}\right)\\
\leq& \sum\limits_{t=1}^{T'}\sum\limits_{\text{client} \, i \in \mathcal{F}_t}\sum\limits_{\text{benign client} \, j}
\norm{\frac{\hat{\mathbf{g}}^{ti}}{\sqrt{\hat{\mathbf{r}}^{ti}}}-\frac{\mathbf{g}^{tj}}{\sqrt{\mathbf{r}^{tj}}}}
+\sum\limits_{t=T'+1}^T\sum\limits_{\text{client} \, i \in \mathcal{F}_t}\sum\limits_{\text{benign client} \, j}
g_{max} \norm{\frac{1}{\sqrt{\hat{\mathbf{r}}^{ti}}}-\frac{1}{\sqrt{\mathbf{r}^{tj}}}} \\
&+\sum\limits_{t=T'+1}^T\sum\limits_{\text{client} \, i \in \mathcal{F}_t}\sum\limits_{\text{benign client} \, j}
\frac{1}{\sqrt{  r_{min}}} \norm{\hat{\mathbf{g}}^{ti}-\mathbf{g}^{tj}}\\
&+\sum\limits_{t=T'+1}^T\sum\limits_{\text{client} \, i \in \mathcal{F}_t}\sum\limits_{\text{benign client} \, j}
\norm{\hat{\mathbf{g}}^{ti}-\mathbf{g}^{tj}} \norm{\frac{1}{\sqrt{\hat{\mathbf{r}}^{ti}}}-\frac{1}{\sqrt{\mathbf{r}^{tj}}}}.
\end{split}
\end{equation}
The first inequality is due to Eq. (\ref{eq:ag_u5}). The second inequality is due to Assumption 3.
Since $\sum\limits_{t=T'+1}^T\sum\limits_{\text{client} \, i \in \mathcal{F}_t}\sum\limits_{\text{benign client} \, j}
\norm{\hat{\mathbf{g}}^{ti}-\mathbf{g}^{tj}}\leq C_g$ (assumption of Theorem 3) and $\sum\limits_{t=T'+1}^T\sum\limits_{\text{client} \, i \in \mathcal{F}_t}\sum\limits_{\text{benign client} \, j}\norm{\frac{1}{\sqrt{\hat{\mathbf{r}}^{ti}}}-\frac{1}{\sqrt{\mathbf{r}^{tj}}}}\leq C''_r$ (Eq. (\ref{eq:ag_vb6})), then
\begin{equation}\label{eq:ag_u10}
\begin{split}
\sum\limits_{t=T'+1}^T\sum\limits_{\text{client} \, i \in \mathcal{F}_t}\sum\limits_{\text{benign client} \, j}
\norm{\hat{\mathbf{g}}^{ti}-\mathbf{g}^{tj}} \norm{\frac{1}{\sqrt{\hat{\mathbf{r}}^{ti}}}-\frac{1}{\sqrt{\mathbf{r}^{tj}}}}
\leq C_gC''_r.
\end{split}
\end{equation}
The inequality follows from Lemma 1.
We draw our attention back to Eq. (\ref{eq:ag_u9}):
\begin{equation}\label{eq:ag_u11}
\begin{split}
&\sum\limits_{t=1}^T\sum\limits_{\text{client} \, i \in \mathcal{F}_t}\sum\limits_{\text{benign client} \, j}
\norm{\frac{\hat{\mathbf{g}}^{ti}}{\sqrt{\hat{\mathbf{r}}^{ti}}}-\frac{\mathbf{g}^{tj}}{\sqrt{\mathbf{r}^{tj}}}}\\
\leq& \sum\limits_{t=1}^{T'}\sum\limits_{\text{client} \, i \in \mathcal{F}_t}\sum\limits_{\text{benign client} \, j}
\norm{\frac{\hat{\mathbf{g}}^{ti}}{\sqrt{\hat{\mathbf{r}}^{ti}}}-\frac{\mathbf{g}^{tj}}{\sqrt{\mathbf{r}^{tj}}}}
+\sum\limits_{t=T'+1}^T\sum\limits_{\text{client} \, i \in \mathcal{F}_t}\sum\limits_{\text{benign client} \, j}g_{max} \norm{\frac{1}{\sqrt{\hat{\mathbf{r}}^{ti}}}-\frac{1}{\sqrt{\mathbf{r}^{tj}}}}\\
&+\sum\limits_{t=T'+1}^T\sum\limits_{\text{client} \, i \in \mathcal{F}_t}\sum\limits_{\text{benign client} \, j}\frac{1}{\sqrt{  r_{min}}} \norm{\hat{\mathbf{g}}^{ti}-\mathbf{g}^{tj}} \\
&+\sum\limits_{t=T'+1}^T\sum\limits_{\text{client} \, i \in \mathcal{F}_t}\sum\limits_{\text{benign client} \, j}\norm{\hat{\mathbf{g}}^{ti}-\mathbf{g}^{tj}} \norm{\frac{1}{\sqrt{\hat{\mathbf{r}}^{ti}}}-\frac{1}{\sqrt{\mathbf{r}^{tj}}}}\\
\leq& \sum\limits_{t=1}^{T'}\sum\limits_{\text{client} \, i \in \mathcal{F}_t}\sum\limits_{\text{benign client} \, j}
\norm{\frac{\hat{\mathbf{g}}^{ti}}{\sqrt{\hat{\mathbf{r}}^{ti}}}-\frac{\mathbf{g}^{tj}}{\sqrt{\mathbf{r}^{tj}}}}
+\sum\limits_{t=T'+1}^T\sum\limits_{\text{client} \, i \in \mathcal{F}_t}\sum\limits_{\text{benign client} \, j}g_{max} \norm{\frac{1}{\sqrt{\hat{\mathbf{r}}^{ti}}}-\frac{1}{\sqrt{\mathbf{r}^{tj}}}}\\
&+\sum\limits_{t=T'+1}^T\sum\limits_{\text{client} \, i \in \mathcal{F}_t}\sum\limits_{\text{benign client} \, j}\frac{1}{\sqrt{  r_{min}}} \norm{\hat{\mathbf{g}}^{ti}-\mathbf{g}^{tj}}+C_gC''_r\\
\leq& \sum\limits_{t=1}^{T'}\sum\limits_{\text{client} \, i \in \mathcal{F}_t}\sum\limits_{\text{benign client} \, j}
\norm{\frac{\hat{\mathbf{g}}^{ti}}{\sqrt{\hat{\mathbf{r}}^{ti}}}-\frac{\mathbf{g}^{tj}}{\sqrt{\mathbf{r}^{tj}}}}
+\frac{C_g}{\sqrt{  r_{min}}}+g_{max}C''_r+C_gC''_r\\
\end{split}
\end{equation}
The first inequality is due to Eq. (\ref{eq:ag_u9}). The second inequality follows from Eq. (\ref{eq:ag_u10}). The third inequality is due to Eq. (\ref{eq:ag_vb6}) and the assumption of Theorem 3.
Let $C_\theta=\sum\limits_{t=1}^{T'}\sum\limits_{\text{client} \, i \in \mathcal{F}_t}\sum\limits_{\text{benign client} \, j}
\norm{\frac{\hat{\mathbf{g}}^{ti}}{\sqrt{\hat{\mathbf{r}}^{ti}}}-\frac{\mathbf{g}^{tj}}{\sqrt{\mathbf{r}^{tj}}}}
+\frac{C_g}{\sqrt{  r_{min}}}+g_{max}C''_r+C_gC''_r$. Then $C_\theta$ is a positive constant  number, and  \begin{equation}\label{eq:ag_u12}
\begin{split}
\sum\limits_{t=1}^T\sum\limits_{\text{client} \, i \in \mathcal{F}_t}\sum\limits_{\text{benign client} \, j}
\norm{\frac{\hat{\mathbf{g}}^{ti}}{\sqrt{\hat{\mathbf{r}}^{ti}}}- \frac{\mathbf{g}^{tj}}{\sqrt{\mathbf{r}^{tj}}}}\leq C_\theta.
\end{split}
\end{equation}
When $T\leq T'$,
\begin{equation}\label{eq:ag_u13}
\begin{split}
\sum\limits_{t=1}^T\sum\limits_{\text{client} \, i \in \mathcal{F}_t}\sum\limits_{\text{benign client} \, j}
\norm{\frac{\hat{\mathbf{g}}^{ti}}{\sqrt{\hat{\mathbf{r}}^{ti}}}- \frac{\mathbf{g}^{tj}}{\sqrt{\mathbf{r}^{tj}}}}
&\leq\sum\limits_{t=1}^{T'}\sum\limits_{\text{client} \, i \in \mathcal{F}_t}\sum\limits_{\text{benign client} \, j}
\norm{\frac{\hat{\mathbf{g}}^{ti}}{\sqrt{\hat{\mathbf{r}}^{ti}}}- \frac{\mathbf{g}^{tj}}{\sqrt{\mathbf{r}^{tj}}}}\\
&\leq C_\theta.
\end{split}
\end{equation}
Thus Eq. (\ref{eq:ag_u12}) still holds.

The fourth part is to prove $\hat{\boldsymbol{\theta}}^{ti}$ is close to $\boldsymbol{\theta}^{tj}$. According to the definition of $\hat{\boldsymbol{\theta}}^{ti}$ and $\boldsymbol{\theta}^{tj}$ (for simplicity, we omit the learning rate),
\begin{equation}\label{eq:ag_theta1}
\begin{split}
\sum\limits_{t=1}^T\sum\limits_{\text{client} \, i \in \mathcal{F}_t}\sum\limits_{\text{benign client} \, j}
\norm{\hat{\boldsymbol{\theta}}^{ti}-\boldsymbol{\theta}^{tj}}
&=\sum\limits_{t=1}^T\sum\limits_{\text{client} \, i \in \mathcal{F}_t}\sum\limits_{\text{benign client} \, j}
\norm{\overline{\boldsymbol{\theta}}^{t-1}-\frac{\hat{\mathbf{g}}^{ti}}{\sqrt{\hat{\mathbf{r}}^{ti}}}-\overline{\boldsymbol{\theta}}^{t-1}+\frac{\mathbf{g}^{tj}}{\sqrt{\mathbf{r}^{tj}}}}\\
&=\sum\limits_{t=1}^T\sum\limits_{\text{client} \, i \in \mathcal{F}_t}\sum\limits_{\text{benign client} \, j}
\norm{\frac{\hat{\mathbf{g}}^{ti}}{\sqrt{\hat{\mathbf{r}}^{ti}}}- \frac{\mathbf{g}^{tj}}{\sqrt{\mathbf{r}^{tj}}}} \\
&\leq C_\theta.
\end{split}
\end{equation}
The inequality follows from Eq. (\ref{eq:ag_u12}).
Therefore, condition 2 of AdaGrad-Byzantine resilience holds.

Since condition 1 and condition 2 of AdaGrad-Byzantine resilience all hold, AG-RFRS is AdaGrad-Byzantine resilient.
\end{proof}

\section{FRS based on RMSProp}
In this section, we propose the definition of RMSProp-Byzantine resilience and show that our robust learning strategy is suitable in FRS based on RMSProp optimizer~\cite{Tieleman2012} with theoretical guarantee.

The algorithm of RMSProp of client $i$ at round $t$ is:
\begin{equation}\label{eq:rms_rmsprop}
\begin{split}
\mathbf{r}^{ti}&=\beta_4 \overline{\mathbf{r}}^{t-1}+(1-\beta_4)  \mathbf{g}^{ti}\odot \mathbf{g}^{ti}\\
\mathbf{u}^{ti}&=\frac{\mathbf{g}^{ti}}{\sqrt{\mathbf{r}^{ti}}+\epsilon}\\
\boldsymbol{\theta}^{ti}&=\overline{\boldsymbol{\theta}}^{t-1}-\eta  \mathbf{u}^{ti}
\end{split}
\end{equation}
where $\mathbf{g}^{ti}$, $\mathbf{r}^{ti}$, $\mathbf{u}^{ti}$ and $\boldsymbol{\theta}^{ti}$ are gradient, squared gradient, update, and model parameter of client $i$ at round $t$. $\eta$ is the learning rate. $\beta_4$ is a hyperparameter controlling the weight of squared gradient. $\overline{\mathbf{r}}^{t-1}$ and $\overline{\boldsymbol{\theta}}^{t-1}$ represent aggregated squared gradient and aggregated model parameter at round ($t-1$). $\epsilon$ is a small constant for numerical stability.

Suppose $\widetilde{n}$ out of $n$ clients are Byzantine.
Let $G^t=\{\mathbf{g}^{ti}|i\in\{1,...,n-\widetilde{n}\}\}$ be the gradient set of $n-\widetilde{n}$ benign clients at round $t$. Let  $\widetilde{G}^t=\{\widetilde{\mathbf{g}}^{ti}|i\in\{1,...,\widetilde{n}\}\}$ be the gradient set of $\widetilde{n}$ Byzantine clients at round $t$.
Let $\mathcal{F}_t$ be the set of selected clients for aggregation.
Let $\hat{G}^t=\{\hat{\mathbf{g}}^{ti}|\text{client} \, i\in\mathcal{F}_t\}$ be the gradient set of clients in $\mathcal{F}_t$.
We define RMSProp-Byzantine resilience as follows:

\textbf{Definition 4} RMSProp-Byzantine Resilience.  For any client $i$ in $\mathcal{F}_t$, we denote its squared gradient and model parameter as $\hat{\mathbf{r}}^{ti}$ and $\hat{\boldsymbol{\theta}}^{ti}$. For any benign client $j$, we denote its squared gradient and model parameter as $\mathbf{r}^{tj}$ and $\boldsymbol{\theta}^{tj}$. A defense method is RMSProp-Byzantine resilient, if for the round $T$ there exists positive constant numbers $C_r$ and $C_\theta$, such that:
\begin{enumerate}
    \item $\sum\limits_{t=1}^T\sum\limits_{\text{client} \, i \in \mathcal{F}_t}\sum\limits_{\text{benign client} \, j}\norm{\hat{\mathbf{r}}^{ti}-\mathbf{r}^{tj}}\leq C_r$ ;
    \item $\sum\limits_{t=1}^T\sum\limits_{\text{client} \, i \in \mathcal{F}_t}\sum\limits_{\text{benign client} \, j}\norm{\hat{\boldsymbol{\theta}}^{ti}-\boldsymbol{\theta}^{tj}}\leq C_\theta$.
\end{enumerate}

We propose RMSProp-based robust federated recommendation system (R-RFRS), which utilize gradients to detect Byzantine clients. The algorithm of R-RFRS is the same to A-RFRS except the former uses RMSProp to learn the model while the latter uses Adam to learn the model.

\textbf{Assumption 5}. For any gradient $\mathbf{g}$, its norm is upper bounded by a positive constant number $g_{max}$.
Formally, $\norm{\mathbf{g}}\leq g_{max}$ with $\mathbf{g}\in G^t\cup \widetilde{G}^t, t\in \mathbb{N}^*$.

\textbf{Assumption 6}. After $T'$ rounds of training, each component of $\overline{\mathbf{r}}^{t-1}$ is lower bounded by a positive constant number $r_{min}$. Formally, for any round $t$ with $t>T'$, $\overline{r}^{t-1}_k\geq r_{min}$, where $\overline{r}^{t-1}_k$ denotes the $k$-th component of $\overline{\mathbf{r}}^{t-1}$.

\textbf{Theorem 4}. R-RFRS is RMSProp-Byzantine resilient, if Assumption 5 and Assumption 6 hold, and for any client $i$ in $\mathcal{F}_t$ with gradient $\hat{\mathbf{g}}^{ti} \in \hat{G}^t$, for any benign client $j$ with gradient $\mathbf{g}^{tj}\in G^t$, and for the round $T$, there exist a positive constant number $C_g$, such that
$$\sum\limits_{t=1}^T\sum\limits_{\text{client} \, i \in \mathcal{F}_t}\sum\limits_{\text{benign client} \, j}\norm{\hat{\mathbf{g}}^{ti}-\mathbf{g}^{tj}}\leq C_g.$$

\begin{proof}
Intuitively, since $\hat{\mathbf{g}}^{ti}$ is not far from $\mathbf{g}^{tj}$, $\hat{\mathbf{r}}^{ti}$ and $\hat{\boldsymbol{\theta}}^{ti}$ should also be close to $\mathbf{r}^{tj}$ and $\boldsymbol{\theta}^{tj}$. We will prove them step by step below.

The first step is to prove $\hat{\mathbf{r}}^{ti}$ is close to $\mathbf{r}^{tj}$.
\begin{equation}\label{eq:rms_v1}
\begin{split}
\norm{\hat{\mathbf{r}}^{ti}-\mathbf{r}^{tj}}&=\norm{\beta_4  \overline{\mathbf{r}}^{t-1}+(1-\beta_4)  \hat{\mathbf{g}}^{ti}\odot \hat{\mathbf{g}}^{ti}-(\beta_4  \overline{\mathbf{r}}^{t-1}+(1-\beta_4)  \mathbf{g}^{tj}\odot \mathbf{g}^{tj})} \\
&= (1-\beta_4) \norm{\hat{\mathbf{g}}^{ti}\odot\hat{\mathbf{g}}^{ti} -\mathbf{g}^{tj}\odot \mathbf{g}^{tj}} \\
&= (1-\beta_4)   \norm{(\hat{\mathbf{g}}^{ti}+\mathbf{g}^{tj})\odot(\hat{\mathbf{g}}^{ti}-\mathbf{g}^{tj})} \\
&\leq (1-\beta_4)   \norm{\hat{\mathbf{g}}^{ti}+\mathbf{g}^{tj}}  \norm{\hat{\mathbf{g}}^{ti}-\mathbf{g}^{tj}} \\
&= (1-\beta_4)   \norm{(\hat{\mathbf{g}}^{ti}-\mathbf{g}^{tj})+2  \mathbf{g}^{tj}}  \norm{\hat{\mathbf{g}}^{ti}-\mathbf{g}^{tj}}  \\
&\leq (1-\beta_4)   \norm{\hat{\mathbf{g}}^{ti}-\mathbf{g}^{tj}}\left(\norm{\hat{\mathbf{g}}^{ti}-\mathbf{g}^{tj}}+2\norm{\mathbf{g}^{tj}}\right) \\
&= (1-\beta_4) \norm{\hat{\mathbf{g}}^{ti}-\mathbf{g}^{tj}}^2
+2(1-\beta_4) \norm{\hat{\mathbf{g}}^{ti}-\mathbf{g}^{tj}}\norm{\mathbf{g}^{tj}}.
\end{split}
\end{equation}
The first equality is due to the definition of squared gradient. The second equality is due to absolutely homogeneous of matrix norm. The third equality follows from square of the difference formula. The first inequality is due to submultiplicativity of matrix norm. The second inequality is due to triangle inequality.
By adding round $1$ to round $T$, all clients in $\mathcal{F}_t$ and all benign clients together,
\begin{equation}\label{eq:rms_v2}
\begin{split}
&\sum\limits_{t=1}^T\sum\limits_{\text{client} \, i \in \mathcal{F}_t}\sum\limits_{\text{benign client} \, j}
\norm{\hat{\mathbf{r}}^{ti}-\mathbf{r}^{tj}}\\
\leq&
\sum\limits_{t=1}^T\sum\limits_{\text{client} \, i \in \mathcal{F}_t}\sum\limits_{\text{benign client} \, j}
\left((1-\beta_4) \norm{\hat{\mathbf{g}}^{ti}-\mathbf{g}^{tj}}^2
+2(1-\beta_4) \norm{\hat{\mathbf{g}}^{ti}-\mathbf{g}^{tj}}\norm{\mathbf{g}^{tj}}\right)\\
\leq&
(1-\beta_4)\sum\limits_{t=1}^T\sum\limits_{\text{client} \, i \in \mathcal{F}_t}\sum\limits_{\text{benign client} \, j}
\norm{\hat{\mathbf{g}}^{ti}-\mathbf{g}^{tj}}^2\\
&+2(1-\beta_4)\sum\limits_{t=1}^T\sum\limits_{\text{client} \, i \in \mathcal{F}_t}\sum\limits_{\text{benign client} \, j}
\norm{\hat{\mathbf{g}}^{ti}-\mathbf{g}^{tj}}g_{max} \\
\leq& (1-\beta_4)C_g^2+2(1-\beta_4)g_{max}C_g.
\end{split}
\end{equation}
The first inequality is due to Eq. (\ref{eq:rms_v1}). The second inequality follows from Assumption 5. The third inequality is due to the assumption of Theorem 4 and Lemma 1.
Let $C_r=(1-\beta_4)C_g^2+2(1-\beta_4)g_{max}C_g$. Then,  $C_r$ is a positive constant  number, and
\begin{equation}\label{eq:rms_v3}
\begin{split}
\sum\limits_{t=1}^T\sum\limits_{\text{client} \, i \in \mathcal{F}_t}\sum\limits_{\text{benign client} \, j}
\norm{\hat{\mathbf{r}}^{ti}-\mathbf{r}^{tj}}&\leq C_r.
\end{split}
\end{equation}
Therefore, condition 1 of RMSProp-Byzantine resilience holds.

The second step is to prove $\hat{\boldsymbol{\theta}}^{ti}$ is close to $\boldsymbol{\theta}^{tj}$. According to the definition, $\hat{\boldsymbol{\theta}}^{ti}=\overline{\boldsymbol{\theta}}^{t-1}-\frac{\hat{\mathbf{g}}^{ti}}{\sqrt{\hat{\mathbf{r}}^{ti}}}$, $\boldsymbol{\theta}^{tj}=\overline{\boldsymbol{\theta}}^{t-1}-\frac{\mathbf{g}^{tj}}{\sqrt{\mathbf{r}^{tj}}}$ (for simplicity, we omit the learning rate).
We can decompose the proof into four parts:
\begin{enumerate}
    \item $\sqrt{\hat{\mathbf{r}}^{ti}}$ is close to $\sqrt{\mathbf{r}^{tj}}$
    \item $\frac{1}{\sqrt{\hat{\mathbf{r}}^{ti}}}$ is close to $\frac{1}{\sqrt{\mathbf{r}^{tj}}}$
    \item $\frac{\hat{\mathbf{g}}^{ti}}{\sqrt{\hat{\mathbf{r}}^{ti}}}$ is close to $\frac{\mathbf{g}^{tj}}{\sqrt{\mathbf{r}^{tj}}}$
    \item $\hat{\boldsymbol{\theta}}^{ti}$ is close to $\boldsymbol{\theta}^{tj}$.
\end{enumerate}
The first part is to prove $\sqrt{\hat{\mathbf{r}}^{ti}}$ is close to $\sqrt{\mathbf{r}^{tj}}$.
We focus on one component of $\sqrt{\hat{\mathbf{r}}^{ti}}$ and $\sqrt{\mathbf{r}^{tj}}$.
For any vector $\mathbf{r}$,
let $r_k$ denote the $k$-th component of $\mathbf{r}$.
\begin{equation}\label{eq:rms_va1}
\begin{split}
\left|\sqrt{\hat{r}^{ti}_k}-\sqrt{r^{tj}_k}\right|
&= \left|\frac{\hat{r}^{ti}_k-r^{tj}_k}{\sqrt{\hat{r}^{ti}_k}+\sqrt{r^{tj}_k}}\right| \\
&= \left|\hat{r}^{ti}_k-r^{tj}_k\right| \left|\frac{1}{\sqrt{\hat{r}^{ti}_k}+\sqrt{r^{tj}_k}}\right|.
\end{split}
\end{equation}
Recall that $\hat{r}^{ti}_k=\beta_4 \overline{r}^{t-1}_k+(1-\beta_4) (\hat{g}^{ti}_k)^2  $ and $r^{tj}_k=\beta_4 \overline{r}^{t-1}_k+(1-\beta_4)  (g^{tj}_k)^2$.
When $t>T'$,
\begin{equation}\label{eq:rms_va2}
\begin{split}
\left|\frac{1}{\sqrt{\hat{r}^{ti}_k}+\sqrt{r^{tj}_k}}\right|&= \frac{1}{\sqrt{\beta_4 \overline{r}^{t-1}_k+(1-\beta_4) (\hat{g}^{ti}_k)^2}+\sqrt{\beta_4 \overline{r}^{t-1}_k+(1-\beta_4)  (g^{tj}_k)^2}} \\
&\leq \frac{1}{\sqrt{\beta_4 \overline{r}^{t-1}_k}+\sqrt{\beta_4 \overline{r}^{t-1}_k}} \\
&\leq \frac{1}{2 \sqrt{\beta_4  r_{min}}}.
\end{split}
\end{equation}
The second inequality is due to Assumption 6.
By combining Eq. (\ref{eq:rms_va1}) with Eq. (\ref{eq:rms_va2}),
\begin{equation}\label{eq:rms_va3}
\begin{split}
\left|\sqrt{\hat{r}^{ti}_k}-\sqrt{r^{tj}_k}\right|&\leq \frac{1}{2 \sqrt{ \beta_4  r_{min}}}   \left|\hat{r}^{ti}_k-r^{tj}_k\right|.
\end{split}
\end{equation}
We combine all the components together:
\begin{equation}\label{eq:rms_va4}
\begin{split}
\norm{\sqrt{\hat{\mathbf{r}}^{ti}}-\sqrt{\mathbf{r}^{tj}}}
&\leq \frac{1}{2 \sqrt{\beta_4  r_{min}}}   \norm{\hat{\mathbf{r}}^{ti}-\mathbf{r}^{tj}}.
\end{split}
\end{equation}
By adding round $1$ to round $T$, all clients in $\mathcal{F}_t$ and all benign clients together, when $T>T'$,
\begin{equation}\label{eq:rms_va5}
\begin{split}
&\sum\limits_{t=1}^T\sum\limits_{\text{client} \, i \in \mathcal{F}_t}\sum\limits_{\text{benign client} \, j}
\norm{\sqrt{\hat{\mathbf{r}}^{ti}}-\sqrt{\mathbf{r}^{tj}}}\\
=& \sum\limits_{t=1}^{T'}\sum\limits_{\text{client} \, i \in \mathcal{F}_t}\sum\limits_{\text{benign client} \, j}
\norm{\sqrt{\hat{\mathbf{r}}^{ti}}-\sqrt{\mathbf{r}^{tj}}}+
\sum\limits_{t=T'+1}^T\sum\limits_{\text{client} \, i \in \mathcal{F}_t}\sum\limits_{\text{benign client} \, j}
\norm{\sqrt{\hat{\mathbf{r}}^{ti}}-\sqrt{\mathbf{r}^{tj}}} \\
\leq& \sum\limits_{t=1}^{T'}\sum\limits_{\text{client} \, i \in \mathcal{F}_t}\sum\limits_{\text{benign client} \, j}
\norm{\sqrt{\hat{\mathbf{r}}^{ti}}-\sqrt{\mathbf{r}^{tj}}}+\sum\limits_{t=T'+1}^T\sum\limits_{\text{client} \, i \in \mathcal{F}_t}\sum\limits_{\text{benign client} \, j}
\frac{1}{2  \sqrt{\beta_4  r_{min}}}  \norm{\hat{\mathbf{r}}^{ti}-\mathbf{r}^{tj}} \\
=& \sum\limits_{t=1}^{T'}\sum\limits_{\text{client} \, i \in \mathcal{F}_t}\sum\limits_{\text{benign client} \, j}
\norm{\sqrt{\hat{\mathbf{r}}^{ti}}-\sqrt{\mathbf{r}^{tj}}}+\frac{1}{2  \sqrt{\beta_4  r_{min}}}  \sum\limits_{t=T'+1}^T\sum\limits_{\text{client} \, i \in \mathcal{F}_t}\sum\limits_{\text{benign client} \, j}
\norm{\hat{\mathbf{r}}^{ti}-\mathbf{r}^{tj}} \\
\leq& \sum\limits_{t=1}^{T'}\sum\limits_{\text{client} \, i \in \mathcal{F}_t}\sum\limits_{\text{benign client} \, j}
\norm{\sqrt{\hat{\mathbf{r}}^{ti}}-\sqrt{\mathbf{r}^{tj}}}+\frac{C_r}{2  \sqrt{\beta_4  r_{min}}}.
\end{split}
\end{equation}
The first inequality is due to Eq. (\ref{eq:rms_va4}). The second inequality is due to Eq. (\ref{eq:rms_v3}).
Let $C'_r=\sum\limits_{t=1}^{T'}\sum\limits_{\text{client} \, i \in \mathcal{F}_t}\sum\limits_{\text{benign client} \, j}
\norm{\sqrt{\hat{\mathbf{r}}^{ti}}-\sqrt{\mathbf{r}^{tj}}}+\frac{C_r}{2  \sqrt{\beta_4  r_{min}}}$. Then $C'_r$ is a positive constant  number, and
\begin{equation}\label{eq:rms_va7}
\begin{split}
\sum\limits_{t=1}^T\sum\limits_{\text{client} \, i \in \mathcal{F}_t}\sum\limits_{\text{benign client} \, j}
\norm{\sqrt{\hat{\mathbf{r}}^{ti}}-\sqrt{\mathbf{r}^{tj}}}&\leq C'_r.
\end{split}
\end{equation}
When $T\leq T'$,
\begin{equation}\label{eq:rms_va8}
\begin{split}
\sum\limits_{t=1}^{T}\sum\limits_{\text{client} \, i \in \mathcal{F}_t}\sum\limits_{\text{benign client} \, j}
\norm{\sqrt{\hat{\mathbf{r}}^{ti}}-\sqrt{\mathbf{r}^{tj}}}
\leq&\sum\limits_{t=1}^{T'}\sum\limits_{\text{client} \, i \in \mathcal{F}_t}\sum\limits_{\text{benign client} \, j}
\norm{\sqrt{\hat{\mathbf{r}}^{ti}}-\sqrt{\mathbf{r}^{tj}}}\\
\leq& C'_r.
\end{split}
\end{equation}
Thus Eq. (\ref{eq:rms_va7}) still holds.

The second part is to prove $\frac{1}{\sqrt{\hat{\mathbf{r}}^{ti}}}$ is close to $\frac{1}{\sqrt{\mathbf{r}^{tj}}}$. Similar to $\sqrt{\hat{\mathbf{r}}^{ti}}$ and $\sqrt{\mathbf{r}^{tj}}$, we also focus on one components of $\frac{1}{\sqrt{\hat{\mathbf{r}}^{ti}}}$ and $\frac{1}{\sqrt{\mathbf{r}^{tj}}}$.
\begin{equation}\label{eq:rms_vb1}
\begin{split}
\left|\frac{1}{\sqrt{\hat{r}^{ti}_k}}-\frac{1}{\sqrt{r^{tj}_k}}\right|
&= \left|\frac{\sqrt{r^{tj}_k}-\sqrt{\hat{r}^{ti}_k}}{\sqrt{\hat{r}^{ti}_k}  \sqrt{r^{tj}_k}}\right| \\
&= \left|\sqrt{\hat{r}^{ti}_k}-\sqrt{r^{tj}_k}\right| \left|\frac{1}{\sqrt{\hat{r}^{ti}_k}  \sqrt{r^{tj}_k}}\right|.
\end{split}
\end{equation}
Recall that $\sqrt{\hat{r}^{ti}_k}=\sqrt{\beta_4 \overline{r}^{t-1}_k+(1-\beta_4) (\hat{g}^{ti}_k)^2}$ and $\sqrt{r^{tj}_k}=\sqrt{\beta_4 \overline{r}^{t-1}_k+(1-\beta_4)  (g^{tj}_k)^2}$.
When $t>T'$,
\begin{equation}\label{eq:rms_vb2}
\begin{split}
\left|\frac{1}{\sqrt{\hat{r}^{ti}_k}  \sqrt{r^{tj}_k}}\right| &= \frac{1}{\sqrt{\beta_4 \overline{r}^{t-1}_k+(1-\beta_4) (\hat{g}^{ti}_k)^2} \sqrt{\beta_4 \overline{r}^{t-1}_k+(1-\beta_4)  (g^{tj}_k)^2}} \\
&\leq \frac{1}{\sqrt{\beta_4 \overline{r}^{t-1}_k} \sqrt{\beta_4 \overline{r}^{t-1}_k}} \\
&\leq \frac{1}{\sqrt{\beta_4  r_{min}} \sqrt{\beta_4  r_{min}}} \\
&= \frac{1}{\beta_4  r_{min}}.
\end{split}
\end{equation}
The second inequality is due to Assumption 6.
By combining Eq. (\ref{eq:rms_vb1}) with Eq. (\ref{eq:rms_vb2}),
\begin{equation}\label{eq:rms_vb3}
\begin{split}
\left|\frac{1}{\sqrt{\hat{r}^{ti}_k}}-\frac{1}{\sqrt{r^{tj}_k}}\right| &\leq \frac{1}{\beta_4  r_{min}}  \left|\sqrt{\hat{r}^{ti}_k}-\sqrt{r^{tj}_k}\right|.
\end{split}
\end{equation}
We combine all the components together:
\begin{equation}\label{eq:rms_vb4}
\begin{split}
\norm{\frac{1}{\sqrt{\hat{\mathbf{r}}^{ti}}}-\frac{1}{\sqrt{\mathbf{r}^{tj}}}}&\leq \frac{1}{\beta_4  r_{min}}   \norm{\sqrt{\hat{\mathbf{r}}^{ti}}-\sqrt{\mathbf{r}^{tj}}}.
\end{split}
\end{equation}
By adding round $1$ to round $T$, all clients in $\mathcal{F}_t$ and all benign clients together, when $T>T'$,
\begin{equation}\label{eq:rms_vb5}
\begin{split}
&\sum\limits_{t=1}^T\sum\limits_{\text{client} \, i \in \mathcal{F}_t}\sum\limits_{\text{benign client} \, j}
\norm{\frac{1}{\sqrt{\hat{\mathbf{r}}^{ti}}}-\frac{1}{\sqrt{\mathbf{r}^{tj}}}}\\
=&\sum\limits_{t=1}^{T'}\sum\limits_{\text{client} \, i \in \mathcal{F}_t}\sum\limits_{\text{benign client} \, j}
\norm{\frac{1}{\sqrt{\hat{\mathbf{r}}^{ti}}}-\frac{1}{\sqrt{\mathbf{r}^{tj}}}}
+\sum\limits_{t=T'+1}^T\sum\limits_{\text{client} \, i \in \mathcal{F}_t}\sum\limits_{\text{benign client} \, j}
\norm{\frac{1}{\sqrt{\hat{\mathbf{r}}^{ti}}}-\frac{1}{\sqrt{\mathbf{r}^{tj}}}} \\
\leq& \sum\limits_{t=1}^{T'}\sum\limits_{\text{client} \, i \in \mathcal{F}_t}\sum\limits_{\text{benign client} \, j}
\norm{\frac{1}{\sqrt{\hat{\mathbf{r}}^{ti}}}-\frac{1}{\sqrt{\mathbf{r}^{tj}}}}
+\sum\limits_{t=T'+1}^T\sum\limits_{\text{client} \, i \in \mathcal{F}_t}\sum\limits_{\text{benign client} \, j}
\frac{1}{\beta_4  r_{min}}   \norm{\sqrt{\hat{\mathbf{r}}^{ti}}-\sqrt{\mathbf{r}^{tj}}} \\
=&\sum\limits_{t=1}^{T'}\sum\limits_{\text{client} \, i \in \mathcal{F}_t}\sum\limits_{\text{benign client} \, j}
\norm{\frac{1}{\sqrt{\hat{\mathbf{r}}^{ti}}}-\frac{1}{\sqrt{\mathbf{r}^{tj}}}}+\frac{1}{\beta_4  r_{min}}
 \sum\limits_{t=T'+1}^T\sum\limits_{\text{client} \, i \in \mathcal{F}_t}\sum\limits_{\text{benign client} \, j}
\norm{\sqrt{\hat{\mathbf{r}}^{ti}}-\sqrt{\mathbf{r}^{tj}}} \\
\leq& \sum\limits_{t=1}^{T'}\sum\limits_{\text{client} \, i \in \mathcal{F}_t}\sum\limits_{\text{benign client} \, j}
\norm{\frac{1}{\sqrt{\hat{\mathbf{r}}^{ti}}}-\frac{1}{\sqrt{\mathbf{r}^{tj}}}}+\frac{C'_r}{\beta_4  r_{min}}.
\end{split}
\end{equation}
The first inequality is due to Eq. (\ref{eq:rms_vb4}). The second inequality is due to Eq. (\ref{eq:rms_va7}).
Let $C''_r=\sum\limits_{t=1}^{T'}\sum\limits_{\text{client} \, i \in \mathcal{F}_t}\sum\limits_{\text{benign client} \, j}
\norm{\frac{1}{\sqrt{\hat{\mathbf{r}}^{ti}}}-\frac{1}{\sqrt{\mathbf{r}^{tj}}}}+\frac{C'_r}{\beta_4  r_{min}}$. Then $C''_r$ is a positive constant  number, and
\begin{equation}\label{eq:rms_vb6}
\begin{split}
\sum\limits_{t=1}^T\sum\limits_{\text{client} \, i \in \mathcal{F}_t}\sum\limits_{\text{benign client} \, j}
\norm{\frac{1}{\sqrt{\hat{\mathbf{r}}^{ti}}}-\frac{1}{\sqrt{\mathbf{r}^{tj}}}}&\leq C''_r.
\end{split}
\end{equation}
When $T\leq T'$,
\begin{equation}\label{eq:rms_vb7}
\begin{split}
\sum\limits_{t=1}^{T}\sum\limits_{\text{client} \, i \in \mathcal{F}_t}\sum\limits_{\text{benign client} \, j}
\norm{\frac{1}{\sqrt{\hat{\mathbf{r}}^{ti}}}-\frac{1}{\sqrt{\mathbf{r}^{tj}}}}
&\leq\sum\limits_{t=1}^{T'}\sum\limits_{\text{client} \, i \in \mathcal{F}_t}\sum\limits_{\text{benign client} \, j}
\norm{\frac{1}{\sqrt{\hat{\mathbf{r}}^{ti}}}-\frac{1}{\sqrt{\mathbf{r}^{tj}}}}\\
&\leq C''_r.
\end{split}
\end{equation}
Thus Eq. (\ref{eq:rms_vb6}) still holds.

The third part is to prove $\frac{\hat{\mathbf{g}}^{ti}}{\sqrt{\hat{\mathbf{r}}^{ti}}}$ is close to $\frac{\mathbf{g}^{tj}}{\sqrt{\mathbf{r}^{tj}}}$. We also focus on one component of $\frac{\hat{\mathbf{g}}^{ti}}{\sqrt{\hat{\mathbf{r}}^{ti}}}$ and $\frac{\mathbf{g}^{tj}}{\sqrt{\mathbf{r}^{tj}}}$, i.e., $\frac{\hat{g}^{ti}_k}{\sqrt{\hat{r}^{ti}_k}}$ and $\frac{g^{tj}_k }{\sqrt{r^{tj}_k}}$.
Let $\triangle g^t_k=\hat{g}^{ti}_k-g^{tj}_k$ and $\triangle r^t_k=\frac{1}{\sqrt{\hat{r}^{ti}_k}}-\frac{1}{\sqrt{r^{tj}_k}}$.
\begin{equation}\label{eq:rms_u2}
\begin{split}
\left|\frac{\hat{g}^{ti}_k}{\sqrt{\hat{r}^{ti}_k}}-\frac{g^{tj}_k }{\sqrt{r^{tj}_k}}\right|
&=\left|\hat{g}^{ti}_k\frac{1}{\sqrt{\hat{r}^{ti}_k}}-g^{tj}_k\frac{1}{\sqrt{r^{tj}_k}}\right|\\
&= \left|(g^{tj}_k+\triangle g^t_k) (\frac{1}{\sqrt{r^{tj}_k}}+\triangle r^t_k)-g^{tj}_k  \frac{1}{\sqrt{r^{tj}_k}}\right| \\
&= \left|g^{tj}_k  \frac{1}{\sqrt{r^{tj}_k}}+g^{tj}_k \triangle r^t_k+\frac{1}{\sqrt{r^{tj}_k}} \triangle g^t_k+\triangle g^t_k \triangle r^t_k-g^{tj}_k  \frac{1}{\sqrt{r^{tj}_k}}\right|\\
&= \left|g^{tj}_k \triangle r^t_k+\frac{1}{\sqrt{r^{tj}_k}} \triangle g^t_k+\triangle g^t_k \triangle r^t_k\right| \\
&\leq \left|g^{tj}_k\right| \left|\triangle r^t_k\right|+\left|\frac{1}{\sqrt{r^{tj}_k}}\right| \left|\triangle g^t_k\right|+\left|\triangle g^t_k\right| \left|\triangle r^t_k\right|.
\end{split}
\end{equation}
When $t>T'$,
\begin{equation}\label{eq:rms_u3}
\begin{split}
\left|\frac{1}{\sqrt{r^{tj}_k}}\right|
&= \frac{1}{\sqrt{r^{tj}_k}}\\
&= \frac{1}{\sqrt{\beta_4 \overline{r}^{t-1}_k+(1-\beta_4)  (g^{tj}_k)^2}}\\
&\leq \frac{1}{\sqrt{\beta_4 \overline{r}^{t-1}_k}} \\
&\leq \frac{1}{\sqrt{\beta_4  r_{min}}}.
\end{split}
\end{equation}
The second equality is due to the definition of $\frac{1}{\sqrt{r^{tj}_k}}$. The second inequality is due to Assumption 6.
By combining Eq. (\ref{eq:rms_u2}) with Eq. (\ref{eq:rms_u3}),
\begin{equation}\label{eq:rms_u4}
\begin{split}
\left|\hat{g}^{ti}_k  \frac{1}{\sqrt{\hat{r}^{ti}_k}}-g^{tj}_k  \frac{1}{\sqrt{r^{tj}_k}}\right|\leq& \left|g^{tj}_k\right| \left|\triangle r^t_k\right|+\frac{1}{\sqrt{\beta_4  r_{min}}} \left|\triangle g^t_k\right|+\left|\triangle g^t_k\right| \left|\triangle r^t_k\right|\\
=& \left|g^{tj}_k\right| \left|\frac{1}{\sqrt{\hat{r}^{ti}_k}}-\frac{1}{\sqrt{r^{tj}_k}}\right|\\
&+\frac{1}{\sqrt{\beta_4  r_{min}}} \left|\hat{g}^{ti}_k-g^{tj}_k\right|+
\left|\hat{g}^{ti}_k-g^{tj}_k\right| \left|\frac{1}{\sqrt{\hat{r}^{ti}_k}}-\frac{1}{\sqrt{r^{tj}_k}}\right|.
\end{split}
\end{equation}
We combine all the components together:
\begin{equation}\label{eq:rms_u5}
\begin{split}
&\norm{\frac{\hat{\mathbf{g}}^{ti}}{\sqrt{\hat{\mathbf{r}}^{ti}}}- \frac{\mathbf{g}^{tj}}{\sqrt{\mathbf{r}^{tj}}}}\\
\leq& \norm{\mathbf{g}^{tj}} \norm{\frac{1}{\sqrt{\hat{\mathbf{r}}^{ti}}}-\frac{1}{\sqrt{\mathbf{r}^{tj}}}}+
\frac{1}{\sqrt{\beta_4  r_{min}}} \norm{\hat{\mathbf{g}}^{ti}-\mathbf{g}^{tj}}+
\norm{\hat{\mathbf{g}}^{ti}-\mathbf{g}^{tj}} \norm{\frac{1}{\sqrt{\hat{\mathbf{r}}^{ti}}}-\frac{1}{\sqrt{\mathbf{r}^{tj}}}}.
\end{split}
\end{equation}
By adding round $1$ to round $T$, all clients in $\mathcal{F}_t$ and all benign clients together, when $T>T'$,
\begin{equation}\label{eq:rms_u9}
\begin{split}
&\sum\limits_{t=1}^T\sum\limits_{\text{client} \, i \in \mathcal{F}_t}\sum\limits_{\text{benign client} \, j}
\norm{\frac{\hat{\mathbf{g}}^{ti}}{\sqrt{\hat{\mathbf{r}}^{ti}}}- \frac{\mathbf{g}^{tj}}{\sqrt{\mathbf{r}^{tj}}}}
\\
=& \sum\limits_{t=1}^{T'}\sum\limits_{\text{client} \, i \in \mathcal{F}_t}\sum\limits_{\text{benign client} \, j}
\norm{\frac{\hat{\mathbf{g}}^{ti}}{\sqrt{\hat{\mathbf{r}}^{ti}}}-\frac{\mathbf{g}^{tj}}{\sqrt{\mathbf{r}^{tj}}}}
+\sum\limits_{t=T'+1}^T\sum\limits_{\text{client} \, i \in \mathcal{F}_t}\sum\limits_{\text{benign client} \, j}
\norm{\frac{\hat{\mathbf{g}}^{ti}}{\sqrt{\hat{\mathbf{r}}^{ti}}}-\frac{\mathbf{g}^{tj}}{\sqrt{\mathbf{r}^{tj}}}}\\
\leq& \sum\limits_{t=1}^{T'}\sum\limits_{\text{client} \, i \in \mathcal{F}_t}\sum\limits_{\text{benign client} \, j}
\norm{\frac{\hat{\mathbf{g}}^{ti}}{\sqrt{\hat{\mathbf{r}}^{ti}}}-\frac{\mathbf{g}^{tj}}{\sqrt{\mathbf{r}^{tj}}}}
+\sum\limits_{t=T'+1}^T\sum\limits_{\text{client} \, i \in \mathcal{F}_t}\sum\limits_{\text{benign client} \, j}
\left(\norm{\mathbf{g}^{tj}} \norm{\frac{1}{\sqrt{\hat{\mathbf{r}}^{ti}}}-\frac{1}{\sqrt{\mathbf{r}^{tj}}}}\right.\\
&\left.+\frac{1}{\sqrt{\beta_4  r_{min}}} \norm{\hat{\mathbf{g}}^{ti}-\mathbf{g}^{tj}}+
\norm{\hat{\mathbf{g}}^{ti}-\mathbf{g}^{tj}} \norm{\frac{1}{\sqrt{\hat{\mathbf{r}}^{ti}}}-\frac{1}{\sqrt{\mathbf{r}^{tj}}}}\right)\\
\leq& \sum\limits_{t=1}^{T'}\sum\limits_{\text{client} \, i \in \mathcal{F}_t}\sum\limits_{\text{benign client} \, j}
\norm{\frac{\hat{\mathbf{g}}^{ti}}{\sqrt{\hat{\mathbf{r}}^{ti}}}-\frac{\mathbf{g}^{tj}}{\sqrt{\mathbf{r}^{tj}}}}
+\sum\limits_{t=T'+1}^T\sum\limits_{\text{client} \, i \in \mathcal{F}_t}\sum\limits_{\text{benign client} \, j}
g_{max} \norm{\frac{1}{\sqrt{\hat{\mathbf{r}}^{ti}}}-\frac{1}{\sqrt{\mathbf{r}^{tj}}}} \\
&+\sum\limits_{t=T'+1}^T\sum\limits_{\text{client} \, i \in \mathcal{F}_t}\sum\limits_{\text{benign client} \, j}
\frac{1}{\sqrt{\beta_4  r_{min}}} \norm{\hat{\mathbf{g}}^{ti}-\mathbf{g}^{tj}}\\
&+\sum\limits_{t=T'+1}^T\sum\limits_{\text{client} \, i \in \mathcal{F}_t}\sum\limits_{\text{benign client} \, j}
\norm{\hat{\mathbf{g}}^{ti}-\mathbf{g}^{tj}} \norm{\frac{1}{\sqrt{\hat{\mathbf{r}}^{ti}}}-\frac{1}{\sqrt{\mathbf{r}^{tj}}}}.
\end{split}
\end{equation}
The first inequality is due to Eq. (\ref{eq:rms_u5}). The second inequality is due to Assumption 5.
Since $\sum\limits_{t=T'+1}^T\sum\limits_{\text{client} \, i \in \mathcal{F}_t}\sum\limits_{\text{benign client} \, j}
\norm{\hat{\mathbf{g}}^{ti}-\mathbf{g}^{tj}}\leq C_g$ (assumption of Theorem 4) and $\sum\limits_{t=T'+1}^T\sum\limits_{\text{client} \, i \in \mathcal{F}_t}\sum\limits_{\text{benign client} \, j}\norm{\frac{1}{\sqrt{\hat{\mathbf{r}}^{ti}}}-\frac{1}{\sqrt{\mathbf{r}^{tj}}}}\leq C''_r$ (Eq. (\ref{eq:rms_vb6})), then
\begin{equation}\label{eq:rms_u10}
\begin{split}
\sum\limits_{t=T'+1}^T\sum\limits_{\text{client} \, i \in \mathcal{F}_t}\sum\limits_{\text{benign client} \, j}
\norm{\hat{\mathbf{g}}^{ti}-\mathbf{g}^{tj}} \norm{\frac{1}{\sqrt{\hat{\mathbf{r}}^{ti}}}-\frac{1}{\sqrt{\mathbf{r}^{tj}}}}
\leq C_gC''_r.
\end{split}
\end{equation}
The inequality follows from Lemma 1.
We draw our attention back to Eq. (\ref{eq:rms_u9}):
\begin{equation}\label{eq:rms_u11}
\begin{split}
&\sum\limits_{t=1}^T\sum\limits_{\text{client} \, i \in \mathcal{F}_t}\sum\limits_{\text{benign client} \, j}
\norm{\frac{\hat{\mathbf{g}}^{ti}}{\sqrt{\hat{\mathbf{r}}^{ti}}}-\frac{\mathbf{g}^{tj}}{\sqrt{\mathbf{r}^{tj}}}}\\
\leq& \sum\limits_{t=1}^{T'}\sum\limits_{\text{client} \, i \in \mathcal{F}_t}\sum\limits_{\text{benign client} \, j}
\norm{\frac{\hat{\mathbf{g}}^{ti}}{\sqrt{\hat{\mathbf{r}}^{ti}}}-\frac{\mathbf{g}^{tj}}{\sqrt{\mathbf{r}^{tj}}}}
+\sum\limits_{t=T'+1}^T\sum\limits_{\text{client} \, i \in \mathcal{F}_t}\sum\limits_{\text{benign client} \, j}g_{max} \norm{\frac{1}{\sqrt{\hat{\mathbf{r}}^{ti}}}-\frac{1}{\sqrt{\mathbf{r}^{tj}}}}\\
&+\sum\limits_{t=T'+1}^T\sum\limits_{\text{client} \, i \in \mathcal{F}_t}\sum\limits_{\text{benign client} \, j}\frac{1}{\sqrt{\beta_4  r_{min}}} \norm{\hat{\mathbf{g}}^{ti}-\mathbf{g}^{tj}} \\
&+\sum\limits_{t=T'+1}^T\sum\limits_{\text{client} \, i \in \mathcal{F}_t}\sum\limits_{\text{benign client} \, j}\norm{\hat{\mathbf{g}}^{ti}-\mathbf{g}^{tj}} \norm{\frac{1}{\sqrt{\hat{\mathbf{r}}^{ti}}}-\frac{1}{\sqrt{\mathbf{r}^{tj}}}}\\
\leq& \sum\limits_{t=1}^{T'}\sum\limits_{\text{client} \, i \in \mathcal{F}_t}\sum\limits_{\text{benign client} \, j}
\norm{\frac{\hat{\mathbf{g}}^{ti}}{\sqrt{\hat{\mathbf{r}}^{ti}}}-\frac{\mathbf{g}^{tj}}{\sqrt{\mathbf{r}^{tj}}}}
+\sum\limits_{t=T'+1}^T\sum\limits_{\text{client} \, i \in \mathcal{F}_t}\sum\limits_{\text{benign client} \, j}g_{max} \norm{\frac{1}{\sqrt{\hat{\mathbf{r}}^{ti}}}-\frac{1}{\sqrt{\mathbf{r}^{tj}}}}\\
&+\sum\limits_{t=T'+1}^T\sum\limits_{\text{client} \, i \in \mathcal{F}_t}\sum\limits_{\text{benign client} \, j}\frac{1}{\sqrt{\beta_4  r_{min}}} \norm{\hat{\mathbf{g}}^{ti}-\mathbf{g}^{tj}}+C_gC''_r\\
\leq& \sum\limits_{t=1}^{T'}\sum\limits_{\text{client} \, i \in \mathcal{F}_t}\sum\limits_{\text{benign client} \, j}
\norm{\frac{\hat{\mathbf{g}}^{ti}}{\sqrt{\hat{\mathbf{r}}^{ti}}}-\frac{\mathbf{g}^{tj}}{\sqrt{\mathbf{r}^{tj}}}}
+\frac{C_g}{\sqrt{\beta_4  r_{min}}}+g_{max}C''_r+C_gC''_r\\
\end{split}
\end{equation}
The first inequality is due to Eq. (\ref{eq:rms_u9}). The second inequality follows from Eq. (\ref{eq:rms_u10}). The third inequality is due to Eq. (\ref{eq:rms_vb6}) and the assumption of Theorem 4.
Let $C_\theta=\sum\limits_{t=1}^{T'}\sum\limits_{\text{client} \, i \in \mathcal{F}_t}\sum\limits_{\text{benign client} \, j}
\norm{\frac{\hat{\mathbf{g}}^{ti}}{\sqrt{\hat{\mathbf{r}}^{ti}}}-\frac{\mathbf{g}^{tj}}{\sqrt{\mathbf{r}^{tj}}}}
+\frac{C_g}{\sqrt{\beta_4  r_{min}}}+g_{max}C''_r+C_gC''_r$. Then $C_\theta$ is a positive constant  number, and  \begin{equation}\label{eq:rms_u12}
\begin{split}
\sum\limits_{t=1}^T\sum\limits_{\text{client} \, i \in \mathcal{F}_t}\sum\limits_{\text{benign client} \, j}
\norm{\frac{\hat{\mathbf{g}}^{ti}}{\sqrt{\hat{\mathbf{r}}^{ti}}}- \frac{\mathbf{g}^{tj}}{\sqrt{\mathbf{r}^{tj}}}}\leq C_\theta.
\end{split}
\end{equation}
When $T\leq T'$,
\begin{equation}\label{eq:rms_u13}
\begin{split}
\sum\limits_{t=1}^T\sum\limits_{\text{client} \, i \in \mathcal{F}_t}\sum\limits_{\text{benign client} \, j}
\norm{\frac{\hat{\mathbf{g}}^{ti}}{\sqrt{\hat{\mathbf{r}}^{ti}}}- \frac{\mathbf{g}^{tj}}{\sqrt{\mathbf{r}^{tj}}}}
&\leq\sum\limits_{t=1}^{T'}\sum\limits_{\text{client} \, i \in \mathcal{F}_t}\sum\limits_{\text{benign client} \, j}
\norm{\frac{\hat{\mathbf{g}}^{ti}}{\sqrt{\hat{\mathbf{r}}^{ti}}}- \frac{\mathbf{g}^{tj}}{\sqrt{\mathbf{r}^{tj}}}}\\
&\leq C_\theta.
\end{split}
\end{equation}
Thus Eq. (\ref{eq:rms_u12}) still holds.

The fourth part is to prove $\hat{\boldsymbol{\theta}}^{ti}$ is close to $\boldsymbol{\theta}^{tj}$. According to the definition of $\hat{\boldsymbol{\theta}}^{ti}$ and $\boldsymbol{\theta}^{tj}$ (for simplicity, we omit the learning rate),
\begin{equation}\label{eq:rms_theta1}
\begin{split}
\sum\limits_{t=1}^T\sum\limits_{\text{client} \, i \in \mathcal{F}_t}\sum\limits_{\text{benign client} \, j}
\norm{\hat{\boldsymbol{\theta}}^{ti}-\boldsymbol{\theta}^{tj}}
&=\sum\limits_{t=1}^T\sum\limits_{\text{client} \, i \in \mathcal{F}_t}\sum\limits_{\text{benign client} \, j}
\norm{\overline{\boldsymbol{\theta}}^{t-1}-\frac{\hat{\mathbf{g}}^{ti}}{\sqrt{\hat{\mathbf{r}}^{ti}}}-\overline{\boldsymbol{\theta}}^{t-1}+\frac{\mathbf{g}^{tj}}{\sqrt{\mathbf{r}^{tj}}}}\\
&=\sum\limits_{t=1}^T\sum\limits_{\text{client} \, i \in \mathcal{F}_t}\sum\limits_{\text{benign client} \, j}
\norm{\frac{\hat{\mathbf{g}}^{ti}}{\sqrt{\hat{\mathbf{r}}^{ti}}}- \frac{\mathbf{g}^{tj}}{\sqrt{\mathbf{r}^{tj}}}} \\
&\leq C_\theta.
\end{split}
\end{equation}
The inequality follows from Eq. (\ref{eq:rms_u12}).
Therefore, condition 2 of RMSProp-Byzantine resilience holds.

Since condition 1 and condition 2 of RMSProp-Byzantine resilience all hold, R-RFRS is RMSProp-Byzantine resilient.
\end{proof}

\section{Derivation of Byzantine gradient in camouflage attack}
\begin{figure*}[ht]
	\centering
		\includegraphics[width=0.4\linewidth]{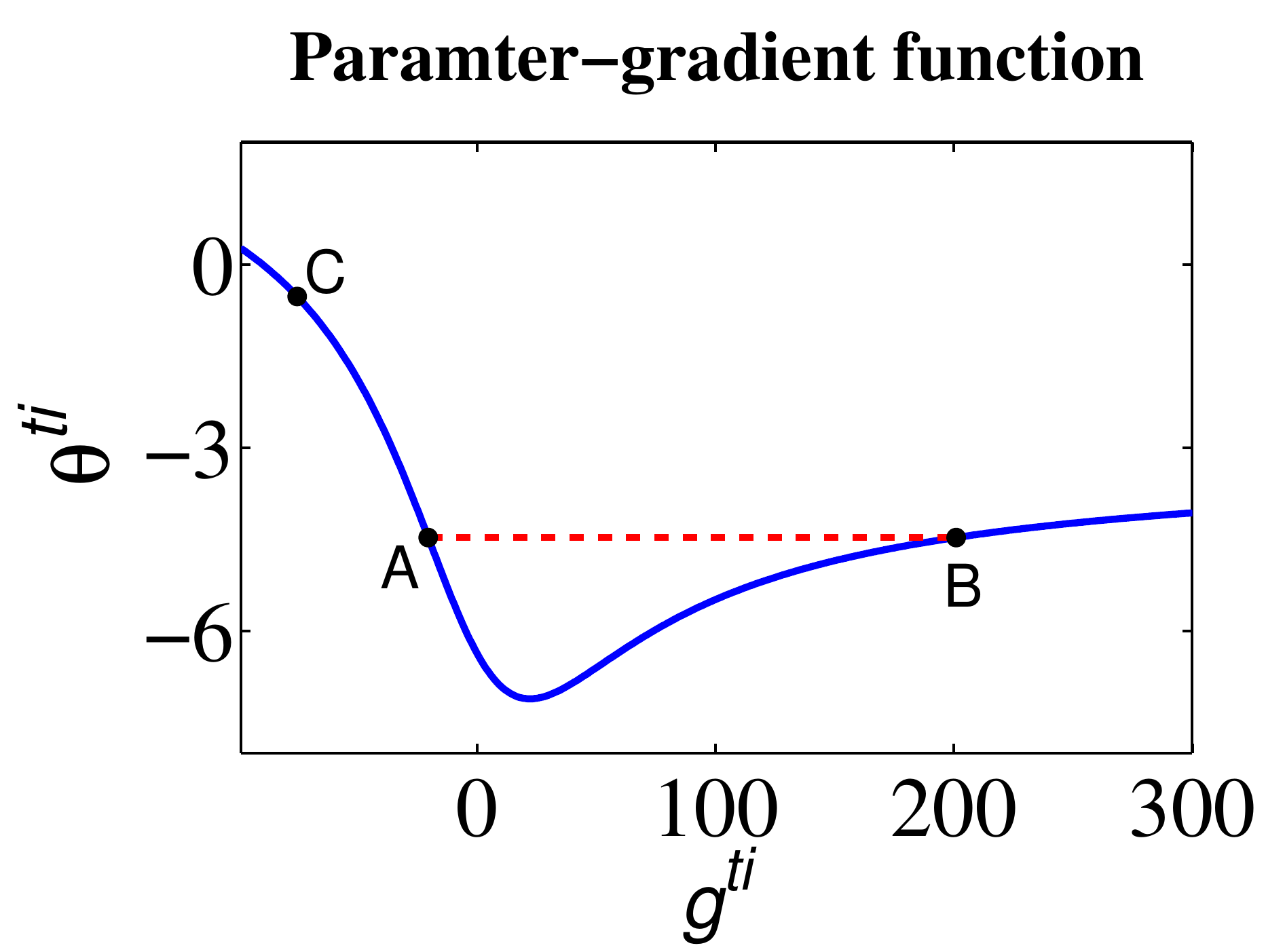}
	\caption{The relation between the model parameter $\theta^{ti}$ and the gradient $g^{ti}$ of client $i$ at round $t$. Suppose point A is the benign point with benign $\theta^{ti}$ and $g^{ti}$, point B is a Byzantine point with the same $\theta^{ti}$ but different $\widetilde{g}^{ti}$. point C has a unique mapping from $\theta^{ti}$ from $g^{ti}$, and thus cannot be camouflaged.}
	\label{fig:app_theta_g}
\end{figure*}
In this section, we demonstrate the derivation of Byzantine gradient in camouflage attack.

As shown in Figure \ref{fig:app_theta_g}, suppose point A is the benign point with benign $\theta^{ti}$ and $g^{ti}$, the Byzantine client can choose Byzantine point B which has the same $\theta^{ti}$ but completely different $\widetilde{g}^{ti}$.
According to Adam update rule, the model parameter of point A and point B are computed by $\theta^{ti}=\overline{\theta}^{t-1}-\eta^t  \frac{\beta_1\overline{m}^{t-1}+\left(1-\beta_1\right)g^{ti}}{\sqrt{\beta_2\overline{v}^{t-1}+\left(1-\beta_2\right)\left(g^{ti}\right)^2}}$ and $\theta^{ti}=\overline{\theta}^{t-1}-\eta^t\frac{\beta_1\overline{m}^{t-1}+\left(1-\beta_1\right)\widetilde{g}^{ti}}{\sqrt{\beta_2\overline{v}^{t-1}+\left(1-\beta_2\right)\left(\widetilde{g}^{ti}\right)^2}}$. Since they have the same update, we can obtain
\begin{equation}\label{eq:cam_1}
\begin{split}
\frac{\beta_1\overline{m}^{t-1}+\left(1-\beta_1\right)g^{ti}}{\sqrt{\beta_2\overline{v}^{t-1}+\left(1-\beta_2\right)\left(g^{ti}\right)^2}}=\frac{\beta_1\overline{m}^{t-1}+\left(1-\beta_1\right)\widetilde{g}^{ti}}{\sqrt{\beta_2\overline{v}^{t-1}+\left(1-\beta_2\right)\left(\widetilde{g}^{ti}\right)^2}}.
\end{split}
\end{equation}
By removing the fractions and square roots,
\begin{equation}\label{eq:cam_2}
\begin{split}
&\left(\beta_1\overline{m}^{t-1}+\left(1-\beta_1\right)g^{ti}\right)^2
\left(\beta_2\overline{v}^{t-1}+\left(1-\beta_2\right)\left(\widetilde{g}^{ti}\right)^2\right)\\
&=\left(\beta_1\overline{m}^{t-1}+\left(1-\beta_1\right)\widetilde{g}^{ti}\right)^2
\left(\beta_2\overline{v}^{t-1}+\left(1-\beta_2\right)\left(g^{ti}\right)^2\right).
\end{split}
\end{equation}
By removing the parentheses,
\begin{equation}\label{eq:cam_3}
\begin{split}
&\left(\beta_1^2\left(\overline{m}^{t-1}\right)^2+2\beta_1\left(1-\beta_1\right)\overline{m}^{t-1}g^{ti}+\left(1-\beta_1\right)^2\left(g^{ti}\right)^2\right)\left(\beta_2\overline{v}^{t-1}+\left(1-\beta_2\right)\left(\widetilde{g}^{ti}\right)^2\right)\\
&=\left(\beta_1^2\left(\overline{m}^{t-1}\right)^2+2\beta_1\left(1-\beta_1\right)\overline{m}^{t-1}\widetilde{g}^{ti}+\left(1-\beta_1\right)^2\left(\widetilde{g}^{ti}\right)^2\right)\left(\beta_2\overline{v}^{t-1}+\left(1-\beta_2\right)\left(g^{ti}\right)^2\right).
\end{split}
\end{equation}

\begin{equation}\label{eq:cam_4}
\begin{split}
&\beta_1^2\left(\overline{m}^{t-1}\right)^2\beta_2\overline{v}^{t-1}+2\beta_1\overline{m}^{t-1}\beta_2\overline{v}^{t-1}\left(1-\beta_1\right)g^{ti}\\
&+\beta_2\overline{v}^{t-1}\left(1-\beta_1\right)^2\left(g^{ti}\right)^2
+\beta_1^2\left(\overline{m}^{t-1}\right)^2\left(1-\beta_2\right)\left(\widetilde{g}^{ti}\right)^2\\
&+2\beta_1\overline{m}^{t-1}\left(1-\beta_1\right)g^{ti}\left(1-\beta_2\right)\left(\widetilde{g}^{ti}\right)^2+\left(1-\beta_1\right)^2\left(g^{ti}\right)^2\left(1-\beta_2\right)\left(\widetilde{g}^{ti}\right)^2\\
=&\beta_1^2\left(\overline{m}^{t-1}\right)^2\beta_2\overline{v}^{t-1}+2\beta_1\overline{m}^{t-1}\beta_2\overline{v}^{t-1}\left(1-\beta_1\right)\widetilde{g}^{ti}\\
&+\beta_2\overline{v}^{t-1}\left(1-\beta_1\right)^2\left(\widetilde{g}^{ti}\right)^2
+\beta_1^2\left(\overline{m}^{t-1}\right)^2\left(1-\beta_2\right)\left(g^{ti}\right)^2\\
&+2\beta_1\overline{m}^{t-1}\left(1-\beta_1\right)\widetilde{g}^{ti}\left(1-\beta_2\right)\left(g^{ti}\right)^2+\left(1-\beta_1\right)^2\left(\widetilde{g}^{ti}\right)^2\left(1-\beta_2\right)\left(g^{ti}\right)^2.
\end{split}
\end{equation}
By deleting the same items,
\begin{equation}\label{eq:cam_5}
\begin{split}
&2\beta_1\overline{m}^{t-1}\beta_2\overline{v}^{t-1}\left(1-\beta_1\right)\left(g^{ti}-\widetilde{g}^{ti}\right)+\beta_2\overline{v}^{t-1}\left(1-\beta_1\right)^2\left(\left(g^{ti}\right)^2-\left(\widetilde{g}^{ti}\right)^2\right)\\
&-\beta_1^2\left(\overline{m}^{t-1}\right)^2\left(1-\beta_2\right)\left(\left(g^{ti}\right)^2-\left(\widetilde{g}^{ti}\right)^2\right)-2\beta_1\overline{m}^{t-1}\left(1-\beta_1\right)\left(1-\beta_2\right)g^{ti}\widetilde{g}^{ti}\left(g^{ti}-\widetilde{g}^{ti}\right)=0.
\end{split}
\end{equation}
Since $g^{ti}$ does not equal to $\widetilde{g}^{ti}$, we can divide the equation by $(g^{ti}-\widetilde{g}^{ti})$,
\begin{equation}\label{eq:cam_6}
\begin{split}
&2\beta_1\overline{m}^{t-1}\beta_2\overline{v}^{t-1}\left(1-\beta_1\right)+\beta_2\overline{v}^{t-1}\left(1-\beta_1\right)^2\left(g^{ti}+\widetilde{g}^{ti}\right)\\
&-\beta_1^2\left(\overline{m}^{t-1}\right)^2\left(1-\beta_2\right)\left(g^{ti}+\widetilde{g}^{ti}\right)-2\beta_1\overline{m}^{t-1}\left(1-\beta_1\right)\left(1-\beta_2\right)g^{ti}\widetilde{g}^{ti}=0.
\end{split}
\end{equation}
Thus, we can get
\begin{equation}\label{eq:cam_7}
\begin{split}
&-\left(\beta_2\overline{v}^{t-1}\left(1-\beta_1\right)^2-\beta_1^2\left(\overline{m}^{t-1}\right)^2\left(1-\beta_2\right)-2\beta_1\overline{m}^{t-1}\left(1-\beta_1\right)\left(1-\beta_2\right)g^{ti}\right)\widetilde{g}^{ti}\\
&=2\beta_1\overline{m}^{t-1}\beta_2\overline{v}^{t-1}\left(1-\beta_1\right)+\beta_2\overline{v}^{t-1}\left(1-\beta_1\right)^2g^{ti}-\beta_1^2\left(\overline{m}^{t-1}\right)^2\left(1-\beta_2\right)g^{ti}.
\end{split}
\end{equation}
Then, we can obtain $\widetilde{g}^{ti}$:
\begin{equation}\label{eq:cam_8}
\begin{split}
\widetilde{g}^{ti}=\frac{2\beta_1\beta_2\left(1-\beta_1\right)\overline{m}^{t-1}\overline{v}^{t-1}+\beta_2\left(1-\beta_1\right)^2\overline{v}^{t-1}g^{ti}-\beta_1^2\left(1-\beta_2\right)\left(\overline{m}^{t-1}\right)^2g^{ti}}{\beta_1^2\left(1-\beta_2\right)\left(\overline{m}^{t-1}\right)^2+2\beta_1\left(1-\beta_1\right)\left(1-\beta_2\right)\overline{m}^{t-1}g^{ti}-\beta_2\left(1-\beta_1\right)^2\overline{v}^{t-1}}.
\end{split}
\end{equation}

It is worth noting that not all $\theta^{ti}$ can be camouflaged. For example, in Figure \ref{fig:app_theta_g}, point C has a unique mapping from $\theta^{ti}$ to $g^{ti}$. In this case, if we compute $\widetilde{g}^{ti}$ by Eq. (\ref{eq:cam_8}), the Byzantine update of model parameter will equal to negative benign update of model parameter (i.e., $\frac{\beta_1\overline{m}^{t-1}+\left(1-\beta_1\right)g^{ti}}{\sqrt{\beta_2\overline{v}^{t-1}+\left(1-\beta_2\right)\left(g^{ti}\right)^2}}=-\frac{\beta_1\overline{m}^{t-1}+\left(1-\beta_1\right)\widetilde{g}^{ti}}{\sqrt{\beta_2\overline{v}^{t-1}+\left(1-\beta_2\right)\left(\widetilde{g}^{ti}\right)^2}}$). Thus, Eq. (\ref{eq:cam_2}-\ref{eq:cam_8}) hold but Eq. (\ref{eq:cam_1}) does not hold.

\section{Datasets, baselines and configuration of experiments}
In this section, we show the details of datasets, baselines and configuration of experiments.

We conduct our experiments on 4 real-world datasets: Last.fm \cite{Cantador_RecSys2011}, ML100K \cite{harper2015movielens}, Citeulike-a \cite{conf/kdd/WangB11}, and Citeulike-t \cite{conf/ijcai/WangCL13}.
Last.fm contains music artist listening information from Last.fm online music system with 1,892 users, 17,632 artists, and 92,834 listening records.
ML100K, which was collected through the MovieLens website, contains 100,000 movie ratings from 943 users on 1,682 movies.
Citeulike-a and Citeulike-t are collected in a real-world community of researchers and their citation. Citeulike-a contains 5,551 users and 16,980 articles with 204,986 user-item pairs. Citeulike-t contains 7,947 users and 25,975 articles with 134,860 user-item pairs.

We compare A-RFRS with three baselines: Krum \cite{conf/nips/BlanchardMGS17}, RFA \cite{journals/corr/abs-1912-13445}, and Trmean \cite{journals/corr/abs-1903-06996}. Krum precludes the model parameters that are too far away and aggregates the remaining model parameters. RFA replaces the weighted arithmetic mean aggregation with an approximate geometric median. Trmean removes model parameters with large norms or small norms.

We set the dimension of $\mathbf{p}^j,\mathbf{q}^j$ (embedding vectors of item $j$) $d=64$, learning rate $\eta=10^{-3}$, client ratio $e=10^{-2}$, hyperparameter $\gamma=1$, hyperparameter of Adam $\beta_1=0.9$, $\beta_2=0.999$, and the regularization coefficient $\lambda=10^{-4}$.
All hyperparameter of baseline methods that are not mentioned above are set to their default value. All experiments are run on the same machine with i7-5820K CPU, 64GB RAM, and three GeForce GTX TITAN X GPU.

\section{Additional experiments}
In this section, we show the results of three additional experiments. Section \ref{sec:recall} shows Recall@$K$ of A-FRS and A-RFRS compared to baseline methods. Section \ref{sec:impact} demonstrates the impact of Adam optimizer in learning an FRS. Section \ref{sec:other_optimizer} shows that our learning strategy is also effective on FRS using other optimizers (SGD with momentum~\cite{conf/icml/SutskeverMDH13} and AdaGrad~\cite{journals/jmlr/DuchiHS11}).

\subsection{Recall@K of A-FRS and A-RFRS}
\label{sec:recall}
\begin{figure*}[!h]
	\centering
	\subfigure{
		\includegraphics[width=0.4\linewidth]{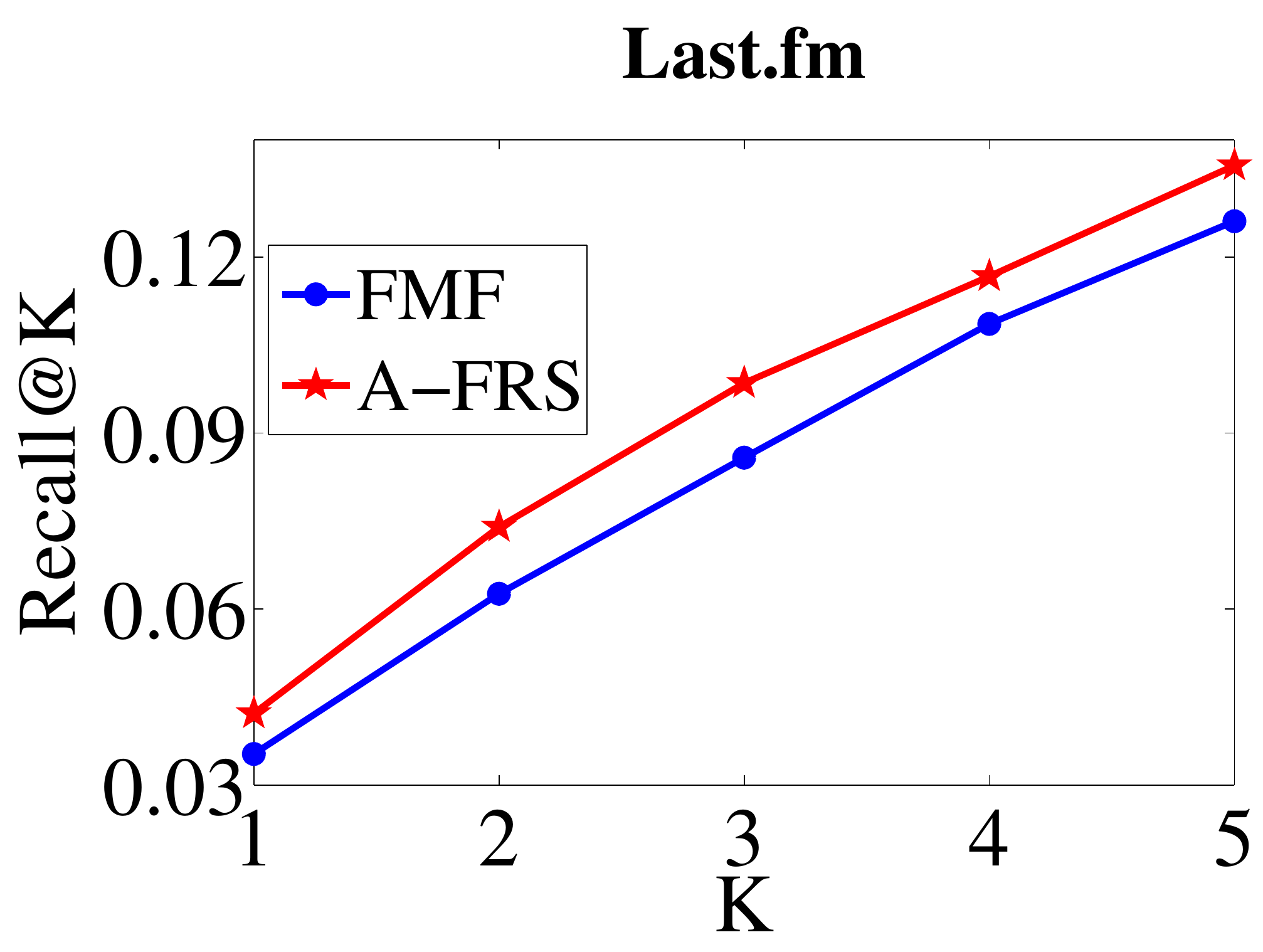}
	}
	\subfigure{
		\includegraphics[width=0.4\linewidth]{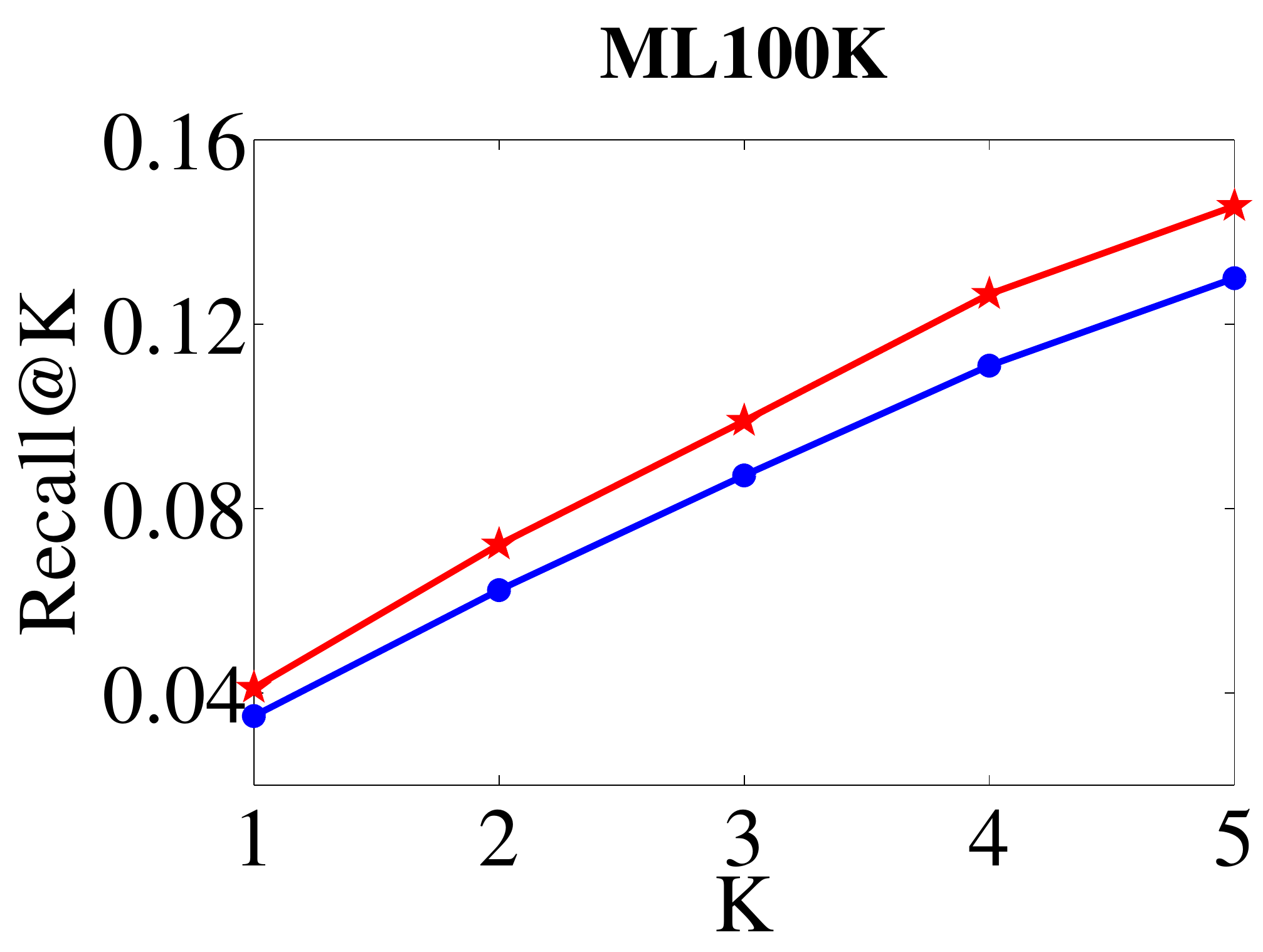}
	}
	\subfigure{
		\includegraphics[width=0.4\linewidth]{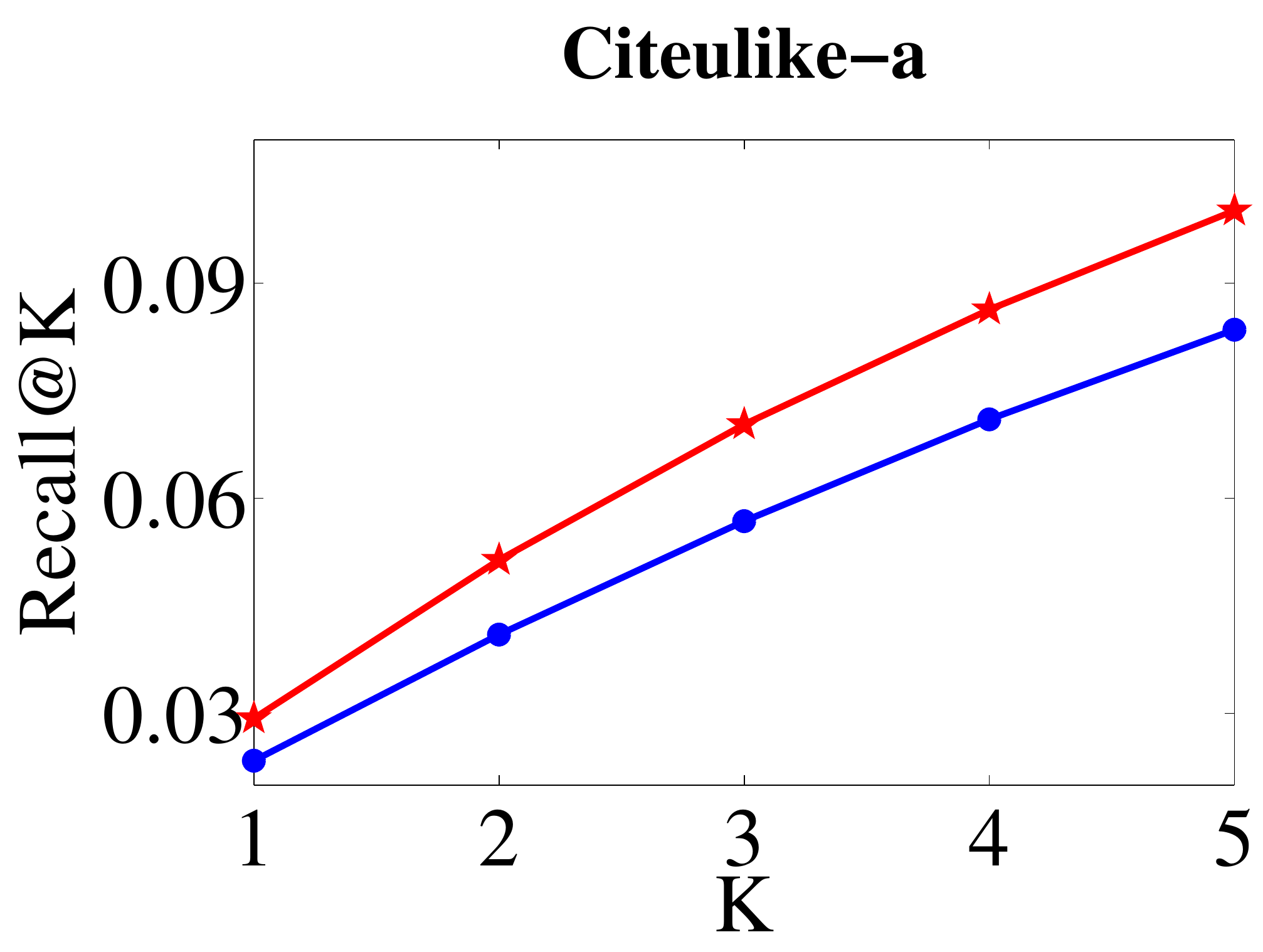}
	}
	\subfigure{
		\includegraphics[width=0.4\linewidth]{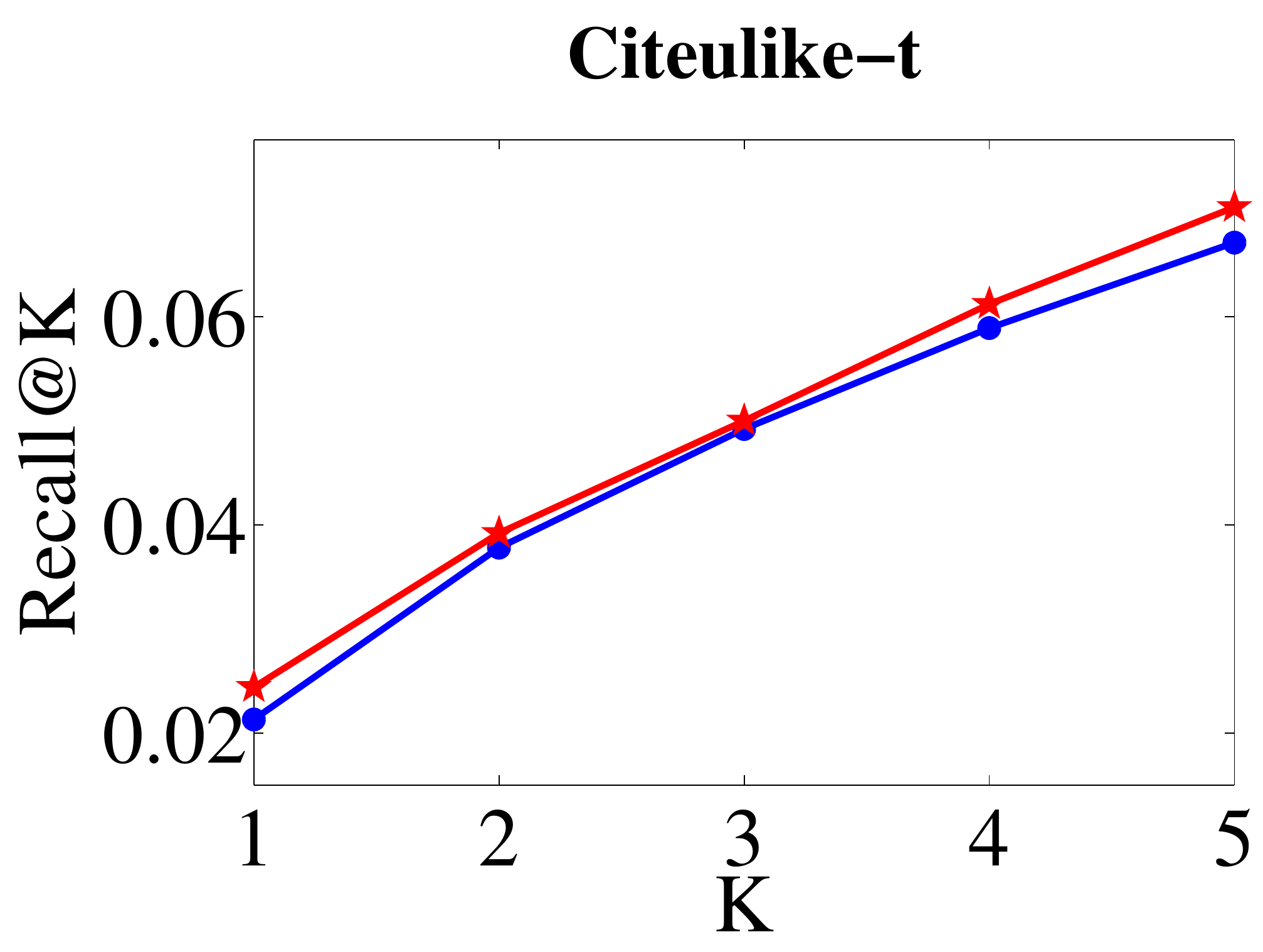}
	}
	\caption{Recall@$K$ of FRSs on 4 datasets. A-FRS (red line) is our proposed federated recommendation method.}
	\label{fig:recall_performance1}
\end{figure*}
\begin{figure*}[!ht]
	\centering
	\subfigure{
		\includegraphics[width=0.31\linewidth]{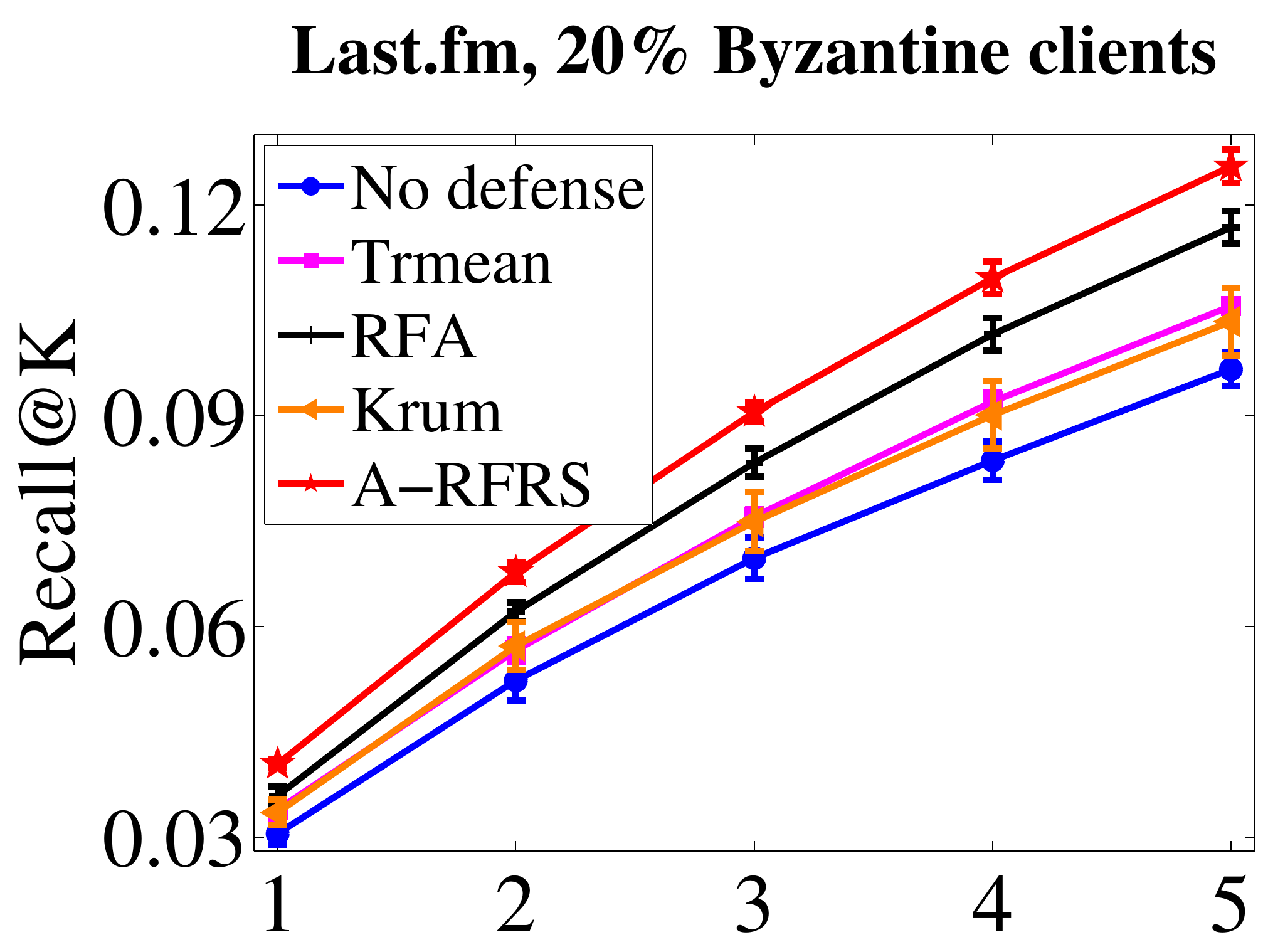}
	}
	\subfigure{
		\includegraphics[width=0.31\linewidth]{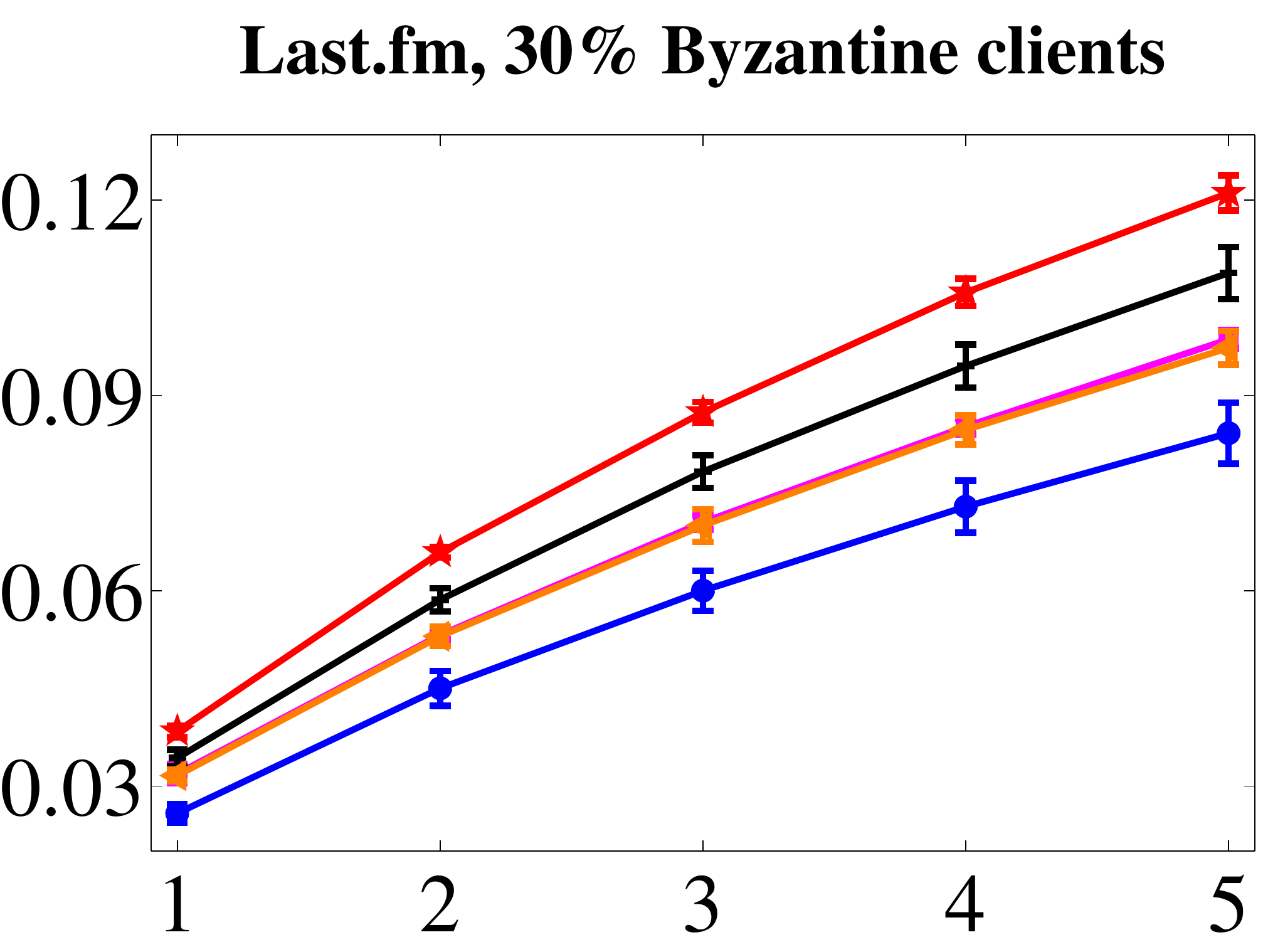}
	}
	\subfigure{
		\includegraphics[width=0.31\linewidth]{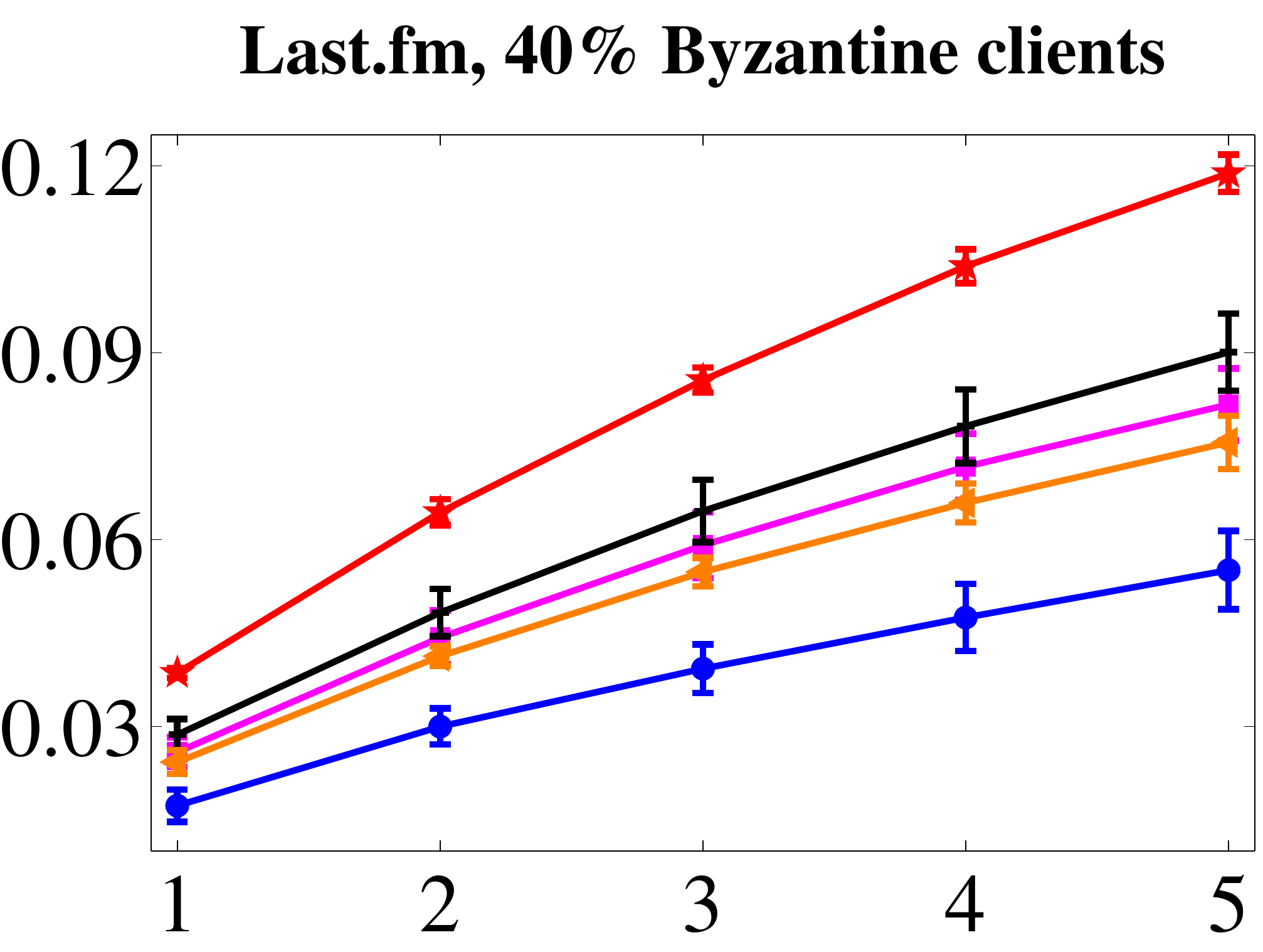}
	}
	\subfigure{
		\includegraphics[width=0.31\linewidth]{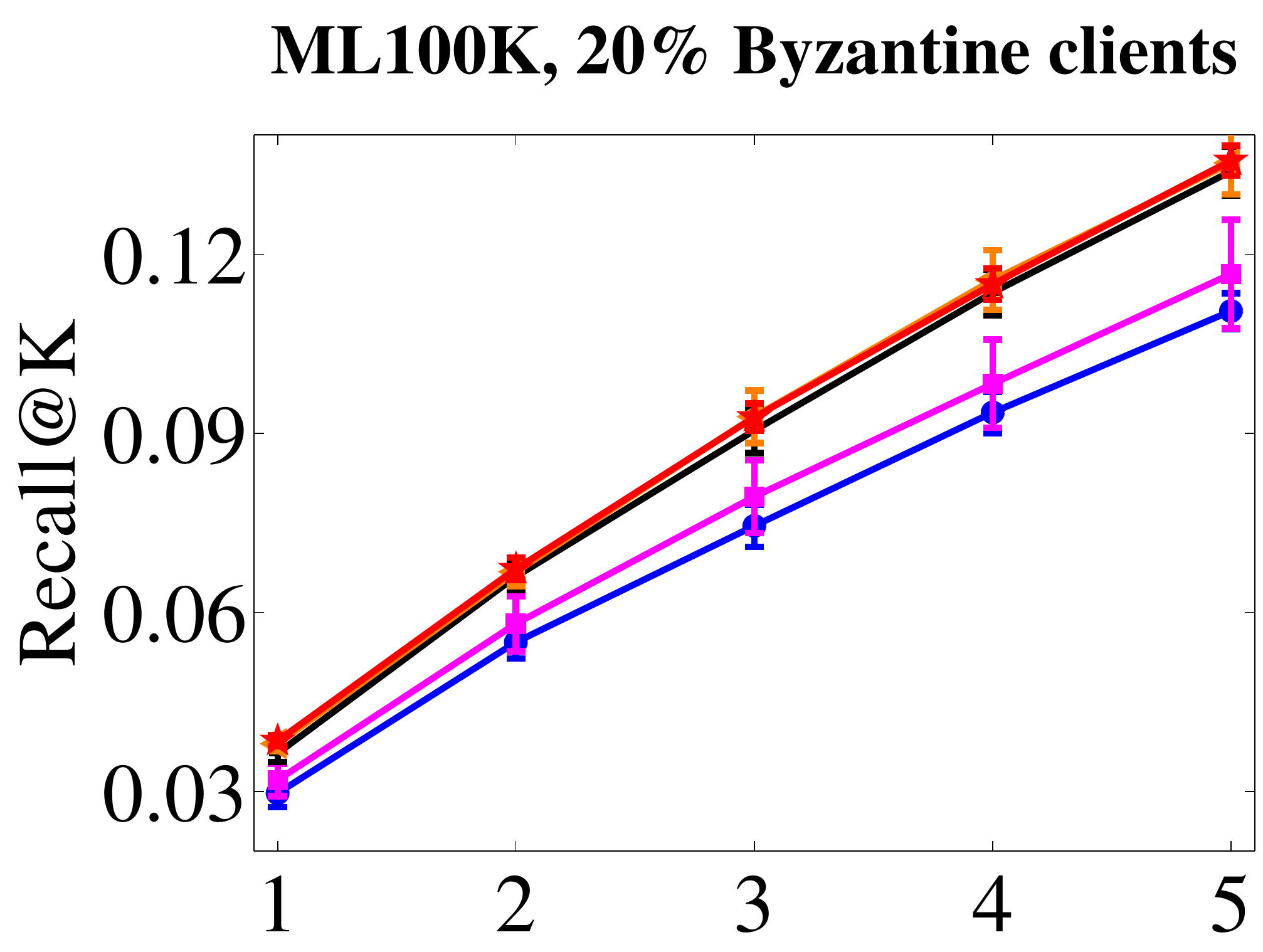}
	}
	\subfigure{
		\includegraphics[width=0.31\linewidth]{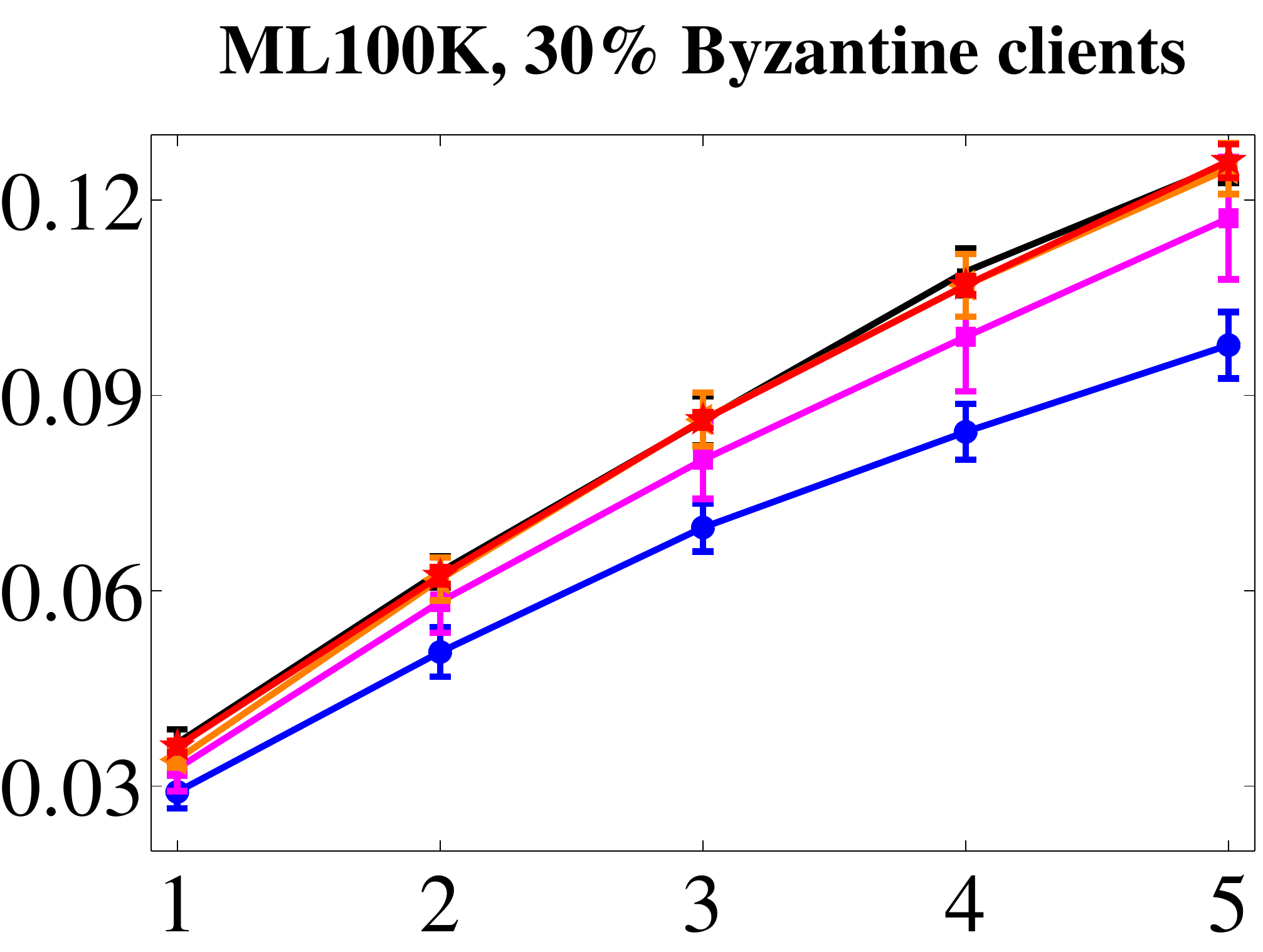}
	}
	\subfigure{
		\includegraphics[width=0.31\linewidth]{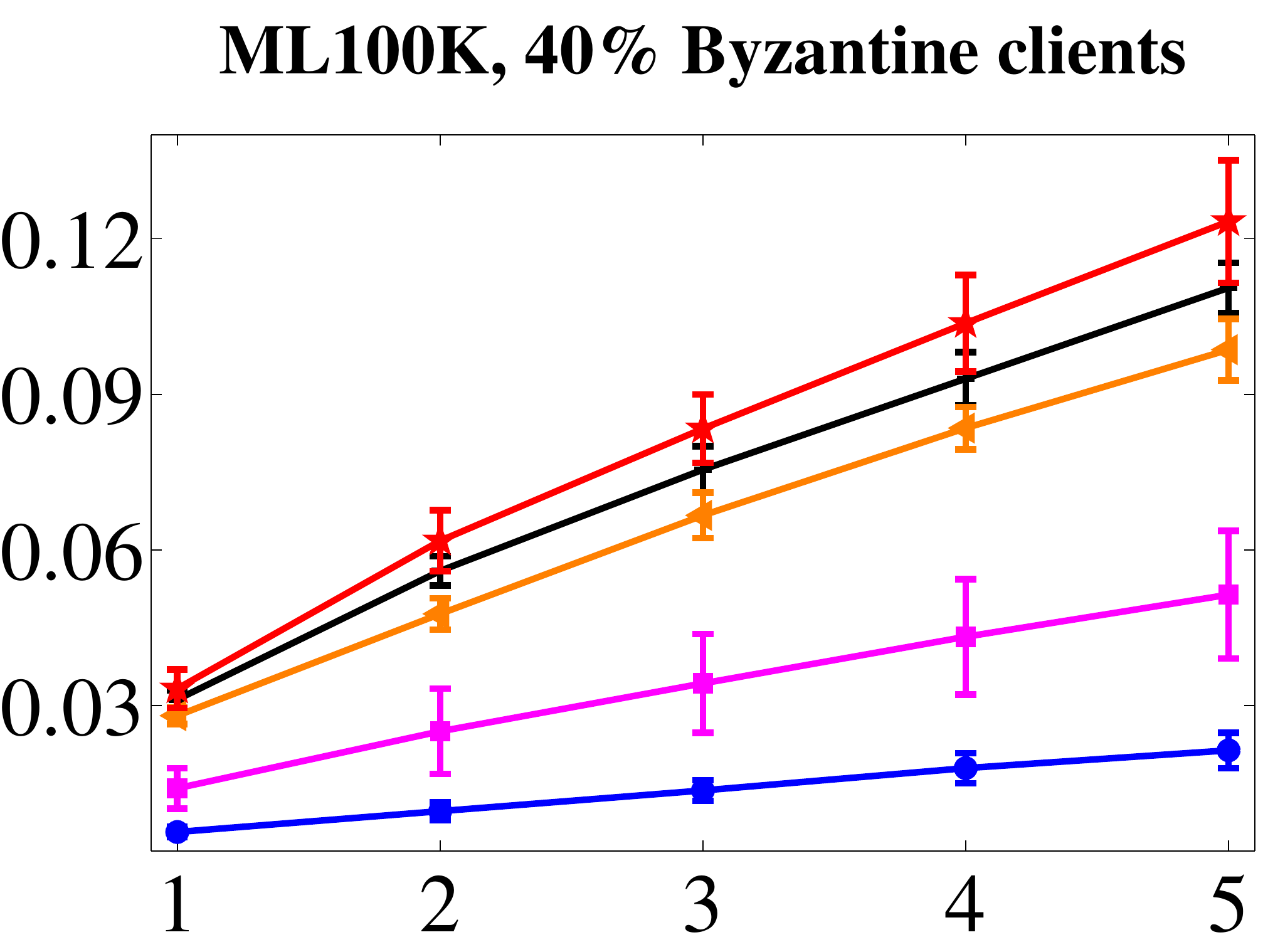}
	}
	\subfigure{
		\includegraphics[width=0.31\linewidth]{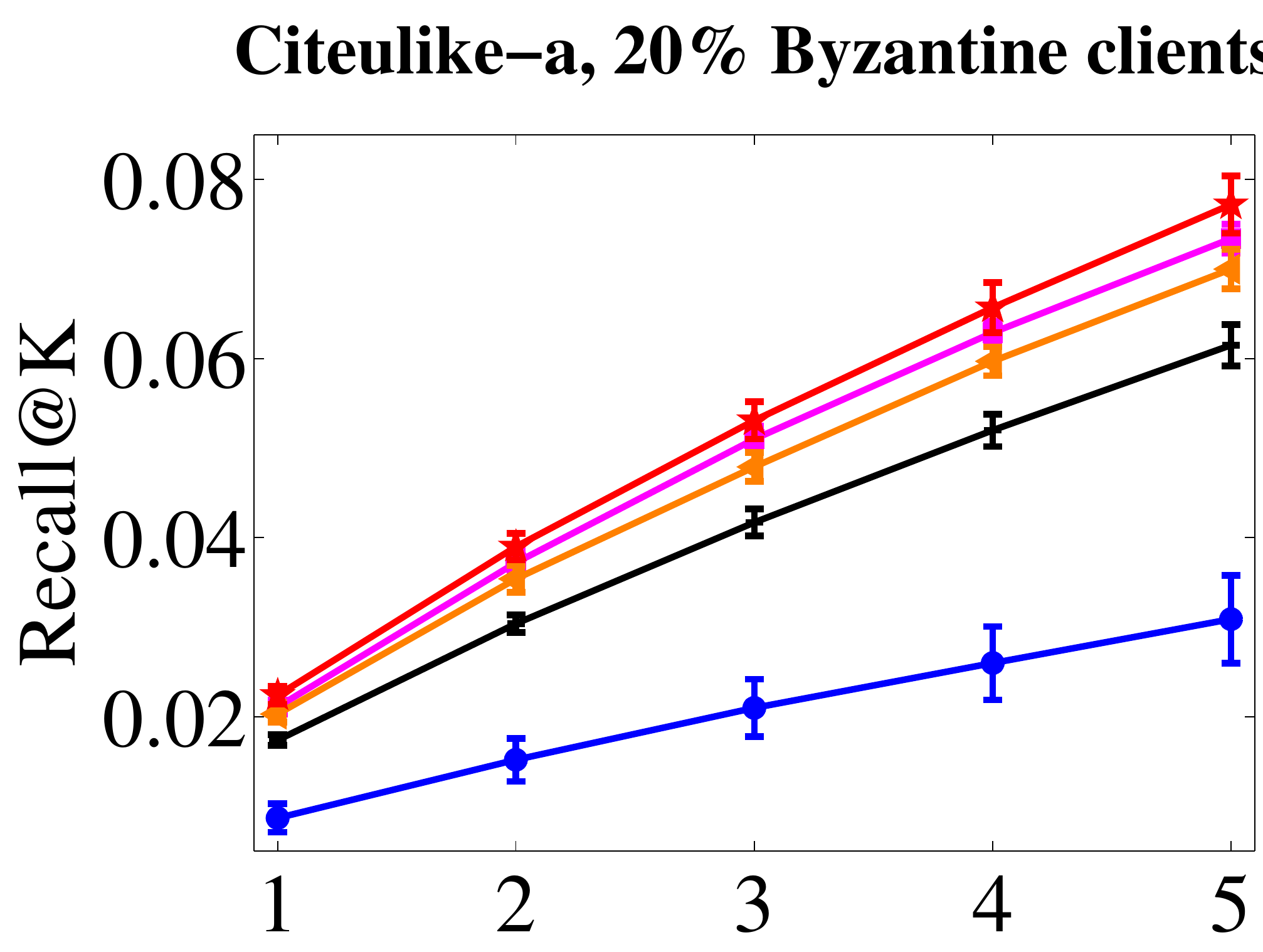}
	}
	\subfigure{
		\includegraphics[width=0.31\linewidth]{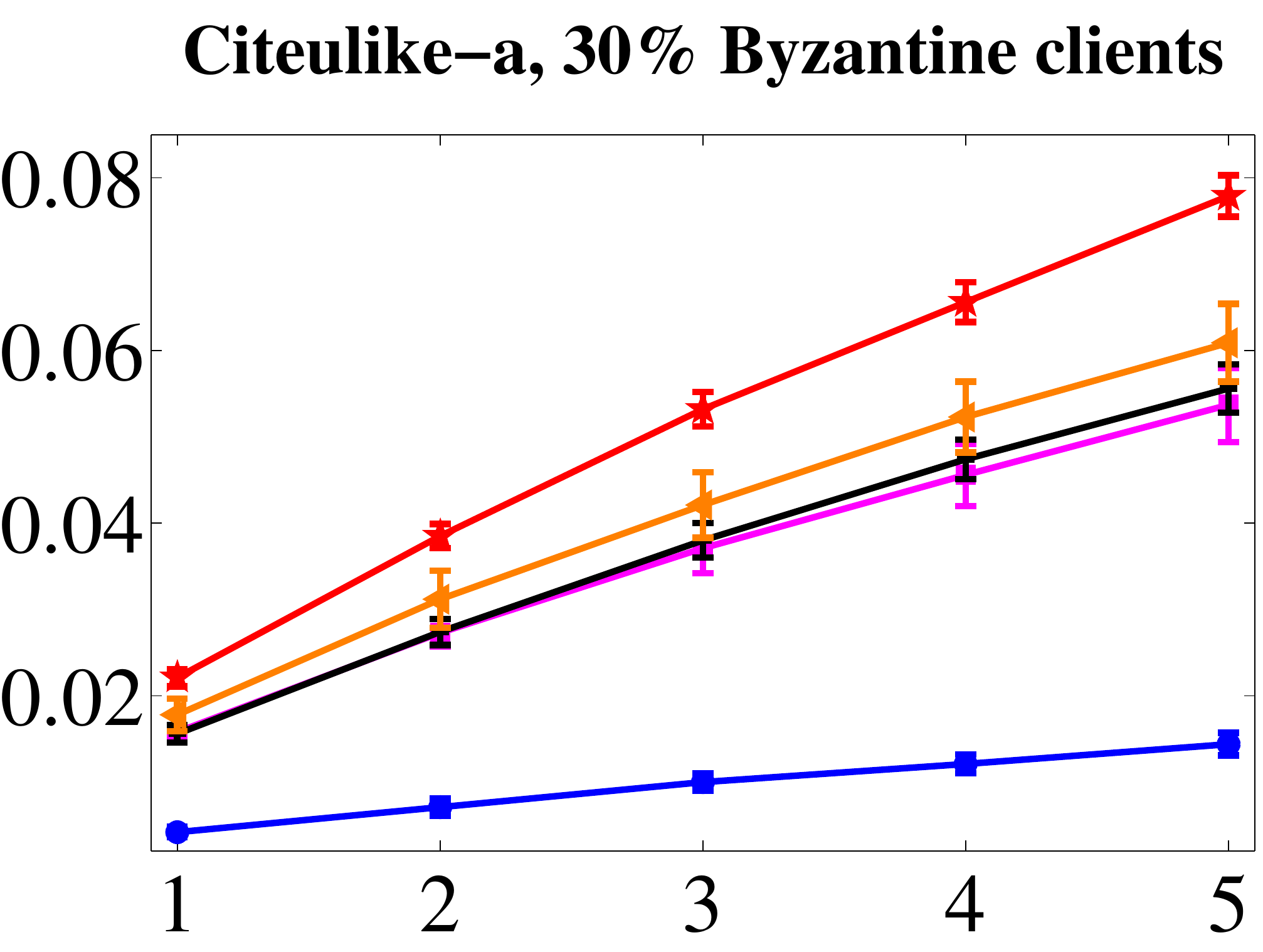}
	}
	\subfigure{
		\includegraphics[width=0.31\linewidth]{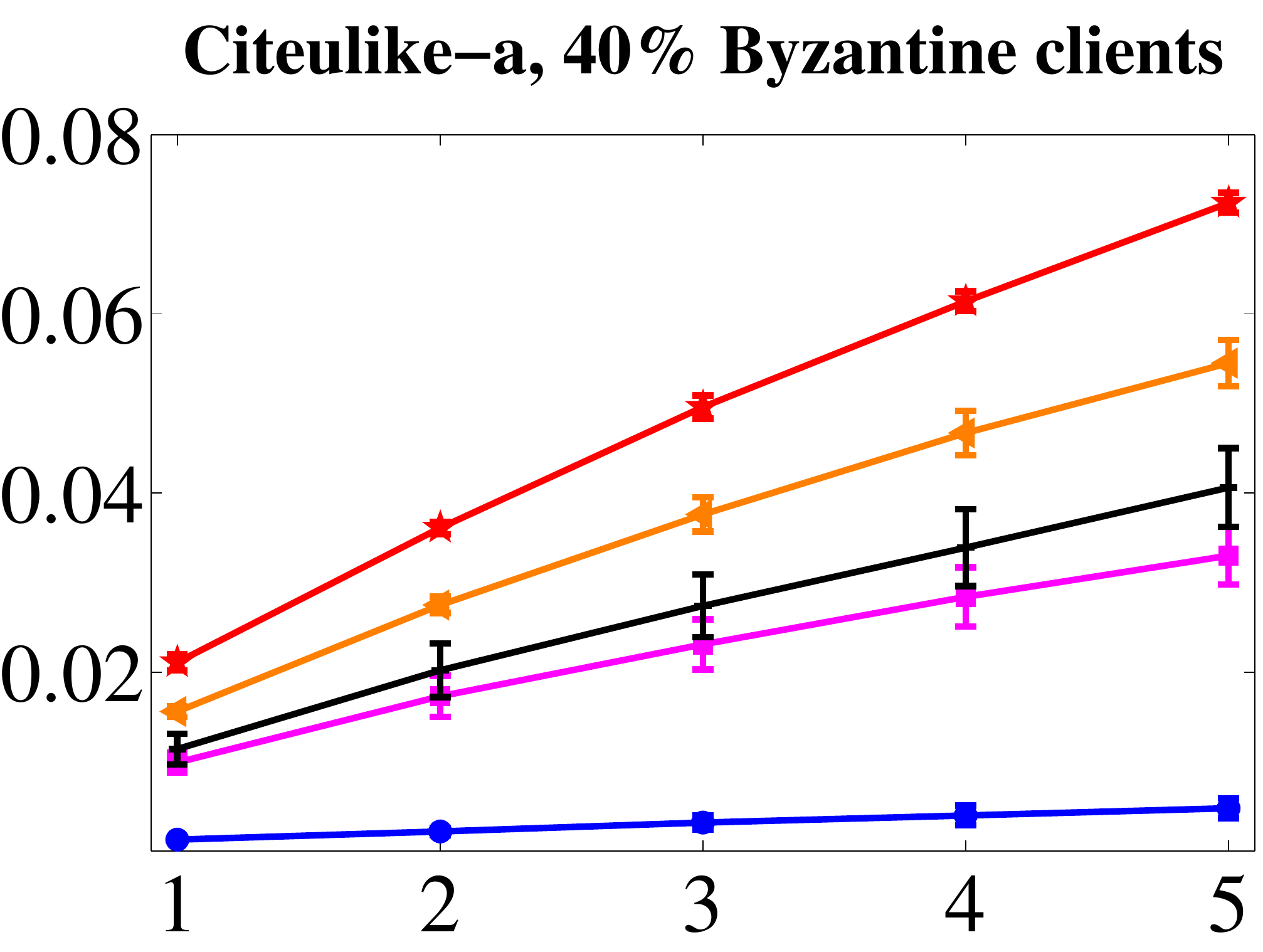}
	}
	\subfigure{
		\includegraphics[width=0.31\linewidth]{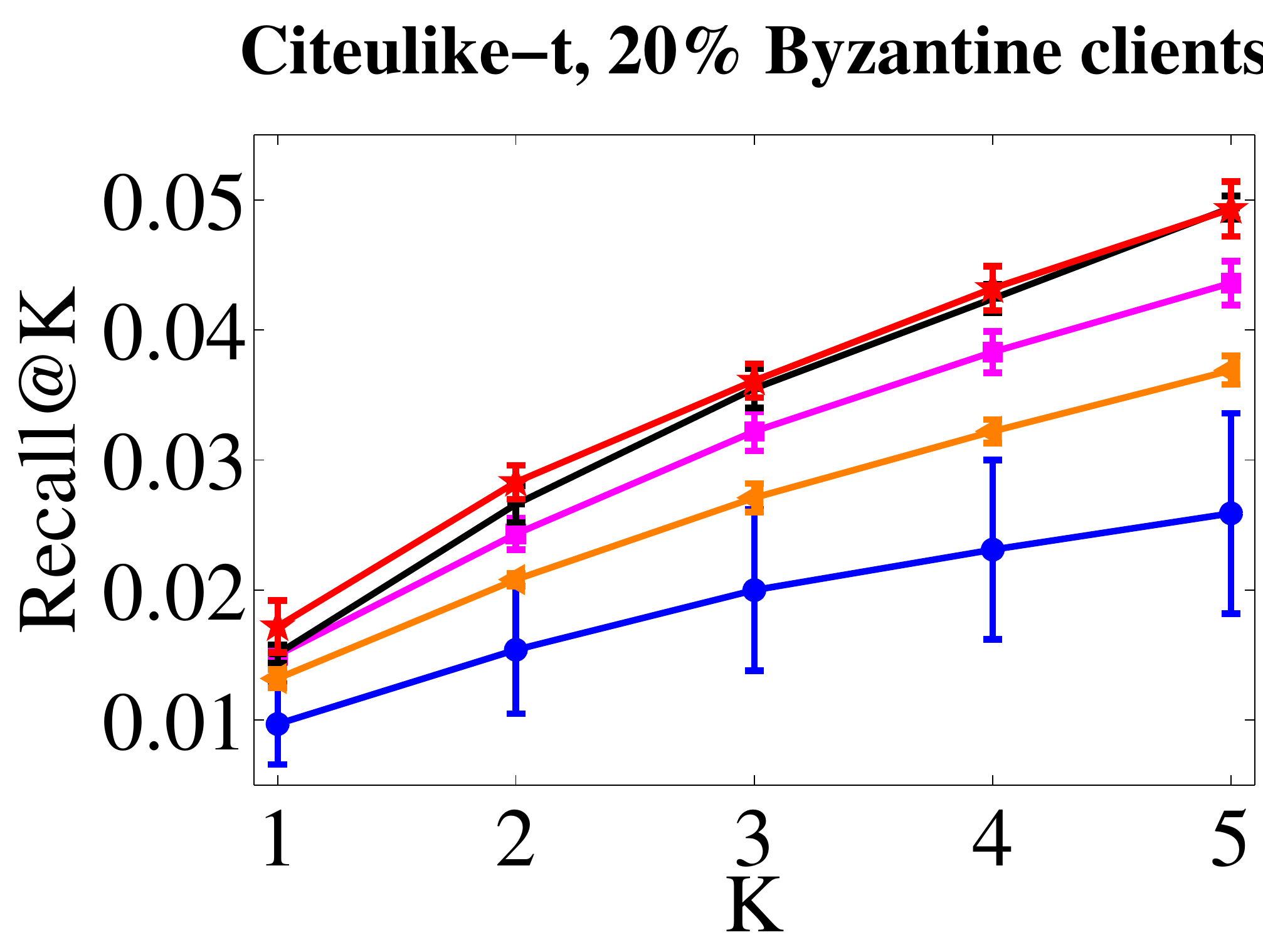}
	}
	\subfigure{
		\includegraphics[width=0.31\linewidth]{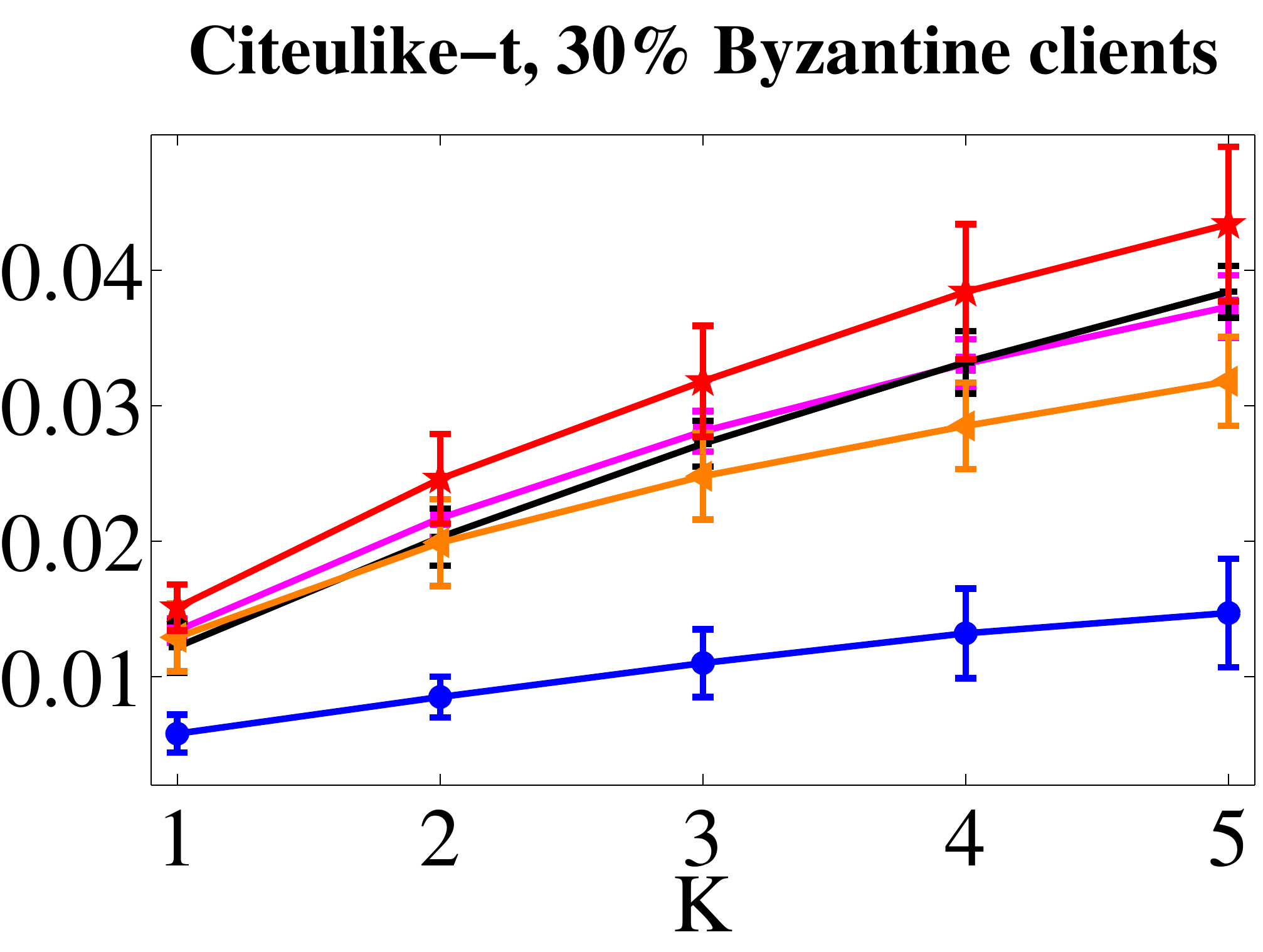}
	}
	\subfigure{
		\includegraphics[width=0.31\linewidth]{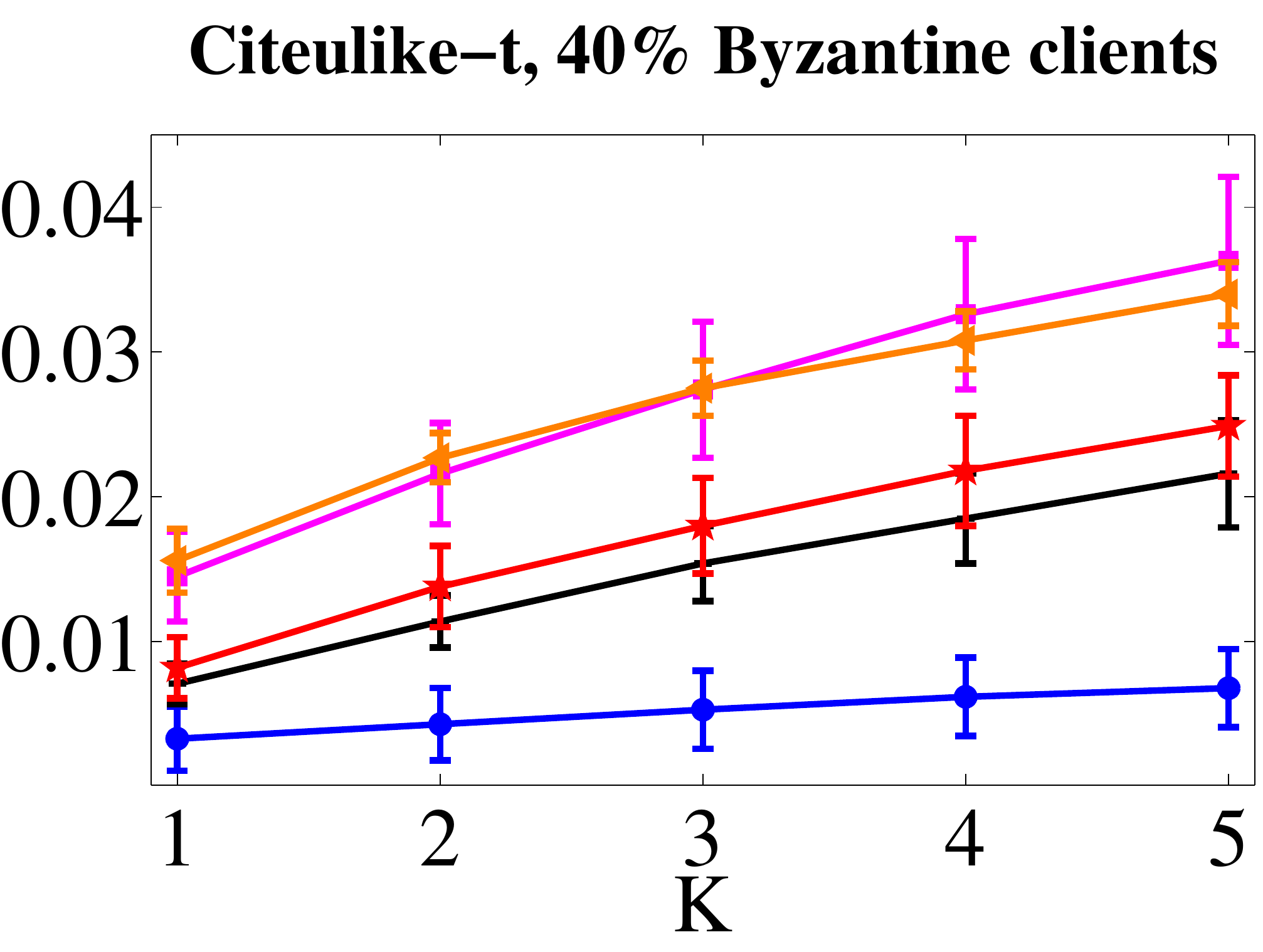}
	}
	\caption{Recall@$K$ (mean and standard deviation) of all methods on 4 datasets and 3 different numbers of Byzantine clients. A-RFRS (red line) is our proposed defense method.}
	\label{fig:recall_performance2}
\end{figure*}
\clearpage
\subsection{Impact of Adam optimizer}
\label{sec:impact}
\begin{figure*}[ht]
	\centering
	\subfigure{
		\includegraphics[width=0.4\linewidth]{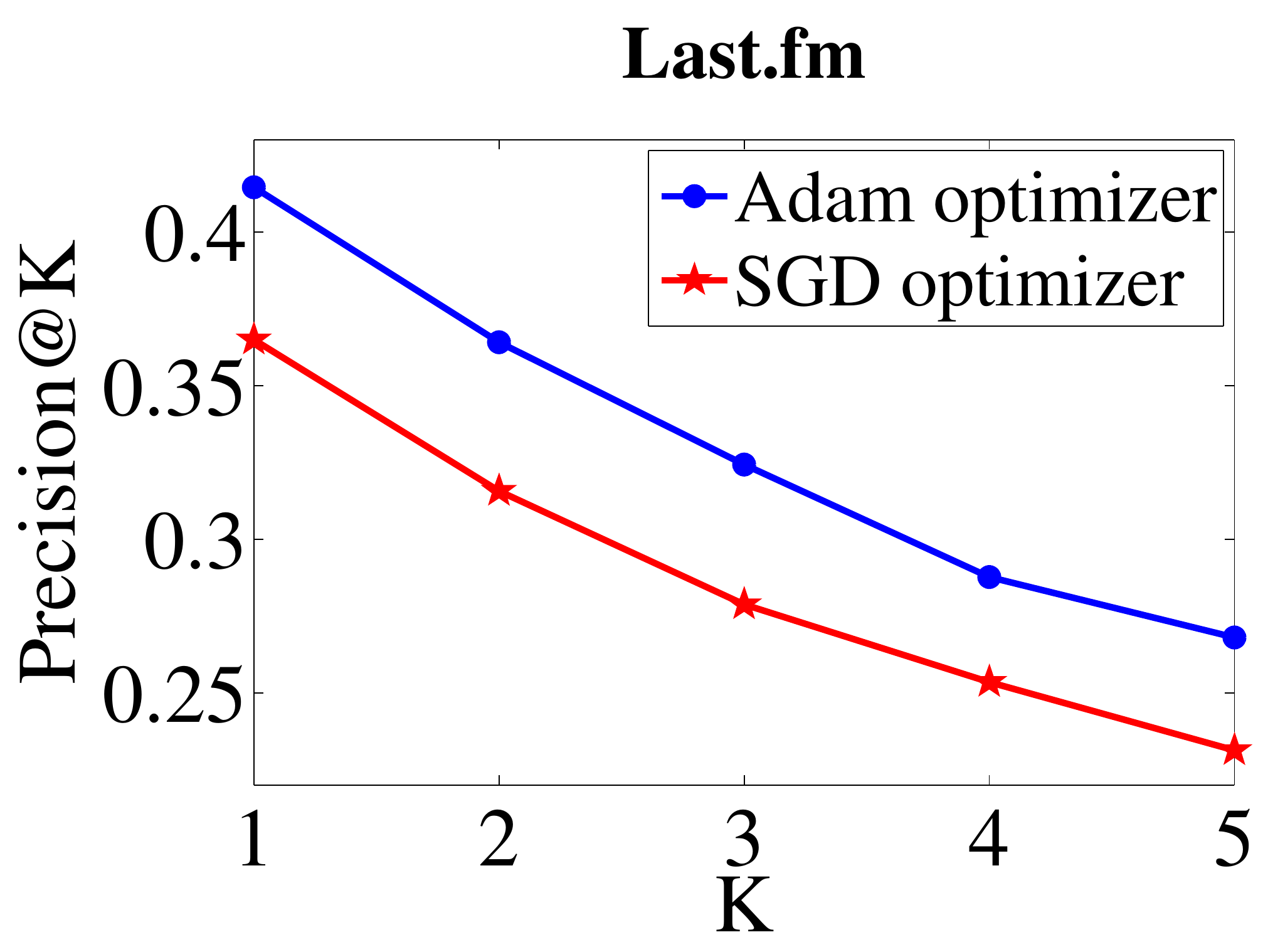}
	}
	\subfigure{
		\includegraphics[width=0.4\linewidth]{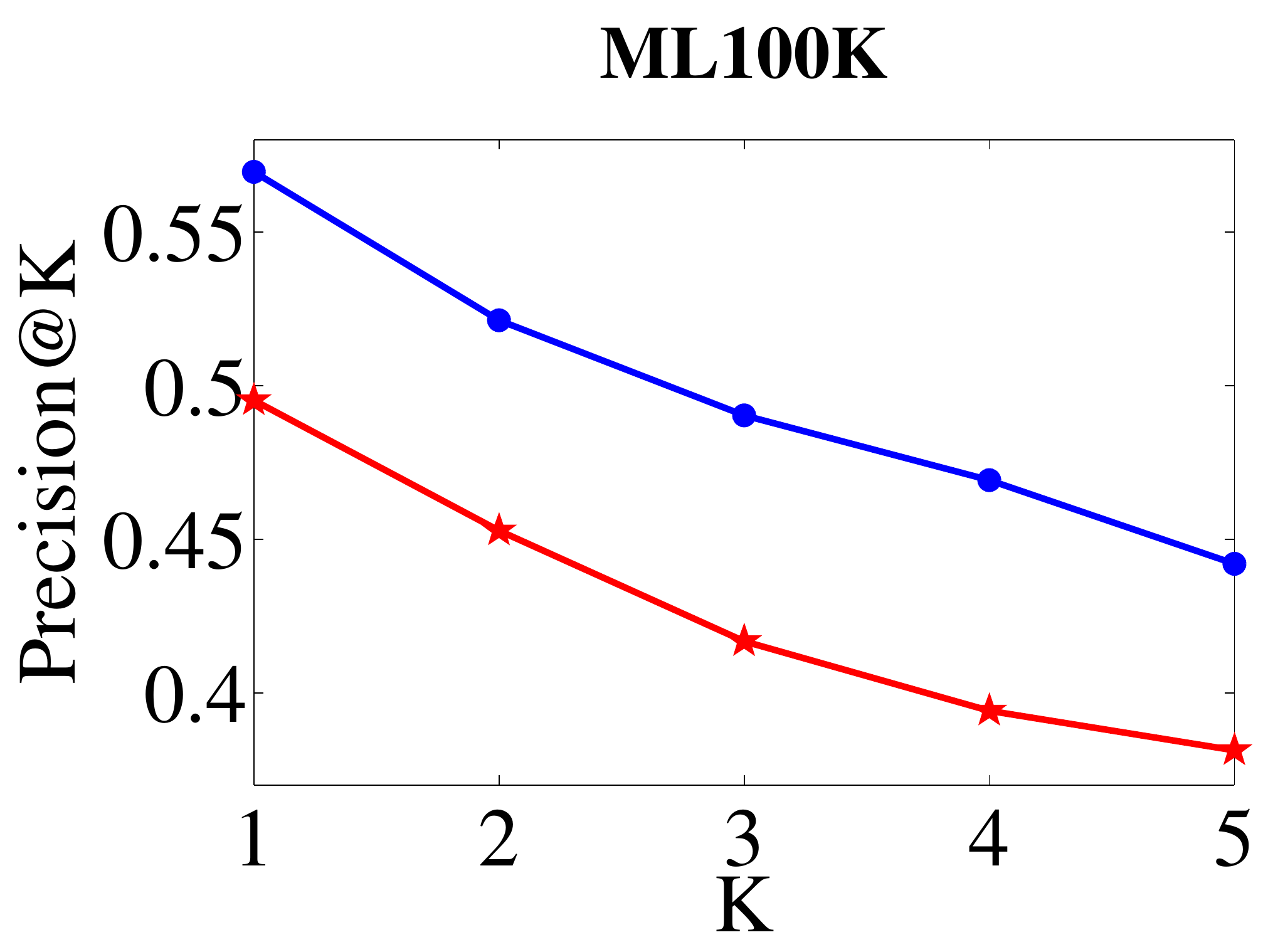}
	}
	\subfigure{
		\includegraphics[width=0.4\linewidth]{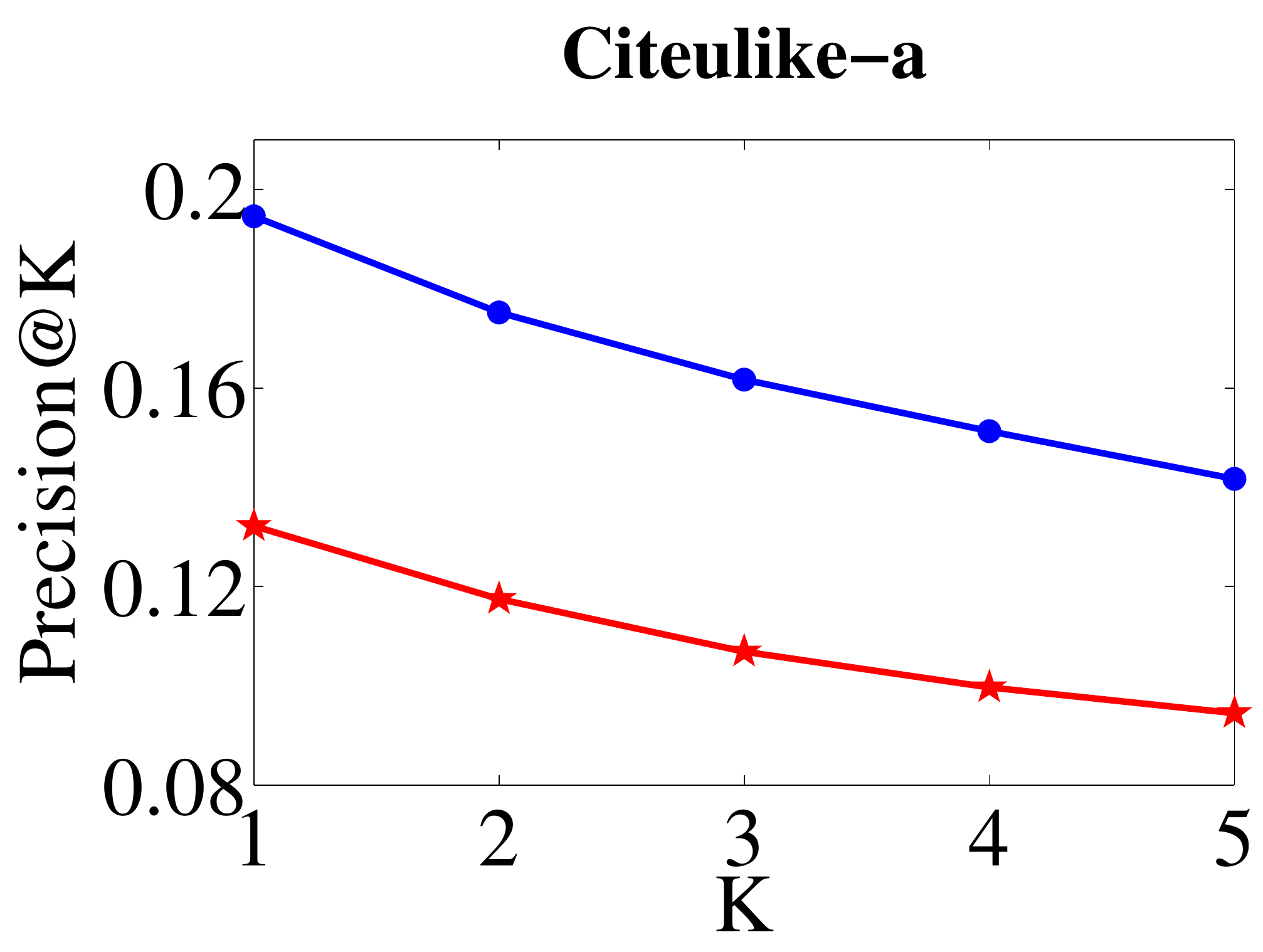}
	}
	\subfigure{
		\includegraphics[width=0.4\linewidth]{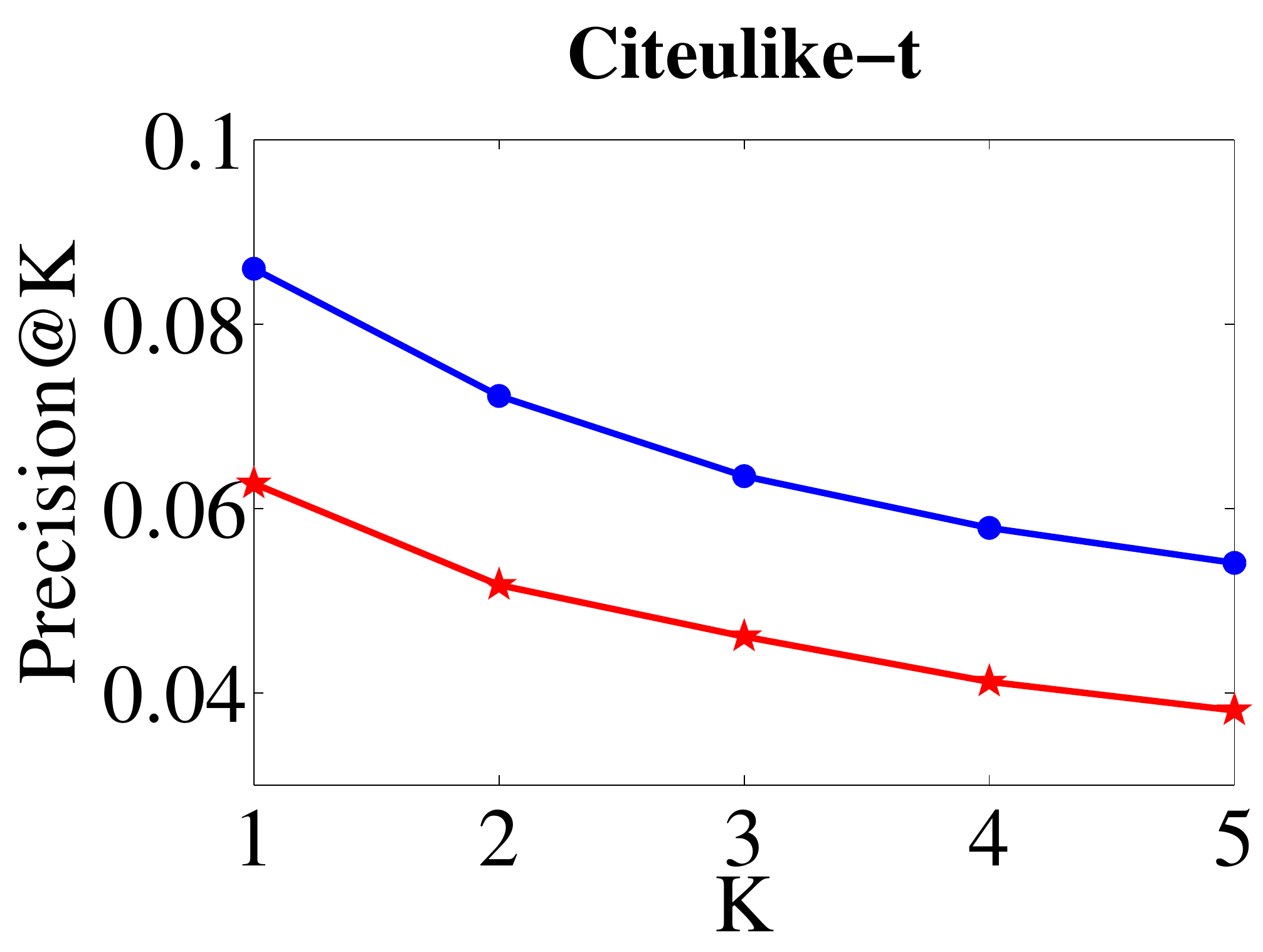}
	}
	\caption{Precision$@K$ of FRS based on Adam and SGD.}
	\label{fig:adam_sgd}
\end{figure*}
\clearpage
\subsection{Performance of defense methods on FRS based on other optimizers}
\label{sec:other_optimizer}

\begin{figure*}[ht]
	\centering
	\subfigure{
		\includegraphics[width=0.31\linewidth]{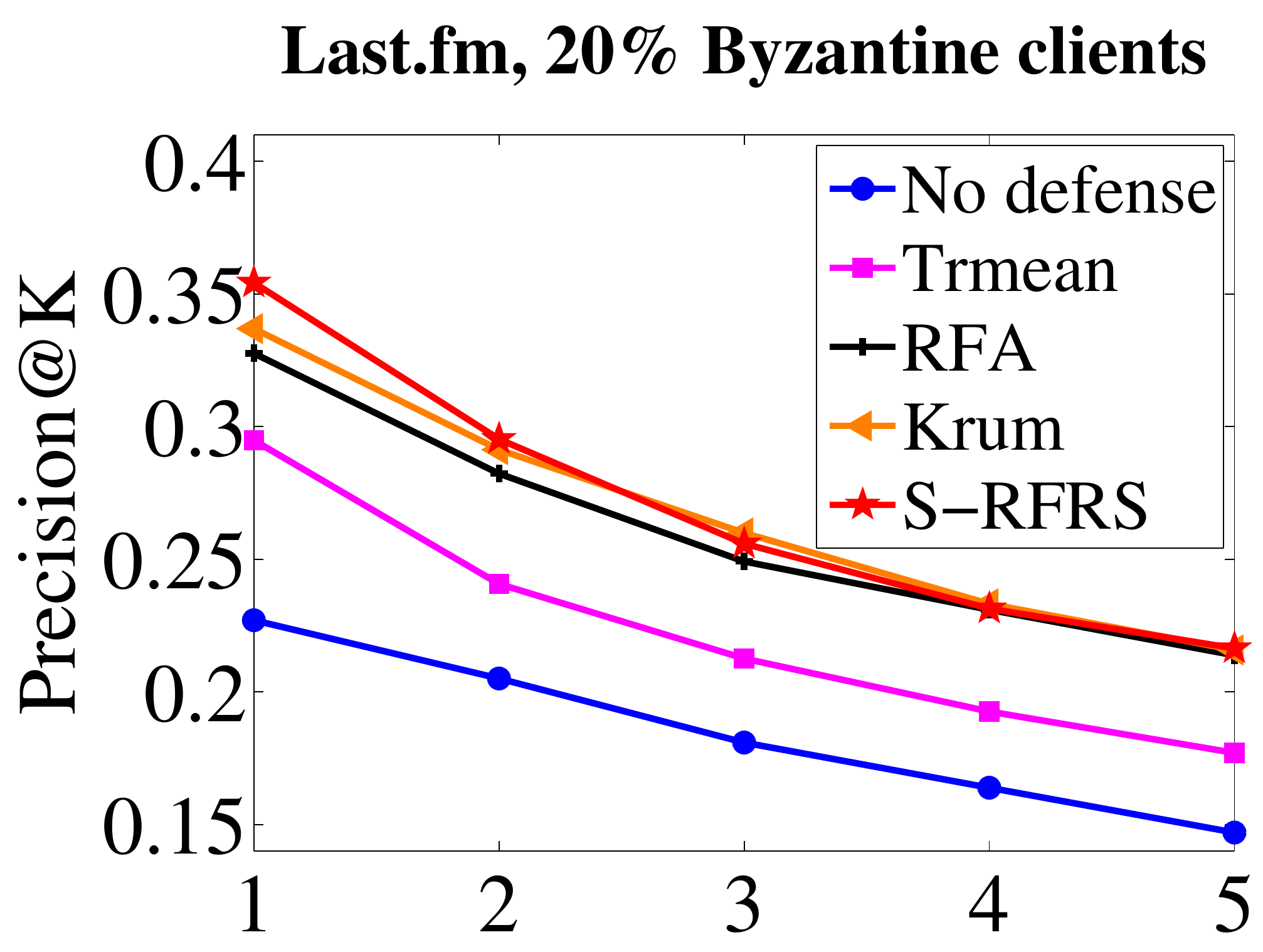}
	}
	\subfigure{
		\includegraphics[width=0.31\linewidth]{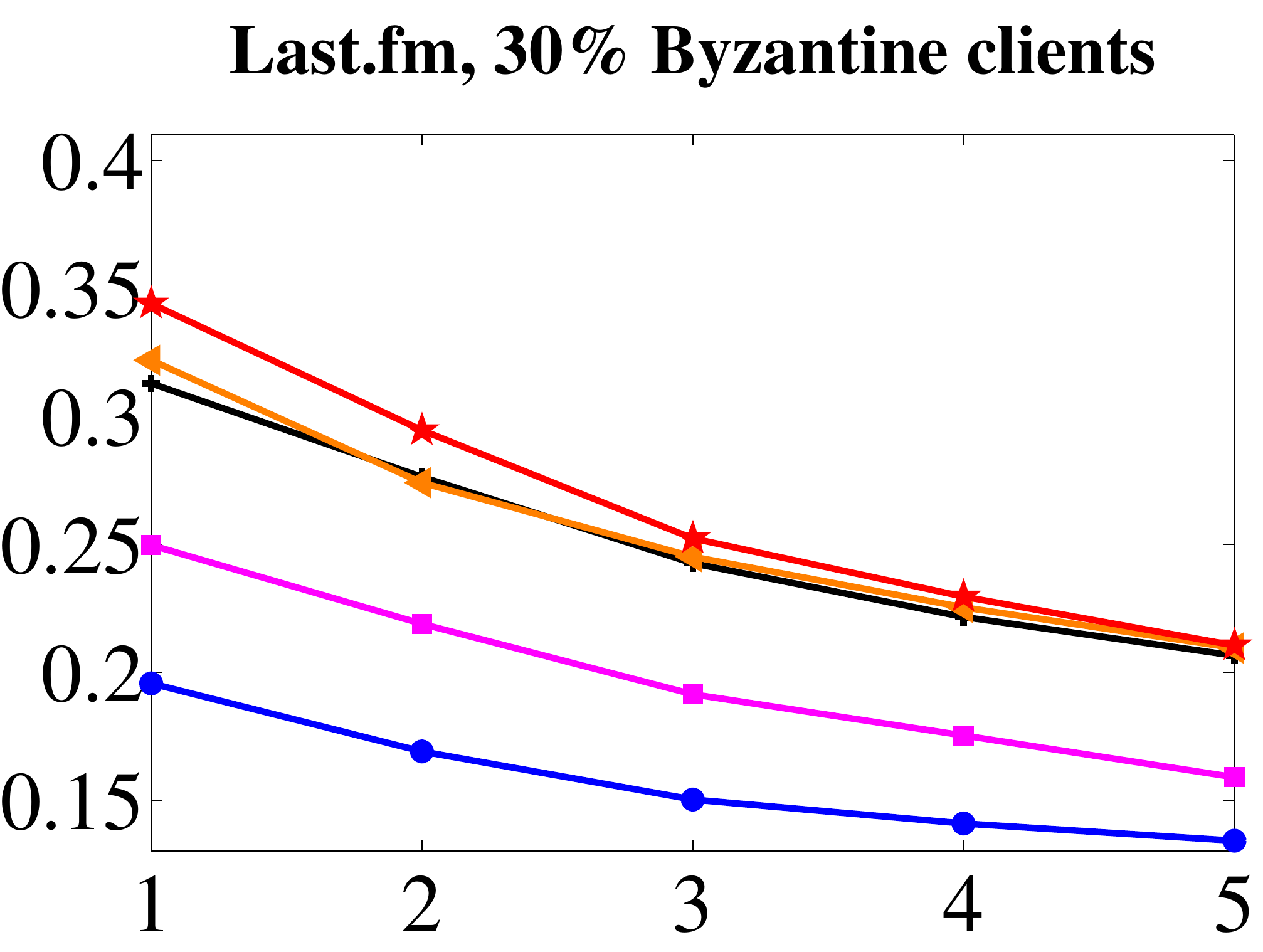}
	}
	\subfigure{
		\includegraphics[width=0.31\linewidth]{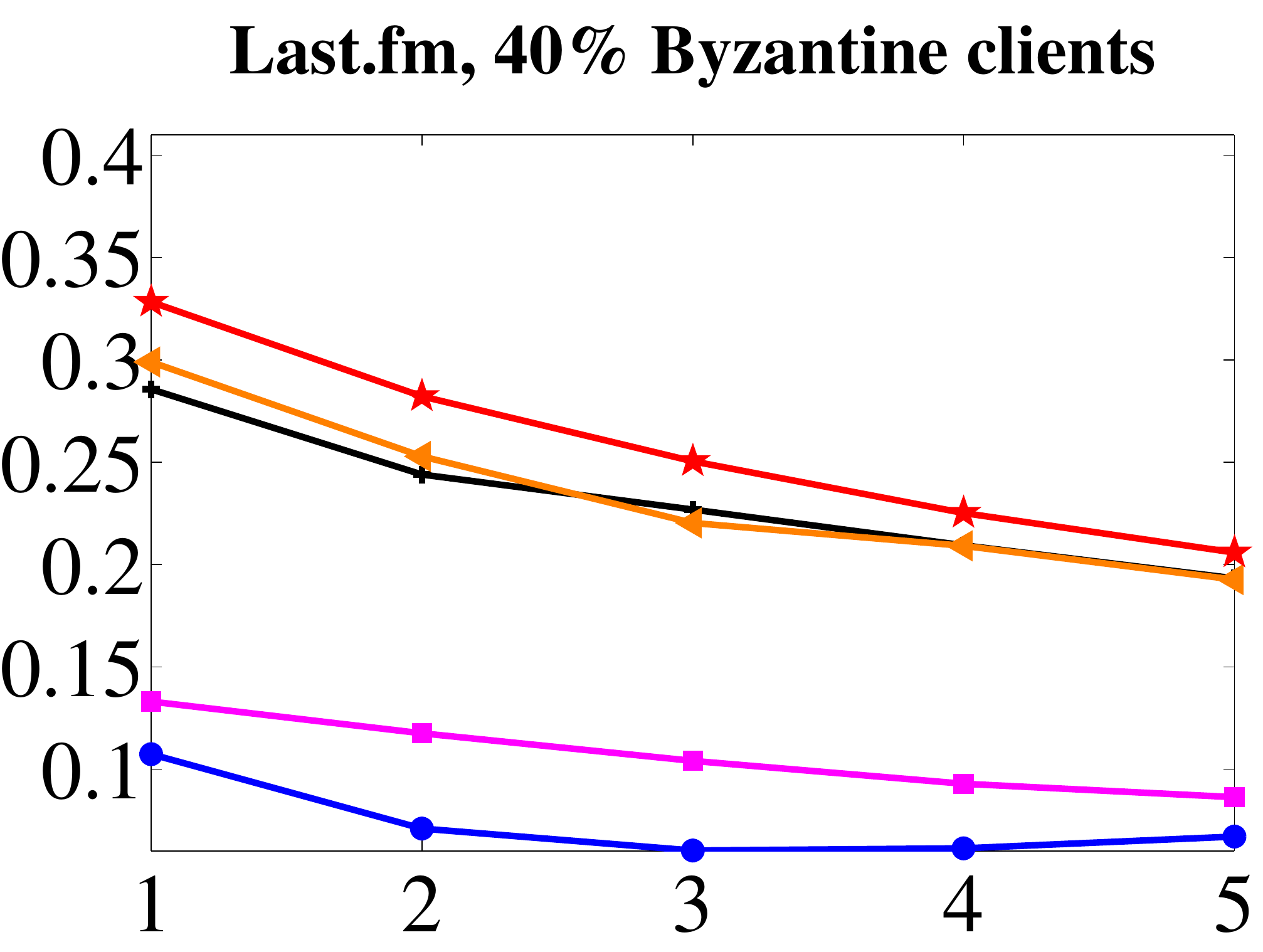}
	}
	\subfigure{
		\includegraphics[width=0.31\linewidth]{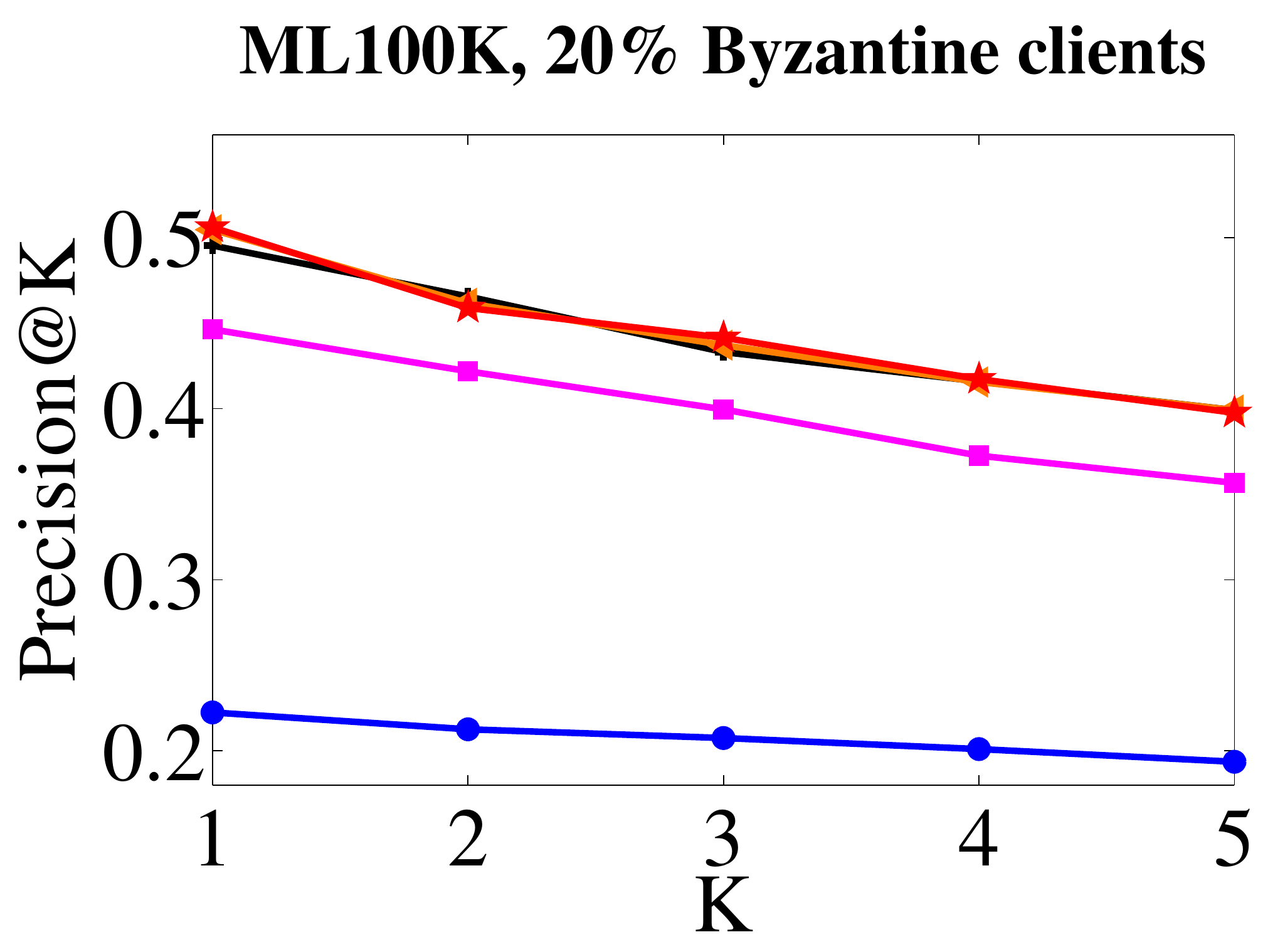}
	}
	\subfigure{
		\includegraphics[width=0.31\linewidth]{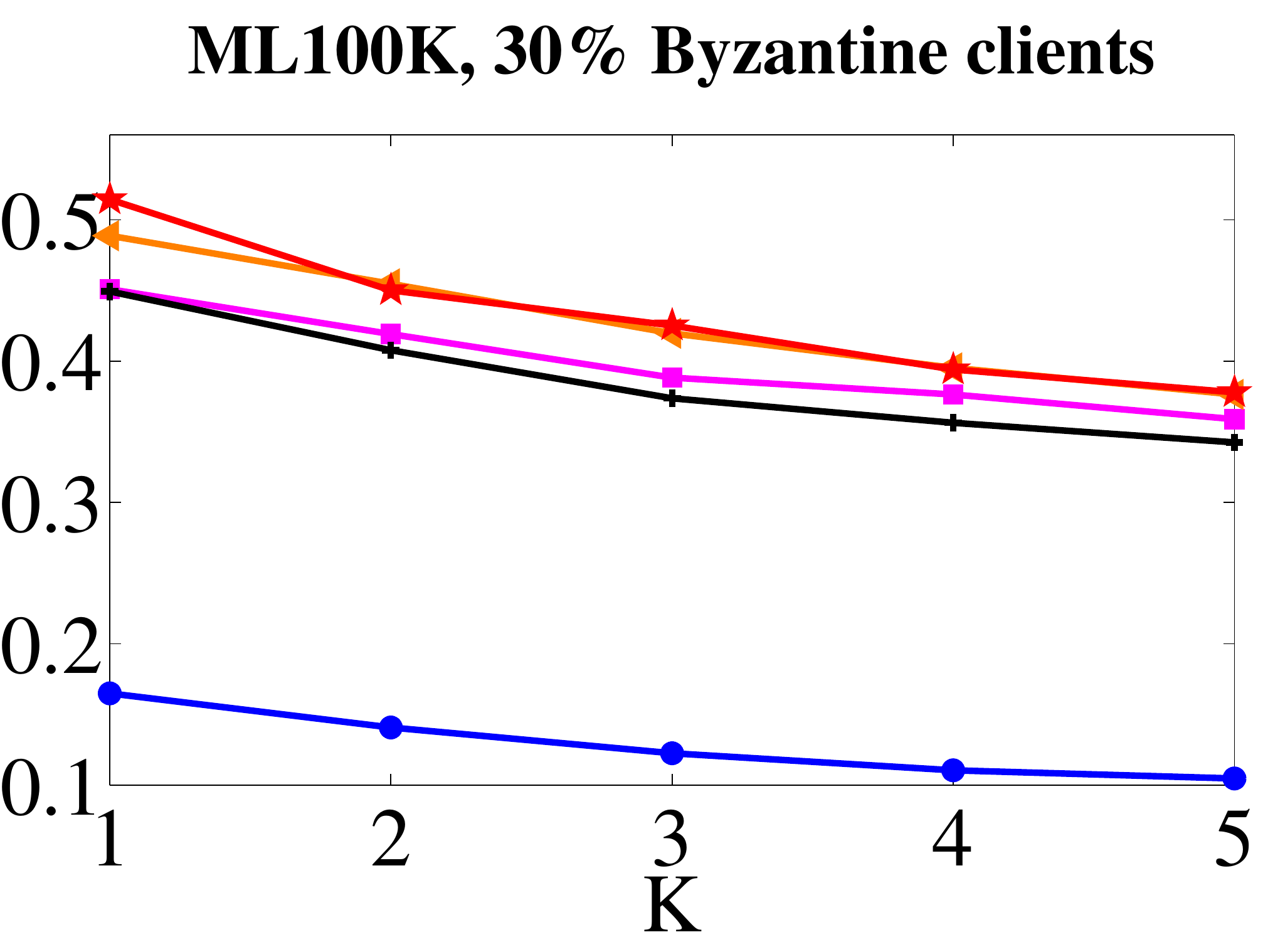}
	}
	\subfigure{
		\includegraphics[width=0.31\linewidth]{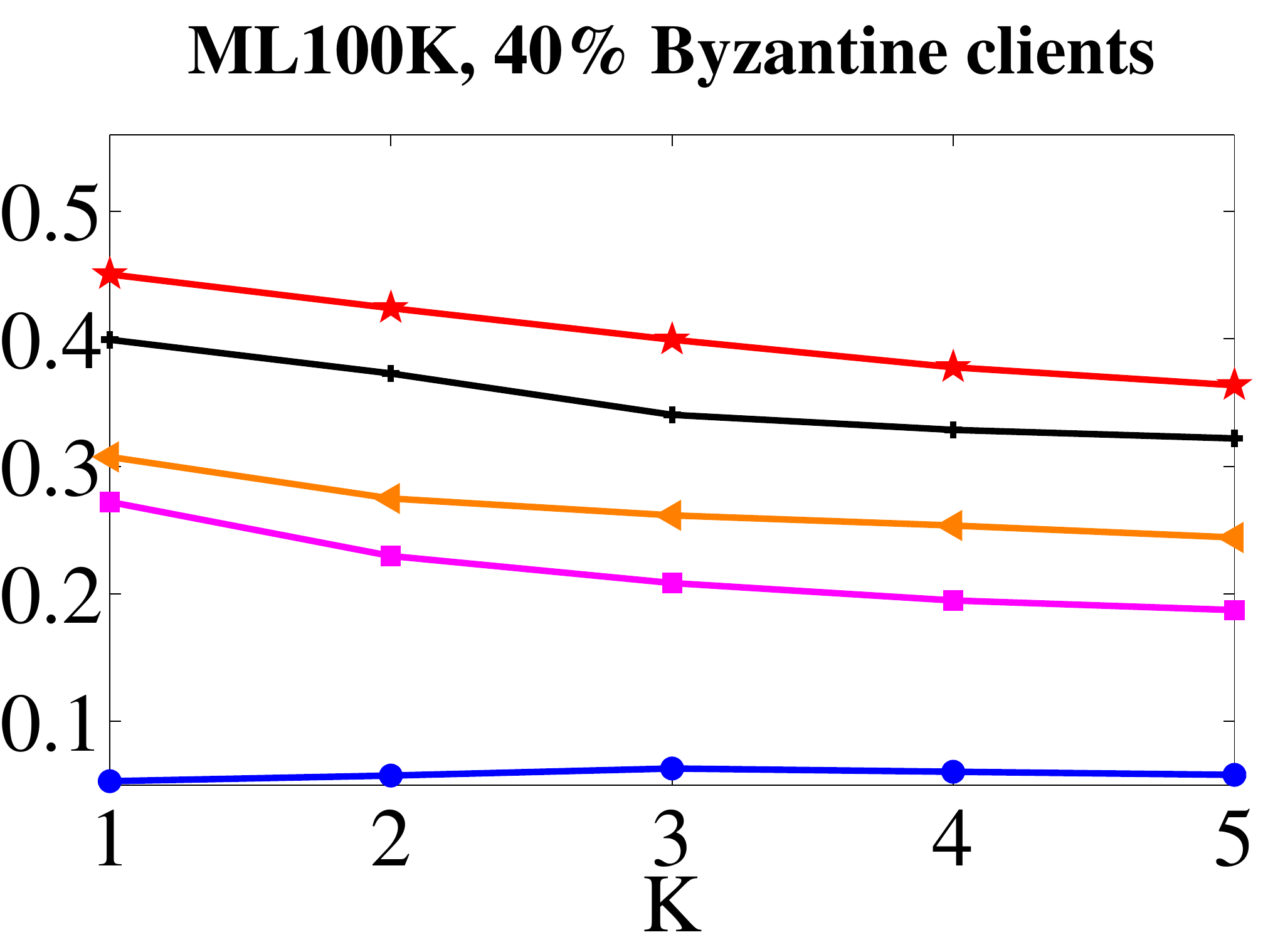}
	}
	\caption{Precision$@K$ of all defense methods in FRS based on SGD with momentum. S-RFRS (red line) is our proposed method.}
	\label{fig:momentum}
\end{figure*}
\begin{figure*}[ht]
	\centering
	\subfigure{
		\includegraphics[width=0.31\linewidth]{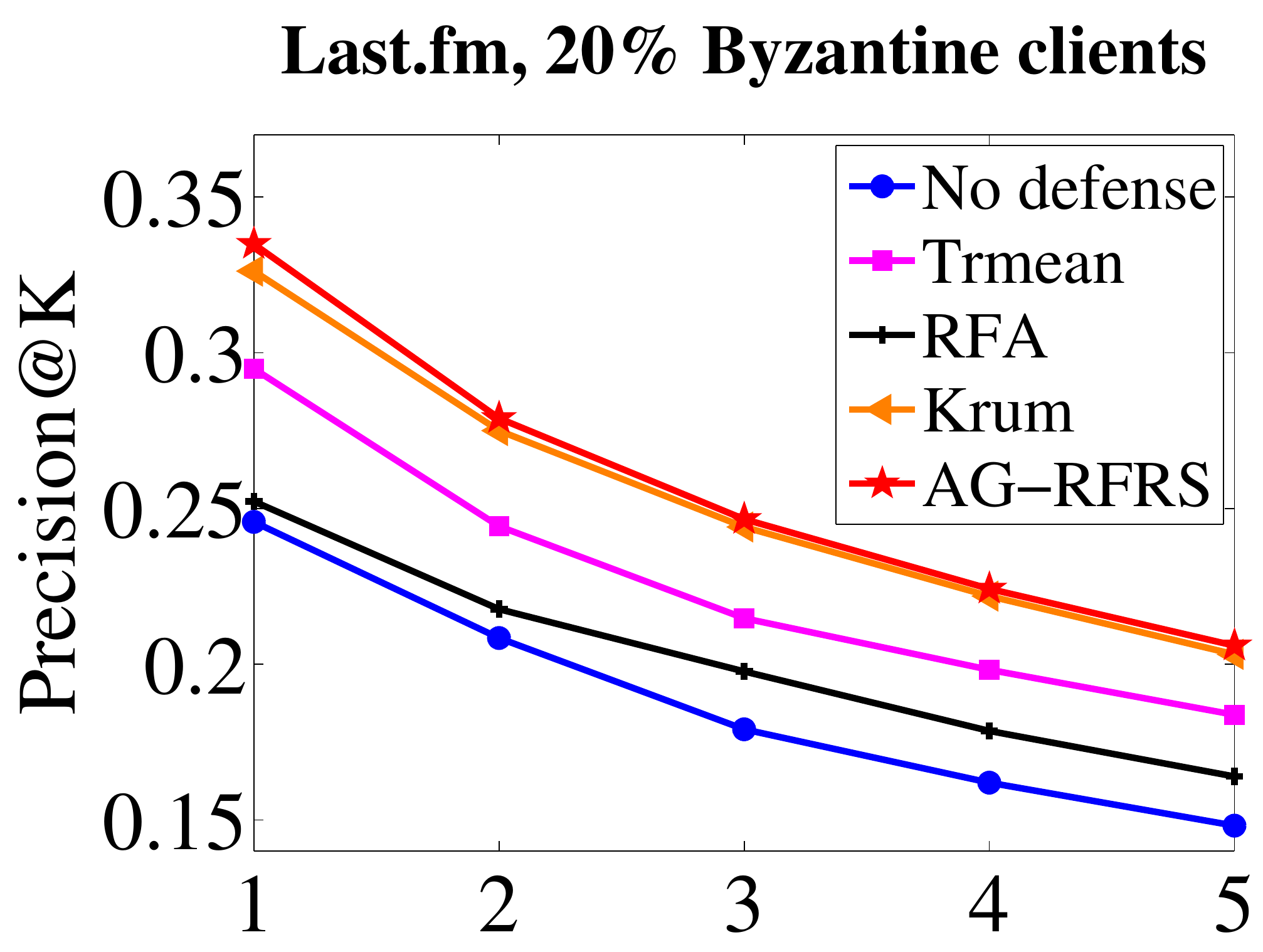}
	}
	\subfigure{
		\includegraphics[width=0.31\linewidth]{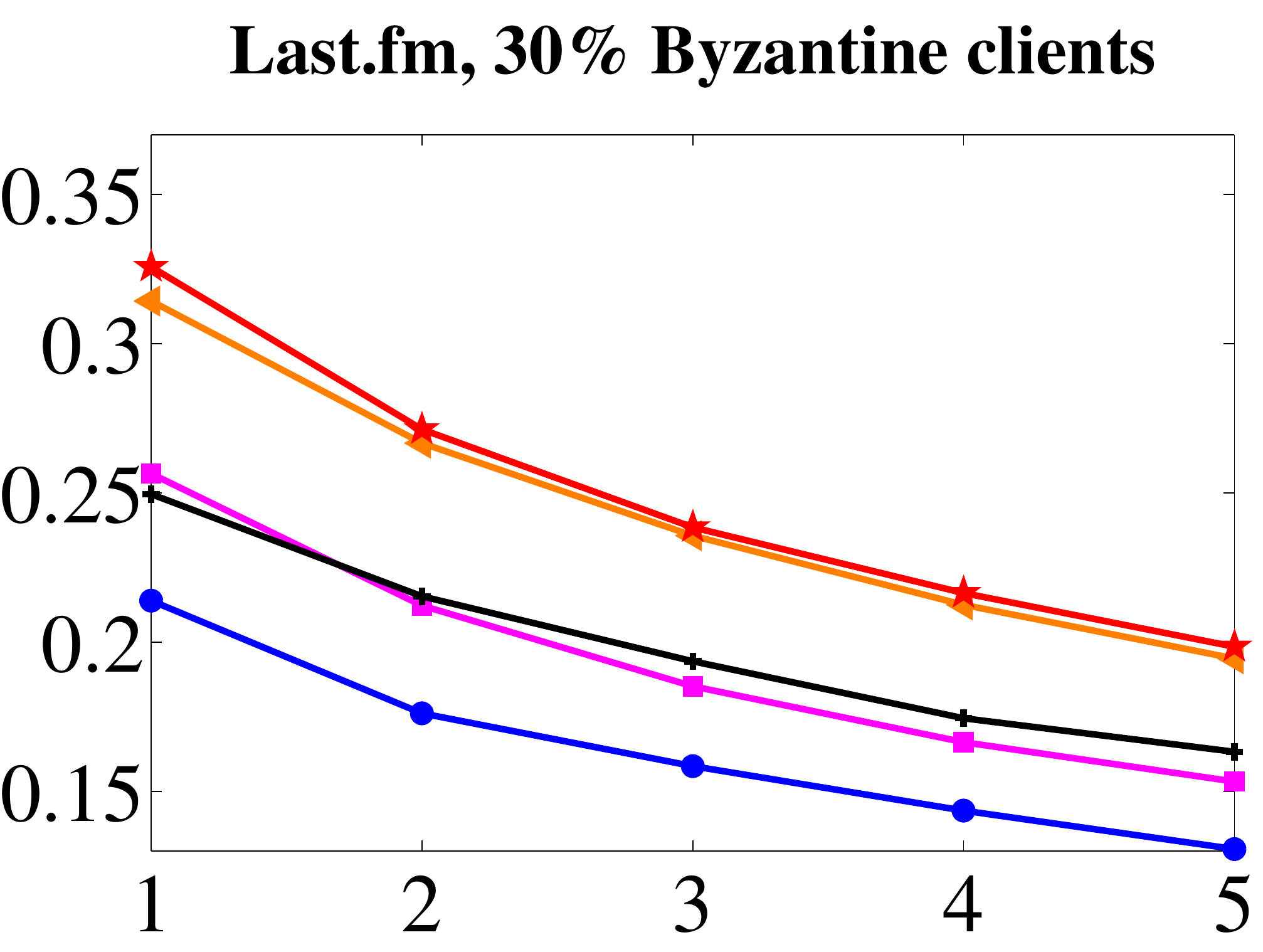}
	}
	\subfigure{
		\includegraphics[width=0.31\linewidth]{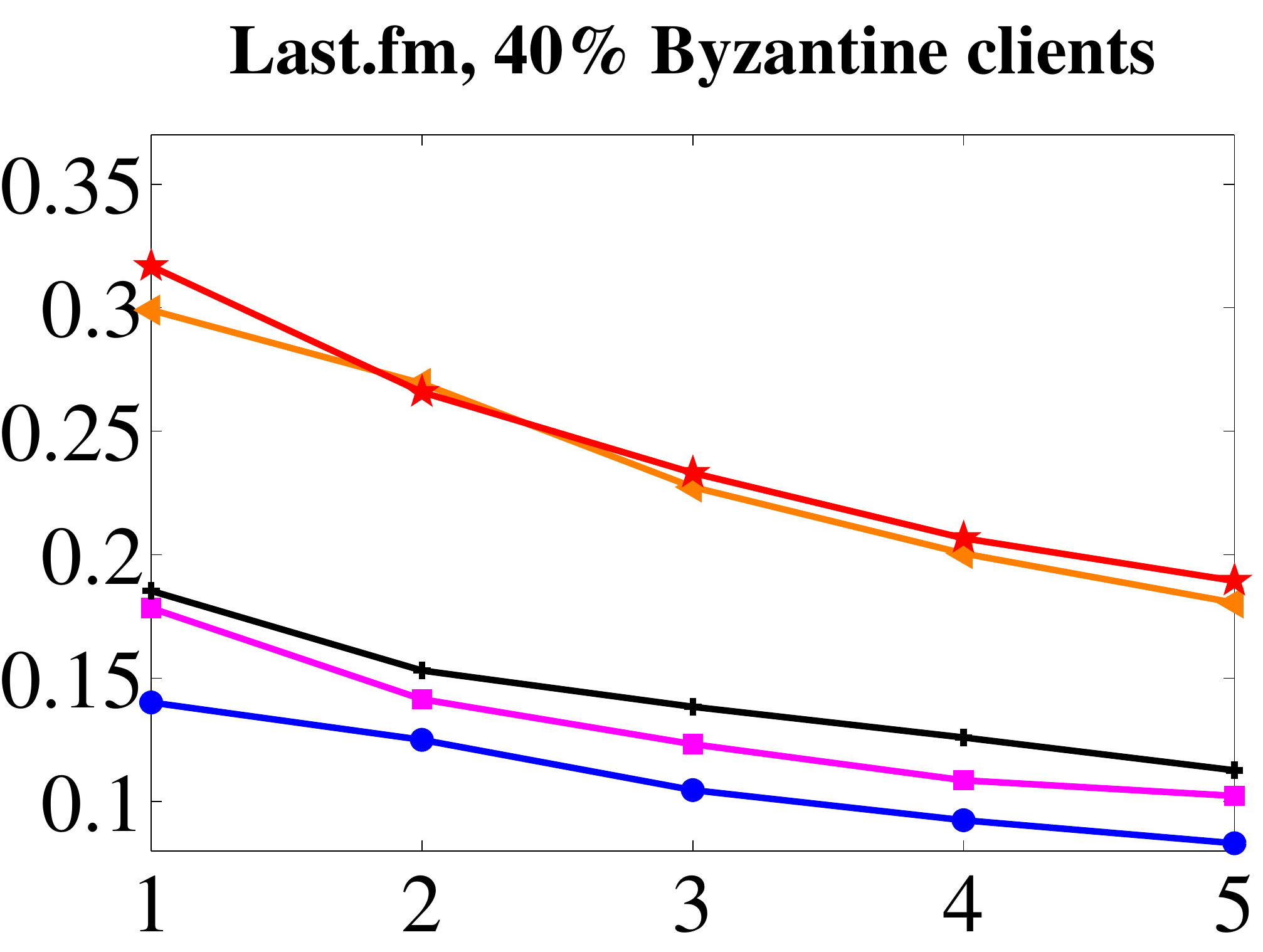}
	}
	\subfigure{
		\includegraphics[width=0.31\linewidth]{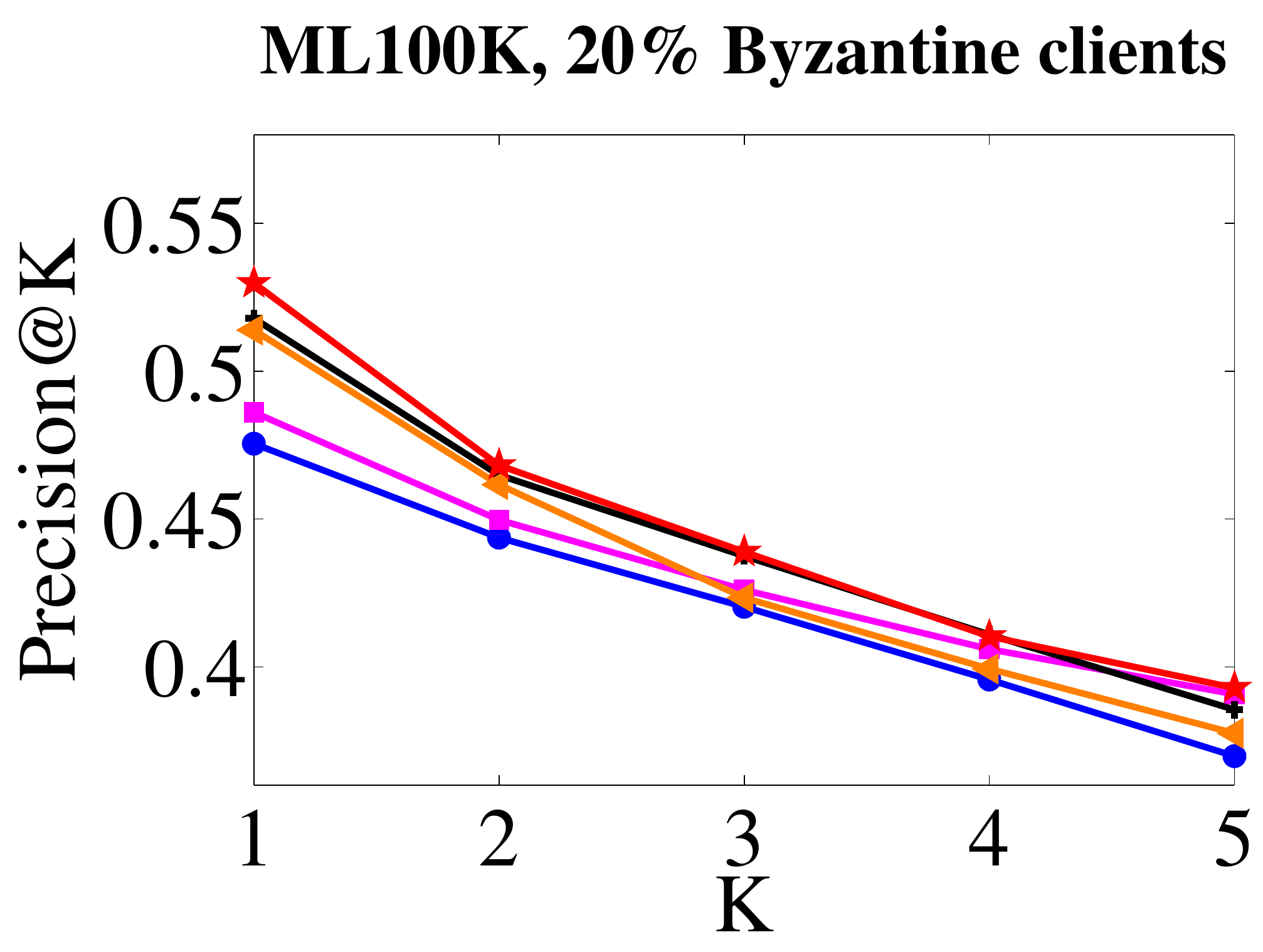}
	}
	\subfigure{
		\includegraphics[width=0.31\linewidth]{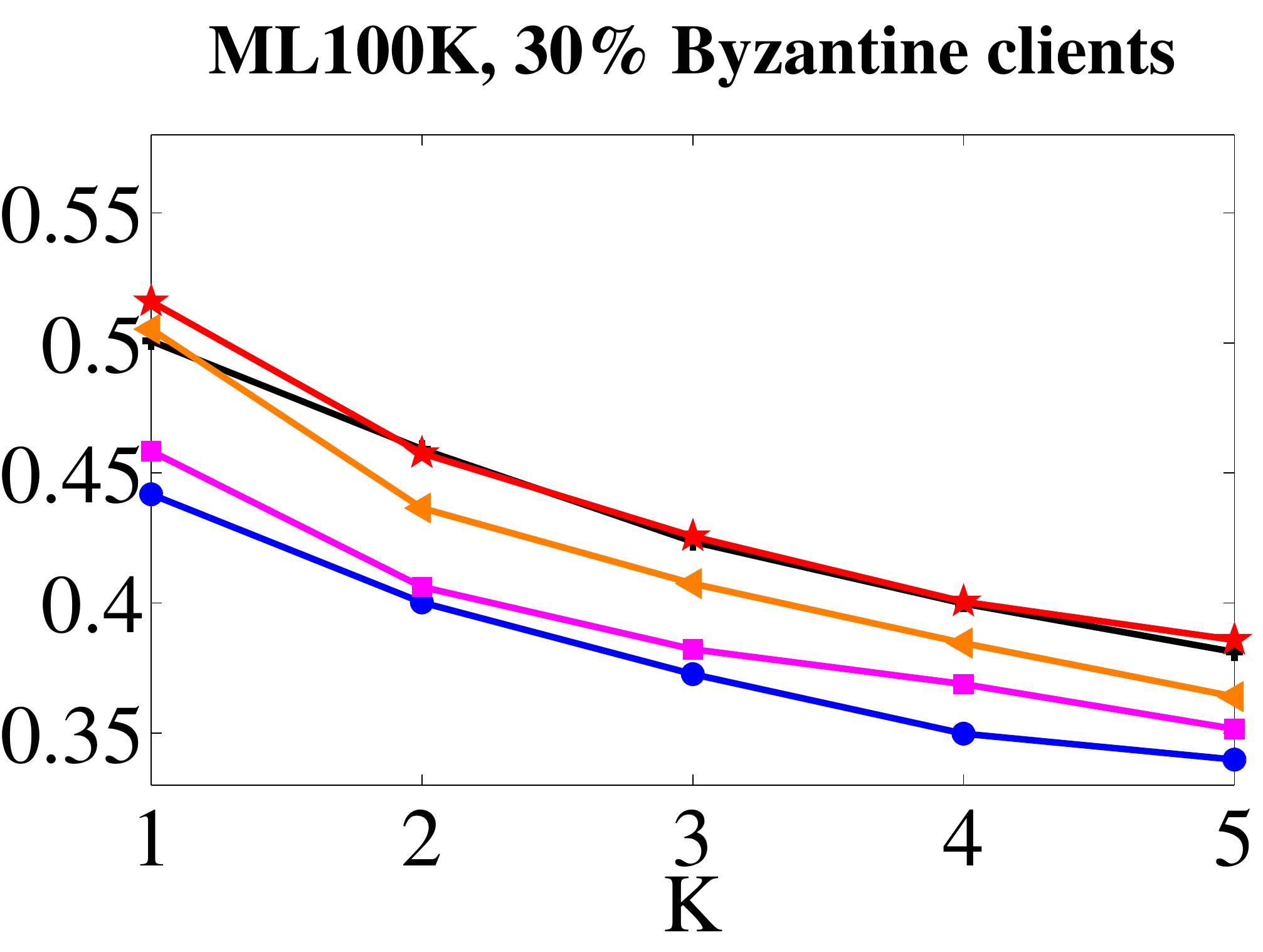}
	}
	\subfigure{
		\includegraphics[width=0.31\linewidth]{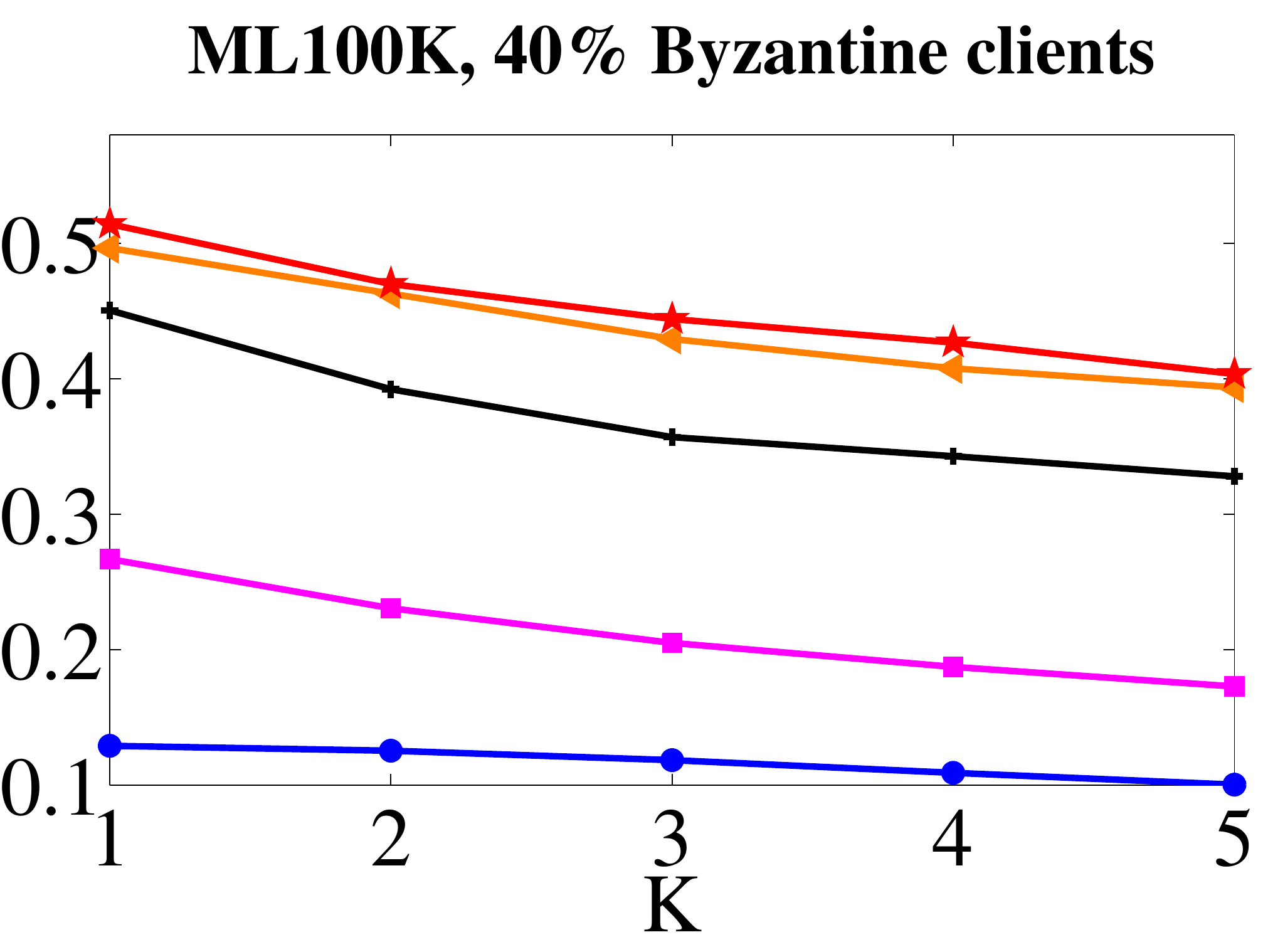}
	}
	\caption{Precision$@K$ of all defense methods in FRS based on AdaGrad. AG-RFRS (red line) is our proposed method.}
	\label{fig:adagrad}
\end{figure*}
\end{document}